\newif\ifisarxiv
\isarxivtrue 
\ifisarxiv

\documentclass[11pt]{article}
\usepackage[margin=2.5cm]{geometry}
\usepackage{natbib}
\usepackage{amsthm}
\else

\documentclass[11pt]{article}
\usepackage{jmlr2e}

\makeatletter
\def\NAT@def@citea{\def\@citea{\NAT@separator}}
\makeatother
\fi

\usepackage{amsmath, amssymb}
\usepackage{graphicx, graphics, epsfig, color}
\usepackage{adjustbox}
\usepackage{subfigure}
\usepackage{tikz, pgfplots}
\usetikzlibrary{plotmarks}
\pgfplotsset{compat=newest}
\usepackage{float}
\usepackage{multirow}
\usepackage{cuted, flushend}
\usepackage{midfloat}
\usepackage{bm}
\usepackage{fancyhdr}
\usepackage{enumerate}
\usepackage{siunitx}
\usepackage{hyperref}
\usepackage{cleveref}
\usepackage{subfigure}

\DeclareMathOperator{\tr}{ {\rm tr} }
\newcommand{\RR}{\mathbb{R}}
\newcommand{\T}{ {\sf T} }
\newcommand{\E}{\mathbb{E}}
\newcommand{\cd}{\xrightarrow{d}}
\newcommand{\cov}{{\rm Cov}}
\newcommand{\var}{{\rm Var}}
\newcommand{\x}{{\mathbf{x}}}
\newcommand{\zeros}{{\mathbf{0}}}
\newcommand{\ones}{{\mathbf{1}}}

\newcommand{\y}{{\mathbf{y}}}
\newcommand{\z}{{\mathbf{z}}}
\newcommand{\w}{{\mathbf{w}}}
\newcommand{\X}{{\mathbf{X}}}

\newcommand{\Z}{{\mathbf{Z}}}

\newcommand{\Q}{{\mathbf{Q}}}
\newcommand{\U}{{\mathbf{U}}}
\newcommand{\V}{{\mathbf{V}}}

\newcommand{\bmu}{ \boldsymbol{\mu} }
\newcommand{\bbeta}{ \boldsymbol{\beta} }
\newcommand{\bxi}{ \boldsymbol{\xi} }
\newcommand{\bdeta}{ \boldsymbol{\eta} }
\newcommand{\bnu}{ \boldsymbol{\nu} }
\newcommand{\bLambda}{ \boldsymbol{\Lambda} }
\newcommand{\bSigma}{ \boldsymbol{\Sigma} }

\newcommand{\C}{{\mathbf{C}}}
\newcommand{\I}{{\mathbf{I}}}

\renewcommand{\u}{{\mathbf{u}}}

\renewcommand{\c}{{\mathbf{c}}}
\renewcommand{\r}{{\mathbf{r}}}
\renewcommand{\v}{{\mathbf{v}}}
\renewcommand{\S}{{\mathbf{S}}}

\renewcommand{\P}{{\mathbf{P}}}

\DeclareMathOperator*{\argmin}{arg\,min}

\DeclareMathOperator{\prox}{{\rm prox}}

\definecolor{RED}{rgb}{0.7,0,0}
\definecolor{BLUE}{rgb}{0,0,0.69}
\definecolor{GREEN}{rgb}{0,0.6,0}
\definecolor{PURPLE}{rgb}{0.69,0,0.8}

\newtheorem{Theorem}{Theorem}
\newtheorem{Proposition}{Proposition}

\newtheorem{Corollary}{Corollary}

\newtheorem{Assumption}{Assumption}

\begin{document}

\title{High Dimensional Classification \\ via  Regularized and Unregularized Empirical Risk Minimization:\\ Precise Error and Optimal Loss}

\ifisarxiv
\author{
  Xiaoyi Mai \\ \texttt{maixiaoyi.31415@gmail.com}
  \and
  Zhenyu Liao\\ \texttt{zhenyu.liao@yahoo.com }
}
\date{\today}

\fi

\maketitle
	\begin{abstract}%
		This article provides, through theoretical analysis, an in-depth understanding of the classification performance of the empirical risk minimization framework, in both ridge-regularized and unregularized cases, when high dimensional data are considered. Focusing on the fundamental problem of separating a two-class Gaussian mixture, the proposed analysis allows for a precise prediction of the classification error for a set of numerous data vectors $\mathbf{x} \in \mathbb R^p$ of sufficiently large dimension $p$. This precise error depends on the loss function, the number of training samples, and the statistics of the mixture data model. It is shown to hold beyond Gaussian distribution under some additional non-sparsity condition of the data statistics. Building upon this quantitative error analysis, we identify the simple square loss as the optimal choice for high dimensional classification in both ridge-regularized and unregularized cases, regardless of the number of training samples. 
	\end{abstract}

	\tableofcontents
	
	\section{Introduction}
	\label{sec:intro}

	Consider the following general classification problem: given a training set of $n$ pre-labelled samples with feature vectors $\x_1 \ldots, \x_n \in \RR^p$ of dimension $p$, the objective is to predict the class label \(y\) (e.g., \(y=\pm 1\)) of a new observation \(\x\in\RR^p\) based on the knowledge of these training samples. The basic setup of a large number of classification algorithms is to obtain the class label $y$ of a new instance by combining its feature vector \(\x\) with a vector of weights \(\bbeta\in\RR^p\) such that \(y={\rm sign}(\bbeta^\T\x)\). The weight vector \(\bbeta\) is usually learned by fitting the known classes of the training samples, for example, by minimizing the classification error (also known as the 0--1 loss) on the training set $\{ (\x_i,y_i) \}_{i=1}^n$. Despite being a natural choice, the minimization of the non-convex 0--1 loss is known to be NP-hard \citep{ben2003difficulty}. To address this issue, the empirical risk minimization principle \citep{vapnik1992principles} suggests to obtain $\bbeta$ by minimizing a certain \emph{convex} surrogate of the 0--1 loss on the training set. Within this framework, the comparison between different designs of loss functions have been long discussed in the machine learning literature \citep{vapnik1992principles,rosasco2004loss,masnadi2009design}, mostly in the setting where the number of training samples $n$ largely exceeds the feature dimension $p$. Besides the computational convenience, the usage of convex loss functions is also theoretically supported by their property of leading to the same Bayes optimal solution that minimizes the 0--1 loss in the limit of $n/p \to \infty$ \citep{rosasco2004loss}. In spite of this remark, the classification accuracy is observed to significantly depend on the loss function when \(n\) is not exceedingly larger than \(p\). While it is of great practical interest to know which loss function to employ at a given number of training samples, much less is known in the more practical regime of finite \(n/p\). 
	
	\smallskip
	
	As a matter of fact, in the current paradigm of big data learning, it is typical to have the dimension $p$ comparable to or even greater than $n$, like for many tasks in the domains of biology \citep{guvenir1997supervised,khan2001classification,tsanas2013objective} and physics \citep{johnson2013classifying,lucas2015designing}.  Recent research works find that the generalization performance in the high dimensional $p \sim n$ regime can be drastically different from what classical statistical learning theory predicts \citep{belkin2018reconciling,hastie2019surprises,mei2019generalization}. In this work, we propose to investigate, under this potentially over-parametrized regime of high dimensional learning, the fundamental problem of classifying Gaussian mixture data using the popular approach of empirical risk minimization. More specifically, we consider data vectors sampled from two multivariate normal populations \(\mathcal{N}(\pm\bmu,\C)\), a classical setting underlying many popular statistical learning algorithms like linear discriminant analysis (LDA) and logistic regression (LR). These two popular methods can be retrieved as special cases under our unified analysis framework, as will be detailed in Section~\ref{sec:problem}. 
	
	
	\subsection{Related works}
	The high dimensional analysis of machine learning methods under the regime of comparably large \(n,p\) has attracted rapidly growing research interest. From a technical perspective, the high dimensional performance of classification methods with explicit solutions were evaluated in \citep{liao2019large,mai2018random,elkhalil2020large,dobriban2018high}, relying mostly on techniques from random matrix theory. When problems with implicit solutions such as M-estimators or logistic regression are concerned, commonly used technical tools to assess the asymptotic performance include:
	\begin{itemize}
	    \item the so-called ``double leave-one-out'' approach adopted in \citep{el2013robust} to study the statistical behavior of M-estimators with no closed-form solution, which hinges on the intuition that, in the large $n,p$ regime, the outcomes of algorithms should remain asymptotically unchanged after excluding one single sample from the training set or one single feature from the feature vector;
	    \item the Approximate Message Passing (AMP) and state evolution approach \citep{donoho2016high};
	    \item the Convex Gaussian Min-max Theorem (CGMT) first introduced by \citet{gordon1985some} and largely popularized by \citet{thrampoulidis2018precise}. 
	\end{itemize}
	Our technical derivation is similar in spirit to the  ``double leave-one-out'' approach, with substantial effort made to adapt to structured data model of correlated features and informative patterns.
	
	In the context of empirical risk minimization, \citet{sur2018modern,candes2020phase} investigated the logistic regression model in classifying ``unstructured'' data having i.i.d.\@ Gaussian features, i.e., $\x \sim \mathcal{N}(\mathbf{0}_p, \I_p)$. \citep{salehi2019impact} evaluated the impact of different regularization schemes (e.g., $\ell_1$ or $\ell_2$) in the context of logistic regression. \citet{mignacco2020role} discussed the impact of ridge regularization in the classification of (noise) Gaussian mixture on a generalized linear model. In pursuit of the optimal loss design within the empirical risk minimization framework, \citet{taheri2020sharp,taheri2020fundamental} derived the optimal convex loss for generalized linear models with non-structured data $\x\sim\mathcal{N}(\mathbf{0}_p,\I_p)$. They showed that the performance of least-squares is suboptimal, while being approximately optimal when the binary response $y$ follows a logistic model. This remark on the suboptimality of the square loss is in contrast with its absolute optimality established for the mixture data model under study.\footnote{Restricted to the Gaussian mixture model $\x_i\sim\mathcal{N}(\pm\bmu,\I_p)$ with \emph{identity covariance matrix} and \emph{unregularized} empirical risk minimization, the optimality of the square loss was also independently proven in \citep{taheri2020optimality}. An earlier version of the present manuscript (\href{https://arxiv.org/abs/1905.13742v1}{https://arxiv.org/abs/1905.13742v1}) had already established the optimality of the square loss in the same unregularized setting for $\x_i\sim\mathcal{N}(\pm\bmu,\C)$ with \emph{arbitrary} covariance matrix $\C$.} 
	
	\subsection{Our contribution}
	The aforementioned previous works being either placed under non-structured data or devoted to the study of particular algorithms without identifying the optimal loss, our study distinguishes itself with a double focus on (i) the classification of mixture-structured data and (ii) the optimal choice of surrogate loss.
	
	\smallskip
	
	Our main findings and contributions are summarized below.

		\begin{itemize}
			
			\item We fully describe, in Theorem~\ref{theo:main}, the high dimensional distribution of the empirical risk minimization solution $\hat \bbeta_\ell(\lambda)=\argmin_{\bbeta \in \RR^p} \frac1n \sum_{i=1}^n \ell(y_i\x_i^\T \bbeta) + \frac{\lambda}2 \| \bbeta \|^2$ for smooth convex loss function $\ell(\cdot)$ and regularization parameter $\lambda \ge 0$. Precisely, we find that $\hat \bbeta_\ell(\lambda)$ is arbitrarily close to a Gaussian random vector $\left(\lambda\I_p+\theta\C\right)^{-1} \times \mathcal{N}\left(\eta\bmu,\gamma^2\C/p\right)$ for sufficiently large $p$, where positive scalars $\theta,\eta,\gamma$ can be determined as functions of $n,p,\bmu$ and $\C$, or estimated empirically using $\hat \bbeta_\ell(\lambda)$ and the training samples $\{ (\x_i,y_i) \}_{i=1}^n$.
			
			\smallskip
			
			\item In terms of the bias-variance decomposition, we find that, irrespective of the loss function, the unregularized solution (if it exists) is aligned in expectation with the oracle solution \(\bbeta_*=2\C^{-1}\bmu\) (we refer to \eqref{eq:beta_star} below for more details on the oracle solution). Since the directional bias has no effect on the classification outcome of a new observation $\x$, which is determined by the sign of $\hat\bbeta^\T\x$, the classification accuracy of unregularized solutions is only affected by the (relative) variance. In the presence of a non-zero regularization term, our results help understand when and how the misclassification rate can be reduced for a better bias-variance trade-off.
			
			\smallskip
			
			\item We prove that the square loss (corresponding to the solution of LDA), rather than the logistic loss (corresponding to the maximal likelihood solution), is optimal in the unregularized case, as it is the one with minimal (relative) variance among all well-defined unregularized solutions (Theorem~\ref{theo:optimal-unreg}). Moreover, this optimality result is extended to show that the square loss solution outperforms, not only \emph{all} other unregularized solutions, but also \emph{any} linear combination of them (Theorem~\ref{theo:optim-linear-combination}).
			
			\smallskip
			
			\item Perhaps more surprisingly, even in the presence of a ridge-regularization term, the square loss remains optimal, in the sense that with an optimally tuned regularization $\lambda \ge 0$, it yields the lowest classification error (Theorem~\ref{theo:optimality reg}). In fact, our results also point out that at a controlled level of directional bias (i.e., with a fixed value of $\lambda/\theta$), the solution given by the square loss has the smallest variance.
			
			\item Finally we show, in Theorem~\ref{theo:universality over non-gaussian data}, that non-Gaussian mixture models have the same expression of high dimensional classification error under some additional non-sparsity assumption of the data statistics (see Assumption~\ref{ass:non-sparsity}). Consequently, the optimality of square loss in unregularized and regularized ERM also holds for generic non-Gaussian mixtures.
		\end{itemize}
	

	In the remainder of this article, we introduce the problem setting in Section~\ref{sec:problem}. The mains results of the precise classification error analysis is presented in Section~\ref{sec:statistical characterization}, with numerical validation provided in Section~\ref{sec:numerical validation}. Based on the precise error analysis, we discuss in Section~\ref{sec:optimality} the impact of loss function and in particular, the optimal loss choice in the unregularized and regularized cases.  The presented results are extended to non-Gaussian data under the additional Assumption~\ref{ass:non-sparsity} in Section~\ref{sec:universality}. The article closes with concluding remarks and envisioned extensions in Section~\ref{sec:conclusion}.

	\section{Problem Setup}
	\label{sec:problem}
	
	We start by introducing some notations that will be employed throughout the article. Boldface lowercase (uppercase) characters stand for vectors (matrices). The notation $(\cdot)^\T$ denotes the transpose operator. The norm $\|\cdot\|$ is the Euclidean norm for vectors and the operator norm for matrices. We follow the convention to use $o_P(1)$ for a sequence of random variables that convergences to zero in probability and $\cd$ for the convergence in distribution. We say that an event occurs with high probability if it happens with probability arbitrarily close to one for sufficiently large \(p\). 
	
	\medskip
	
	In accordance with the statistical hypotheses of linear discriminant analysis (LDA) and logistic regression (LR), we suppose that each data instance \((\x,y)\), with feature vector \(\x\in\RR^p\) and class label \(y=\pm 1\), is drawn independently and uniformly from the following mixture model:
	\begin{equation}\label{eq:mixture}
	\begin{cases}
	y = -1 &\Leftrightarrow \x \sim \mathcal{N}(-\bmu, \C),\\
	y = +1 &\Leftrightarrow \x \sim \mathcal{N}(+\bmu, \C),
	\end{cases}
	\end{equation}
	for some mean $\bmu \in \RR^p$ and positive definite covariance $\C \in \RR^{p \times p}$. The training set \(\{(\x_i,y_i)\}_{i=1}^n \) is composed of $n$ independent observations from \eqref{eq:mixture}. Let \(\X=[\x_1, \ldots, \x_n]\in\RR^{p\times n}\) be the feature matrix of training set, and \(\y=[y_1, \ldots, y_n]^\T\in\RR^p\) the class label vector. 
	
	Note that the conditional class probability $P(y\vert \x)$ is given by:
	 \begin{align*}
		P(y\vert \x)&=\frac{P(y)P(\x\vert y)}{P(\x)} =\frac{e^{-\frac{1}{2}(\x-y\bmu)\C^{-1}(\x-y\bmu)}}{e^{-\frac{1}{2}(\x-\bmu)\C^{-1}(\x-\bmu)}+e^{-\frac{1}{2}(\x+\bmu)\C^{-1}(\x+\bmu)}}=\frac1{1+e^{-2y\bmu^\T \C^{-1}\x}}.
		\end{align*} 
		In other words, $P(y\vert \x)= \sigma(y\bbeta_*^\T\x) $ with
		\begin{align}\label{eq:beta_star}
		\bbeta_*=2\C^{-1}\bmu \quad \text{and} \quad \sigma (t) = (1+e^{-t})^{-1}
		\end{align}
		the \emph{logistic sigmoid} function.
	As such, we shall refer to \(\bbeta_*\) as the oracle solution, or the vector of true parameters throughout this paper, which allows to recover the exact conditional class probability. For a new observation $\x$, it suffices to assign $\x$ to one of the two classes according to the sign of $\bbeta_*^\T\x$ in order to achieve the minimum classification error. In the common setting of supervised learning for classification, the goal is to establish an estimate  $\hat\bbeta$ of $\bbeta_*$ from the training samples such that a new datum $\x$ can be classified using the sign of $\hat\bbeta^\T\x$ with sufficiently small error. We notice that, under \eqref{eq:mixture}, the probability of misclassification is given as
		$$P(y\hat\bbeta^\T\x>0~\vert~\hat\bbeta)=Q \left( \hat\bbeta^\T\bmu/\sqrt{\hat\bbeta^\T\C\hat\bbeta} \right)$$
		for \( (\x,y)\) drawn from \eqref{eq:mixture} and independent of the training samples $\{ (\x_i,y_i) \}_{i=1}^n$, with $Q(t) = \frac1{\sqrt{2\pi}} \int_t^\infty e^{-u^2/2} du$ being the Q-function of the standard Gaussian distribution. In the sequel, we consider
		\begin{align}
		\label{eq:Err}
		{\rm Err}(\hat\bbeta)= Q \left( \hat\bbeta^\T\bmu/\sqrt{\hat\bbeta^\T\C\hat\bbeta} \right)
		\end{align}
		as an \emph{inverse} performance measure of $\hat\bbeta$, which is equal to the (expected) classification error of $\hat\bbeta$ on unseen data.
		
		The method of logistic regression (LR) is based on the principle of maximal likelihood \citep{friedman2001elements} and proposes to estimate \(\bbeta_*\) by
		\begin{equation}
		    \hat\bbeta_{{\rm LR}}= \argmin_{\bbeta \in \RR^p} \prod_{i=1}^n \sigma(y_i\bbeta^\T\x_i).
		\end{equation}
		It should be pointed out that the solution to the above optimization problem is unique and well-defined \emph{if and only if} there exists no hyperplane $\bbeta^\T\x=0$ that separates perfectly the two classes of training samples, in other words, \emph{if and only if} the training set is \emph{not linearly separable} and there does \emph{not} exists $\bbeta\in\RR^p$ such that $y_i\bbeta^\T\x_i\ge 0$ holds for \emph{all} $\{ (\x_i,y_i) \}_{i=1}^n$  \citep{albert1984existence}. This ill-defined issue is often settled by introducing an additional  regularization term
		$$\hat\bbeta_{{\rm LR}}(\lambda)= \argmin_{\bbeta \in \RR^p} \prod_{i=1}^n \sigma(y_i\bbeta^\T\x_i)+ \frac{\lambda}2 \Vert\bbeta\Vert^2,$$ 
		as the existence and uniqueness of $\hat\bbeta_{{\rm LR}}(\lambda)$ is always ensured for $\lambda>0$. The addition Tikhonov regularization term $\lambda\Vert\bbeta\Vert^2/2$ is also justifiable from the perspective of a better bias-variance trade-off and it is argued that the variance of the estimate $\bbeta$ is reduced at larger $\lambda$ at the cost of an increased bias \citep{friedman2001elements}.

		The approach of linear discriminant analysis (LDA), on the other hand, addresses the same classification problem by estimating separately the statistical parameters $\bmu$ and $\C$, with $\bbeta_*$ approximated by \citep{bishop2007pattern}
		\begin{align}\label{eq:beta_LDA}
		\hat \bbeta_{{\rm LDA}} =2\hat\C^{-1}\hat\bmu
		\end{align}
		where \(\hat\bmu=\frac{1}{n}\X\y \) and \( \hat\C=\frac{1}{n}\X\X^\T-\hat\bmu\hat\bmu^\T\) are respectively \emph{consistent} estimators of the population mean \(\bmu\) and covariance \(\C\), in the limit $n/p \to \infty$. Like LR, the solution of LDA is also not always well-defined. It is clear from \eqref{eq:beta_LDA} that the existence of $\hat \bbeta_{{\rm LDA}}$ depends on the invertibility of the sample covariance matrix $\hat\C$. For singular $\hat \C$, a similar regularization term can be introduced, leading thus to the following regularized LDA 
		\begin{align}\label{eq:beta_LDA reg}
		\hat \bbeta_{{\rm LDA}}(\lambda) =2\left(\lambda\I_p+\hat\C\right)^{-1}\hat\bmu.
		\end{align}
		
		It is easily shown that both $\hat\bbeta_{{\rm LDA}} (\lambda=0)$ and $\hat\bbeta_{{\rm LR}}(\lambda=0)$ give the oracle solution $\bbeta_*$ in the limit $n/p \to \infty$, the methods of LR and LDA thus yield the same oracle classification performance when provided with unlimited amounts of labelled data. Here we focus on comparing LR and LDA at finite $n/p$ (that is bounded away from zero), under the high dimensional regime of large $p,n$, as formally stated in the following assumption. As will be shown in the subsequent sections, at finite $n/p$, using regularized solutions with $\lambda>0$ can lead to improved classification accuracy thanks to a better bias-variance trade-off. 

	\begin{Assumption}[High dimensional asymptotics]\label{ass:growth-rate}
		The sample ratio \(n/p \in (0,\infty)\) uniformly for arbitrarily large \(p\). Also, $\| \bmu \| = O(1)$, $\|\C\| = O(1)$ and $\| \C^{-1} \| = O(1)$ with respect to $p$. 
		
	\end{Assumption}
	
	The conditions $\| \bmu \| = O(1)$ and $\|\C\| = O(1)$ imply that the data variation (between or within classes) on one direction remains finite at large $p$. We also ensure that the data vectors $\x_i$ are not lying on an intrinsically lower dimensional manifold with $\| \C^{-1} \| = O(1)$. 
	
	We propose to study the general problem of empirical risk minimization as follows,
	\begin{equation}\label{eq:opt-origin}
	\hat \bbeta_\ell(\lambda)=\argmin_{\bbeta \in \RR^p} \frac1n \sum_{i=1}^n \ell(y_i\x_i^\T \bbeta) + \frac{\lambda}2 \| \bbeta \|^2
	\end{equation}
	for some regularization parameter $\lambda \ge 0$ and $\ell$ some non-negative loss function satisfying Assumption~\ref{ass:loss}.  Sometimes we may use the shortened notation $\hat\bbeta$ for $\hat\bbeta_\ell(\lambda)$.
	
	\begin{Assumption}[Loss function]\label{ass:loss}
		The function $\ell: \RR \to \RR_+$ is convex and three-times continuously differentiable, with $\ell(0)$ finite, $\ell'(0)<0$ and $\ell''(t)$ bounded away from zero or infinity for any finite and non-negligible $t$.
	\end{Assumption}
	
	Assumption~\ref{ass:loss} is satisfied by many popular choices of loss function. In particular, with the logistic loss $\ell(t) = \ln(1+e^{-t})$ that gives the maximum likelihood estimate of $\bbeta_*$, we obtain the logistic regression classifier. The least squares classifier is given by the square loss $\ell(t) = (t-1)^2/2$. Another popular choice is the exponential loss $\ell(t) = \exp(-t)$, widely used in boosting algorithms \citep{freund1999short,rojas2009adaboost}. Even though our results such as precise error are derived assuming the smoothness of loss function $\ell$, they are well-defined for non-smooth losses with the help of proximal mapping (as explained below in \eqref{eq:def-prox}). This provides grounds for approximating a non-smooth loss $\ell$ by a family of smooth losses $\ell_\epsilon$ such that $\ell_\epsilon\to\ell$ as $\epsilon\to0$ in an appropriate sense \citep{el2013robust}.
	
	Remark importantly that the least squares classifier $\hat \bbeta_{{\rm LS}}(\lambda)$ induced by the square loss $\ell(t) = (t-1)^2/2$ is proportional to $\hat \bbeta_{{\rm LDA}}(\lambda)$. More precisely, since 
		\begin{align}
		\label{eq:beta_LS}
		\hat\bbeta_{{\rm LS}}(\lambda)=\left(\lambda\I_p+\frac1n \X\X^\T\right)^{-1} \frac1n \X\y,
		\end{align}
		we have by Sherman-Morrison formula that
		\begin{align*}
		\hat \bbeta_{{\rm LDA}}(\lambda)=2\left[1-\hat\bmu^\T(\lambda \I_p + \X\X^\T/n)^{-1}\hat\bmu\right]^{-1}\hat \bbeta_{{\rm LS}}(\lambda)
		\end{align*}
		where $\hat\bmu^\T(\lambda \I_p + \X\X^\T/n)^{-1}\hat\bmu<1$ since $$\hat\bmu^\T(\lambda \I_p + \X\X^\T/n)^{-1}\hat\bmu=\frac{1}{n^2}\y^\T\X^\T(\lambda \I_p + \X\X^\T/n)^{-1}\X\y<\frac{1}{n^2}\y^\T\X^\T( \X\X^\T/n)^{-1}\X\y<\frac{1}{n}\Vert\y\Vert^2.$$ It is easy to see from \eqref{eq:Err} that ${\rm Err}(\hat\bbeta)={\rm Err}(\alpha\hat\bbeta)$ for any $\alpha>0$, $\hat \bbeta_{{\rm LS}}(\lambda)$ and $\bbeta_{{\rm LDA}}(\lambda)$ give thus the same classification error. In the following, we will speak interchangeably of $\hat \bbeta_{{\rm LS}}(\lambda)$ and $\bbeta_{{\rm LDA}}(\lambda)$ when it comes to the classification performance.


	
	\section{Statistical Characterization}
	\label{sec:statistical characterization}

	Before introducing our main theoretical results, we define some notations that will be used in the theorem. By cancelling the derivative of objective function of the general optimization problem in \eqref{eq:opt-origin}, we obtain $\lambda \hat \bbeta = \frac1n \X_\y \c$ with $\X_\y=[y_1\x_1,\ldots,y_n\x_n] \in \RR^{p \times n}$ and
	\begin{equation}\label{eq:def-c}
	\c=[c_1,\ldots,c_n]^\T =[-\ell'(y_1\x_1^\T \hat \bbeta),\ldots,-\ell'(y_n\x_n^\T \hat \bbeta)]^\T
	\end{equation}
	where we denote \(\ell'(t)\) the derivative of the loss function $\ell$. Additionally, let
	\begin{equation}\label{eq:def-r}
	\r = [r_1, \ldots, r_n]^\T = [y_1\x_1^\T \hat \bbeta - \hat \kappa c_1,\ldots,y_n\x_n^\T \hat \bbeta - \hat \kappa c_n]^\T,
	\end{equation}
	with
	\begin{equation}\label{eq:def-hat-kappa}
	\hat \kappa=\frac{1}{n}\sum_{i=1}^n\frac{\x_i^\T\Q\x_i/n}{1-\ell''(y_i\x_i^\T \hat \bbeta)\x_i^\T\Q\x_i/n},
	\end{equation}
	where $\Q=\left( \frac{1}{n}\sum_{i=1}^n\ell''(y_i\x_i^\T \hat \bbeta)\x_i\x_i^\T + \lambda \I_p \right)^{-1}$.
	
    An interesting property of $\c$ and $\r$ is that they can be computed directly from the training samples $\{(\x_i,y_i) \}_{i=1}^n$ and the learning classifier $\hat\bbeta$ (which is itself function of the training samples). In other words, both $\c$ and $\r$ can be obtained empirically from the training set without knowledge of the statistical parameters $\bmu$ and $\C$ of the data model (which are generally not accessible in practice). As will be shown in our theoretical results, the performance of $\hat\bbeta$ can be assessed by some quantities dependent of $\c$ and $\r$, through what we call a stochastic description of $\hat\bbeta$. 
		
	Our theoretical results also allows one to characterize the asymptotic distribution of $\hat\bbeta$ that depends solely on $\bmu$, $\C$, and $n/p$, without solving the optimization problem \eqref{eq:opt-origin}. To this end, we define, for $\kappa > 0$, the mapping $g: \RR \to \RR$
		\begin{equation}
		\label{eq:def-g}
		g_{\kappa,\ell}(t)=f^{-1}_{\kappa,\ell}(t)
		\end{equation}
		as the inverse function of 
		$$f_{\kappa,\ell}(t)=t+\kappa\ell'(t).$$ It is interesting to note that $g_{\kappa,\ell}(t)$ can be interpreted as a proximal mapping
		\begin{equation}\label{eq:def-prox}
		    g_{\kappa,\ell}(t) = \prox_{\kappa, \ell} (t)=\argmin_{a \in \RR} \ell(a) + \frac1{2\kappa} (a - t)^2,
		\end{equation}
		which is a common object in modern convex optimization problems, from both theoretical \citep{bauschke2011convex} and practical perspectives \citep{parikh2014proximal}. As the proximal mapping $\prox_{\kappa, \ell}(t)$ (and thus $g_{\kappa,\ell}(t)$) is well defined for nondifferentiable convex function $\ell$, it can be computed for a wider range of loss functions outside the scope of this article.

	\subsection{Characterization of a single classifier}

	With the above notations, we introduce now the main technical result of this article, which concerns a stochastic as well as  a deterministic description of the classifier $\hat \bbeta$ defined in \eqref{eq:opt-origin}. We refer to Appendix~\ref{sm:proof-of-theorem-main} for the proof.
	
		\begin{Theorem}[Stochastic and deterministic characterizations of $\hat \bbeta$]\label{theo:main}
			Let Assumptions~\ref{ass:growth-rate}~and~\ref{ass:loss} hold. Then, for any  $\lambda>0$,
			$$\Vert \hat \bbeta_\ell(\lambda)-\tilde \bbeta_\ell(\lambda)\Vert=o_P(1)$$
			where
			\begin{equation}
			\label{eq:tilde beta}
			\tilde \bbeta_\ell(\lambda) = \left(\lambda\I_p+\theta\C\right)^{-1}\left(\eta\bmu+\gamma\C^{\frac{1}{2}}\u\right)
			\end{equation}
			with $\u\sim\mathcal{N}(\zeros_p,\I_p/p)$ and  $\theta, \eta, \gamma$ being positive deterministic constants that 
			\begin{enumerate} 
			\item satisfy the fixed point equations
				\begin{equation}
				\label{eq:theta eta gamma}
				\theta=\frac{-\cov[h(r),r]}{\var[r]},\quad \eta=\E[h(r)],\quad \gamma=\sqrt{\frac{p\E[h^2(r)]}{n}}
				\end{equation}
				where $r \sim \mathcal N(m,\sigma^2)$ with
				\begin{align}
				m=&\eta\bmu^\T\left(\lambda\I_p+\theta\C\right)^{-1}\bmu\nonumber\\ \sigma^2=&\eta^2\bmu^\T\left(\lambda\I_p+\theta\C\right)^{-1}\C\left(\lambda\I_p+\theta\C\right)^{-1}\bmu+\gamma^2\tr\left[\left(\lambda\I_p+\theta\C\right)^{-1}\C\right]^2/p,\label{eq:m sigma}
				\end{align}
				and $h(r) = \frac{g_{\kappa, \ell} (r)-r}{\kappa}$ with $g_{\kappa, \ell} (\cdot)$ as defined in \eqref{eq:def-g} and $\kappa$ given by 
				\begin{align}
				\label{eq:kappa}
				\kappa = \frac1n \tr \C \left( \theta\C + \lambda \I_p \right)^{-1};
				\end{align}
				\item and can be consistently estimated by
				\begin{equation}
				\label{eq:hat theta eta gamma}
				\hat\theta=\frac{-\c^\T[\r-(\ones_n^\T\r/n)\ones_n]}{\Vert\r-(\ones_n^\T\r/n)\ones_n\Vert^2},\quad \hat\eta=\frac{\ones_n^\T\c}{n},\quad \hat\gamma=\frac{\sqrt{p}\Vert\c\Vert}{n}
				\end{equation}
				with $\c$ and $\r$ defined in \eqref{eq:def-c} and \eqref{eq:def-r}, in the sense that $\left\Vert [\theta, \eta, \gamma]-[\hat\theta, \hat\eta, \hat\gamma] \right\Vert=o_P(1)$.
			\end{enumerate}
			Additionally, if $\hat\bbeta_\ell(0)$ is unique with bounded norm for some $n/p> 1$, then we have
			\begin{equation*}
			\left\Vert \hat \bbeta_\ell(0)-\tilde \bbeta_\ell(0)\right\Vert=o_P(1).
			\end{equation*}
			
		\end{Theorem}
		
		Remark first that the above theorem unifies the regularized and unregularized solutions, by saying that, the unregularized solution $\hat \bbeta_\ell(0)$, if being unique and well-defined, can be retrieved by taking $\lambda=  0$ in $\tilde \bbeta_\ell(\lambda)$. As already discussed in Section~\ref{sec:problem}, the existence and uniqueness of the unregularized solution $\hat \bbeta_\ell(0)$ with bounded norm is not always guaranteed, as opposed to the regularized solution $\hat \bbeta_\ell(\lambda)$ with $\lambda>0$. For instance, any loss function $\ell$ with a finite minimizing point $t_{\rm min} = \argmin_t \ell(t)$ yields an infinity of solutions to \eqref{eq:opt-origin} at $\lambda=0$ and $n<p$, as the system of linear equations $y_i\x_i^\T\bbeta=t_{\rm min}$, $i\in\{1,\ldots,n\}$ has infinitely many solutions. This problem can be settled by restricting to $n/p>1$ for the discussion of unregularized classifiers. The condition $n/p>1$ alone however does not ensure a well-behaved unregularized solution $\hat\bbeta_{{\rm LR}}(0)$ for logistic loss.
		The existence of a well-defined $\hat\bbeta_{{\rm LR}}(0)$ was characterized in \citep{candes2020phase} under the same high dimensional regime, however on a different data model. In the remainder of this article, whenever we speak of the unregularized solution $\hat\bbeta_\ell(0)$, it is assumed to be well-defined.
		\medskip
		
		Theorem~\ref{theo:main} provides access to the asymptotic distribution of the solution $\hat\bbeta$, parametrized by three scalar variables $\theta$, $\eta$ and $\gamma$. Roughly speaking, the ``orientation'' of $\hat\bbeta$ is affected by $\theta$, while $\eta$ and $\gamma$ impact the norm of the deterministic information in $\hat\bbeta$ and its random fluctuation, respectively. In fact, it can be shown that the classification error ${\rm Err}(\hat\bbeta)$ (defined in \eqref{eq:Err}) converges to a deterministic value dependent of $\theta$, $\eta$ and $\gamma$: from \eqref{eq:tilde beta} we have that ${\rm Err}(\hat\bbeta_\ell(\lambda))={\rm Err}(\tilde\bbeta_\ell(\lambda))+o_P(1)$, which, together with standard concentration arguments
		\begin{align*}
		&\tilde\bbeta_\ell(\lambda)^\T\bmu=\E[\tilde \bbeta_\ell(\lambda)]^\T\bmu+o_P(1)=m+o_P(1),\\
		&\tilde\bbeta_\ell(\lambda)^\T\C\tilde\bbeta_\ell(\lambda)=\E[\tilde \bbeta_\ell(\lambda)^\T\C\tilde \bbeta_\ell(\lambda)]+o_P(1)=\sigma^2+o_P(1),
		\end{align*}
		leads to the following corollary.
		\begin{Corollary}[Classification error rate]
			\label{cor:classification error}
			Under the conditions of Theorem~\ref{theo:main}, we have
			\begin{equation}
			\label{eq:Err asymtotic}
			{\rm Err}(\hat\bbeta_\ell(\lambda))= Q\left(\frac{m}{\sigma}\right)+o_P(1)
			\end{equation}
			with $m,\sigma^2$ given in \eqref{eq:m sigma} and $Q(\cdot)$ the Gaussian $Q$-function.
		\end{Corollary}
		The values of $\theta$, $\eta$ and $\gamma$ can be computed in a deterministic manner through the fixed-point equations in \eqref{eq:theta eta gamma}, or alternatively be estimated empirically from the training set $\{ (\x_i, y_i) \}_{i=1}^n$, using the estimates in \eqref{eq:hat theta eta gamma}. These two alternative descriptions of $\hat\bbeta$ can help us understand the learning results from different perspectives. As an illustrative example, our discussion on the optimal performance in Section~\ref{sec:optimality} will make use of both descriptions.
		
		One may notice by comparing \eqref{eq:hat theta eta gamma} and \eqref{eq:theta eta gamma} that there is a connection between $c_i$ and $h(r)$. For instance, $\hat\eta$ is the empirical mean of $c_i$ and $\eta$ equals the expectation $\E[h(r)]$. As a matter of fact, as an intermediary result in the proof of Theorem~\ref{theo:main} in Appendix~\ref{sm:proof-of-theorem-main}, the random variable $c_i$ can be shown to converge in distribution to $h(r)$ for $r \sim \mathcal N(m, \sigma^2)$ in the limit of large $p$, as described in Proposition~\ref{prop:distribution c_i} below. We also remark  that, asymptotically, $r$ has the same distribution  as $y\x^\T\tilde\bbeta_\ell(\lambda)$ for some $(\x,y)$ drawn from \eqref{eq:mixture} and independent of $\tilde\bbeta_\ell(\lambda)$. This is in accordance with \eqref{eq:Err asymtotic} where the (expected) classification error of $\tilde\bbeta_\ell(\lambda)$ is well approximated by the probability $P(r>0) = Q(m/\sigma)$.
		\begin{Proposition}[Asymptotic distribution of $c_i$ and $r_i$]
			\label{prop:distribution c_i}
			Let Assumptions~\ref{ass:growth-rate}~and~\ref{ass:loss} hold. Then, for $i\in\{1,\ldots,n\}$, 
			\begin{equation*}
			c_i-h(r_i)=o_P(1)
			\end{equation*}
			with $c_i$ and $r_i$ defined respectively in \eqref{eq:def-c} and \eqref{eq:def-r}. And
			\begin{align*}
			r_i\cd \mathcal{N}(m,\sigma^2)
			\end{align*}
			with $m$, $\sigma^2$ as given in \eqref{eq:m sigma}.
			
		\end{Proposition}

		A few remarks regarding the bias-variance trade-off and the role of regularization can be made. Recall that ${\rm Err}(\hat\bbeta)={\rm Err}(\alpha \hat\bbeta)$ for any $\alpha>0$, meaning that the classification error only depends on the direction of $\hat\bbeta$ and has nothing to do with its amplitude. Interested in the classification error, we say $\hat\bbeta$ is \emph{unbiased} if the expectation of its high dimensional equivalent $\tilde\bbeta$ (given in \eqref{eq:tilde beta}) aligns with the true parameter vector $\bbeta_*$ (defined in \eqref{eq:beta_star}), which yields the oracle classification error. We observe from \eqref{eq:tilde beta} that the unregularized solution $\hat\bbeta_\ell(\lambda = 0)$ is always unbiased, for all licit loss $\ell(\cdot)$ and ratio $n/p$, in line with the conclusion in \citep{rosasco2004loss} in the $n/p \to \infty$ limit. Interestingly, we remark that regularized solutions can also be unbiased, under the necessary and sufficient condition that $\bmu$ lies in the eigenspace of $\C$. In general, the directional bias of regularized classifiers is reflected by $\lambda\theta$, as can be easily seen from \eqref{eq:tilde beta}. Regarding the variance, note that, at a fixed directional bias (i.e., fixed $\lambda/\theta$), ${\rm Err}(\tilde\bbeta_\ell(\lambda))$ is a strictly decreasing function of $\gamma/\eta$ (as can be derived from Corollary~\ref{cor:classification error}), which controls the variance level of $\tilde\bbeta_\ell(\lambda)$. Notice from \eqref{eq:theta eta gamma} that
		\[
		    \frac{\gamma}{\eta}=\sqrt{\frac{p}{n}}\frac{\sqrt{\E[h^2(r)]}}{\E[h(r)]} \ge \sqrt{\frac{p}n},
		\]
		by Cauchy-Schwarz inequality, with equality if and only if $h(r)$ is a deterministic constant. Moreover, recall from Proposition~\ref{prop:distribution c_i} that $c_i\sim h(r)$ asymptotically, and that all $c_i \to \ell'(0)$ as $\lambda\to\infty$ (since $\Vert\hat\bbeta\Vert\to0$), we deduce that the ratio $\gamma/\eta$ reaches its minimum value in the \emph{strong regularization} $\lambda\to\infty$ limit. In summary, 
		\begin{enumerate}
			\item non-zero regularization (with $\lambda>0$) introduces a directional bias in $\E[\hat\bbeta]$ reflected by $\lambda/\theta$, except when $\bmu$ in the eigenspace of $\C$;
			\item at a certain bias level (i.e., with fixed $\lambda/\theta$), the classification error increases with $\gamma/\eta$, which serves as an indicator of the variance of $\hat\bbeta$ that is minimized at $\lambda\to\infty$.
		\end{enumerate}
		Therefore, we see that the bias-variance trade-off will come into play when introducing the regularization $\lambda > 0$. In general, the optimal $\lambda$ is finite and non-zero\footnote{An exception is when $\C=\I_p$. In this case, there is no directional bias for $\lambda>0$ as $\bmu$ is obviously always an eigenvector of $\I_p$, the optimal performance is therefore achieved at $\lambda\to\infty$ with a minimized variance. }, and it depends on the bias-variance trade-off in the classification performance.
		
		\subsection{Characterization of multiple classifiers}
		
		When multiple ERM classifiers are considered, we provide a description of the joint distribution of $\hat\bbeta$ obtained with different loss functions $\ell$ but on the same training samples, as formulated in the following theorem.
		
		\begin{Theorem}[Joint statistical behavior of multiple $\hat \bbeta$]
		\label{theo:joint distribution}
			Let Assumptions~\ref{ass:growth-rate}~and~\ref{ass:loss} hold. For any set of $\hat \bbeta_{\ell_1}(\lambda_1),\ldots, \hat \bbeta_{\ell_M}(\lambda_M)$ solutions to \eqref{eq:opt-origin}, let $\u_i$ denote the associated random vector $\u$ defined in \eqref{eq:tilde beta} for $\hat \bbeta_{\ell_i}(\lambda_i) $, $i=\{1,\ldots,M\}$, and $\c_i$ the associated vector $\c$ as defined in \eqref{eq:def-c}. We have that $\u_1,\ldots,\u_M\sim\mathcal{N}(\zeros_p,\I_p/p)$ are jointly Gaussian vectors with
			\begin{align*}
			\cov[\u_i,\u_j]=\rho_{ij}\I_p/p,\quad \rho_{ij}=\frac{\c_i^\T\c_j}{\Vert\c_i\Vert\Vert\c_j\Vert}+o_P(1), \quad i,j\in\{1,\ldots,M\}.
			\end{align*}
		\end{Theorem}
		
		An application of the above theorem is to establish the \emph{optimal linear combination} of a set of $\hat\bbeta$ obtained from possibly different loss functions $\ell$ to further minimize the classification error (with respect to that of a single $\ell$). For instance, with $\lambda=0$, the high  dimensional equivalent $\tilde\bbeta_\ell$ of $\hat\bbeta_\ell$ is aligned in expectation with the oracle solution $\bbeta_*$, for any $\ell$. According to Theorem~\ref{theo:main}, we have that, for any set of $\hat \bbeta_{\ell_1}(0),\ldots, \hat \bbeta_{\ell_M}(0)$ solutions to \eqref{eq:opt-origin} on the \emph{same} training set $\{(\x_i, y_i)\}_{i=1}^n$ and any $a_1,\ldots,a_M\in\RR$, 
		$$\left\Vert \sum_{i=1}^M a_i\hat \bbeta_{\ell_i}(0)-\sum_{i=1}^M \frac{a_i\eta_i}{\theta_i}\C^{-1}\bmu+\C^{-\frac{1}{2}}\sum_{i=1}^M \frac{a_i\gamma_i}{\theta_i}\u_i\right\Vert=o_P(1)$$
		with $\theta_i, \eta_i, \gamma_i$ and  $\u_i$ associated with $\bbeta_{\ell_i}(0)$ as defined in Theorem~\ref{theo:main}. Plugging the estimators \eqref{eq:hat theta eta gamma} of $\hat\theta,\hat\eta,\hat\gamma$ into the above equation and handling the linear combination $\sum_{i=1}^M \frac{a_i\gamma_i}{\theta_i}\u_i$ with the results of Theorem~\ref{theo:joint distribution}, we have the below result on the linear combination of \emph{unregularized} classifiers. 
		\begin{Corollary}[Linear combination of unregularized classifiers]
		\label{cor:performance of combination unreg}
			Under the assumption and notations of Theorem~\ref{theo:joint distribution}, we have that, for any set of $\hat \bbeta_{\ell_1}(0),\ldots, \hat \bbeta_{\ell_M}(0)$ unregularized solutions to \eqref{eq:opt-origin} and any $a_1,\ldots,a_M\in\RR$,
			$$\left\Vert \sum_{i=1}^M a_i\hat \bbeta_{\ell_i}(0)-\bxi\right\Vert=o_P(1),\quad \bxi = \tau\C^{-1}\bmu+\omega\C^{-\frac{1}{2}}\u'$$
			where $\u'\sim\mathcal{N}(\zeros_p,\I_p/p)$ and
			\begin{align}
			\tau=\frac{\ones_n^\T\left(\sum_{i=1}^M \frac{a_i}{\hat\theta_i}\c_i\right)}{n}+o_P(1)\nonumber\\
			\omega=\frac{\sqrt{p}\left\Vert\sum_{i=1}^M \frac{a_i}{\hat\theta_i}\c_i\right\Vert}{n}+o_P(1)\label{eq:tau omega}
			\end{align}
			with $\hat\theta_i$ given by \eqref{eq:hat theta eta gamma} with respect to $\bbeta_{\ell_i}(0)$.
			
			Consequently,
			\begin{align*}
			{\rm Err}\left(\sum_{i=1}^M a_i\hat \bbeta_{\ell_i}(0)\right)\geq {\rm Err}\left(\sum_{i=1}^M a_i^{{\rm opt}}\hat \bbeta_{\ell_i}(0)\right)
			\end{align*}
			with $a_1^{{\rm opt}},\ldots,a_M^{{\rm opt}}$ the solution to
			\begin{align}
			\label{eq:opt linear comb}
			\min_{a_1,\ldots,a_M\in\RR}\frac{\sqrt{p}\left\Vert\sum_{i=1}^M \frac{a_i}{\hat\theta_i}\c_i\right\Vert}{\ones_n^\T\left(\sum_{i=1}^M \frac{a_i}{\hat\theta_i}\c_i\right)},\quad\textit{s.t. } \ones_n^\T\left(\sum_{i=1}^M \frac{a_i}{\hat\theta_i}\c_i\right)>0.
			\end{align}
		\end{Corollary}
		
		For a given set of unregularized classifiers $\{\hat \bbeta_{\ell_i}(0)\}_{i=1}^M$, the optimal values of $\{a_i\}_{i=1}^M$ can thus be determined via \eqref{eq:opt linear comb}. Since $\hat\theta_i,\c_i$ involved in \eqref{eq:opt linear comb} are empirically accessible from the training samples $\{\x_i,y_i\}_{i=1}^n$, Corollary~\ref{cor:performance of combination unreg} therefore provides a \emph{cross-validation-free} method to derive the theoretically optimal weights. Evidently, the same reasoning can be made upon regularized classifiers. It requires however more computational effort to implement in practice: first we need to find the values of regularization $\lambda_i$ such that all $\E[\hat \bbeta_{\ell_i}(\lambda_i)]$, $i\in\{1,\ldots,M\}$, are on the same direction (i.e., $\hat \bbeta_{\ell_i}(\lambda_i)$ have the same directional bias with respect to the oracle $\bbeta_*$), then use the same method to estimate the optimal weights. This is described in details as follows. 
		
		\begin{Corollary}[Linear combination of regularized classifiers]\label{cor:optimal combination reg}
			Under the assumption and notations of Theorem~\ref{theo:joint distribution}, we have that, for any set of $\hat \bbeta_{\ell_1}(\lambda_1),\ldots, \hat \bbeta_{\ell_M}(\lambda_M)$ solutions to \eqref{eq:opt-origin} such that all $\lambda_i/\hat\theta_i$, $i\in\{1,\ldots,M\}$, are equal, with $\hat\theta_i$ given by \eqref{eq:hat theta eta gamma} for $\bbeta_{\ell_i}(\lambda_i)$, and any $a_1,\ldots,a_M\in\RR$,
			$$\left\Vert \sum_{i=1}^M a_i\hat \bbeta_{\ell_i}(\lambda_i)-\bxi\right\Vert=o_P(1),\quad \bxi = \left(\frac{\lambda_1}{\hat\theta_1}\I_p+\C\right)^{-1}\left(\tau\bmu+\omega\C^{\frac{1}{2}}\u'\right)$$
			where $\u'\sim\mathcal{N}(\zeros_p,\frac{1}{p}\I_p)$ and $\tau,\omega$ as characterized in \eqref{eq:tau omega} with $\hat\theta_i,\c_i$ associated with $\bbeta_{\ell_i}(\lambda_i)$.
			
			Consequently,
			\begin{align*}
			{\rm Err}\left(\sum_{i=1}^M a_i\hat \bbeta_{\ell_i}(\lambda_i)\right)\geq {\rm Err}\left(\sum_{i=1}^M a_i^{{\rm opt}}\hat \bbeta_{\ell_i}(\lambda_i)\right)
			\end{align*}
			with $a_1^{{\rm opt}},\ldots,a_m^{{\rm opt}}$ the solution to \eqref{eq:opt linear comb} by letting $\hat\theta_i,\c_i$ be computed for $\bbeta_{\ell_i}(\lambda_i)$.
		\end{Corollary}

   \section{Numerical Validation}
   \label{sec:numerical validation}
   
   The objective of this section is to empirically validate our theoretical results, which are shown to be remarkably accurate even for moderately large data sets of $n,p$ around hundreds.
   
   With the high dimensional distribution of $\hat\bbeta$ characterized in Theorem~\ref{theo:main}, we have access to the generalization error of classification as presented in Corollary~\ref{cor:classification error}. As can be seen in Figure~\ref{fig:performance-compare-deterministic}, where the actual performance obtained by running the learning algorithm is compared against the theoretical value from Corollary~\ref{cor:classification error},  our high dimensional prediction is almost exact for data vectors of dimension $p=300$.  
   
   \begin{figure}
	\centering
		\begin{tabular}{cc}
			\begin{tikzpicture}[font=\normalsize]
			\renewcommand{\axisdefaulttryminticks}{4} 
			\pgfplotsset{every major grid/.style={densely dashed}}       
			\tikzstyle{every axis y label}+=[yshift=-10pt] 
			\tikzstyle{every axis x label}+=[yshift=5pt]
			\pgfplotsset{every axis legend/.append style={cells={anchor=west},fill=white, at={(1,1)}, anchor=north east, font=\normalsize }}
   			\begin{axis}[
   		    width=.48\linewidth,
	     	height=.35\linewidth,
   			ymin=0.098,
   			ymax=0.122,
   			xmode=log,
   			log basis x ={2},
   			grid=major,
   			ymajorgrids=false,
   			scaled ticks=true,
   			xtick={0.0312,0.25,2,16,128,1024},
   			ytick={0.12,0.11,0.1},
   			yticklabels = {$12$,$11$,$10$},
   			xlabel={$\lambda$},
   			ylabel={ Classification error($\%$)},
   			]
   		  \addplot[only marks,mark=x,mark options={draw=RED,fill=white,mark size=4pt},RED,line width=1.5pt] plot coordinates{
   		(0.01562,0.1182)(0.03125,0.1117)(0.0625,0.1063)(0.125,0.1028)(0.25,0.1017)(0.5,0.1027)(1,0.1052)(2,0.1081)(4,0.1106)(16,0.1136)(64,0.1145)(256,0.1147)(1024,0.1148)
   		};
   			\addlegendentry{{ Empirical }};
   			\addplot[smooth,BLUE,line width=1.5pt] plot coordinates{
 (0.01562,0.1183)(0.03125,0.1118)(0.0625,0.1064)(0.125,0.1030)(0.25,0.1019)(0.5,0.1030)(1,0.1053)(2,0.1083)(4,0.1108)(8,0.1127)(16,0.1137)(32,0.1143)(64,0.1146)(128,0.1149)(256,0.1150)(512,0.1150)(1024,0.1150)
   			};
   			\addlegendentry{{Prediction}};
   			\end{axis}
			\end{tikzpicture}

&
			\begin{tikzpicture}[font=\normalsize]
			\renewcommand{\axisdefaulttryminticks}{4} 
			\pgfplotsset{every major grid/.style={densely dashed}}       
			\tikzstyle{every axis y label}+=[yshift=-10pt] 
			\tikzstyle{every axis x label}+=[yshift=5pt]
			\pgfplotsset{every axis legend/.append style={cells={anchor=west},fill=white, at={(1,1)}, anchor=north east, font=\normalsize }}
 			\begin{axis}[
 		    width=.48\linewidth,
	     	height=.35\linewidth,
   			xmin=750,
   			ymin=0.09,
   			xmax=3150,
   			ymax=0.15,
   			xtick={1200,1800,2400,3000},
   			ytick={0.1,0.12,0.14},
   			yticklabels = {$10$,$12$,$14$},
   			grid=major,
   			ymajorgrids=false,
   			scaled ticks=true,
   			xlabel={$n$},
   			]
   			\addplot[only marks,mark=x,mark options={draw=RED,fill=white,mark size=4pt},RED,line width=1.5pt] plot coordinates{
   			(900,0.1424)(1200,0.1241)(1500,0.1139)(1800,0.1074)(2100,0.1030)(2400,0.0997)(2700,0.0972)(3000,0.0952)
   			};
   			\addlegendentry{{ Empirical }};
   			\addplot[smooth,BLUE,line width=1.5pt] plot coordinates{
   			(900,0.1424)(1200,0.1240)(1500,0.1139)(1800,0.1074)(2100,0.1030)(2400,0.0997)(2700,0.0972)(3000,0.0952)
   			};
   			\addlegendentry{{Prediction}};
   			\end{axis}
			\end{tikzpicture}
	\end{tabular}
   		\caption{Comparison between the empirical classification error and its high dimensional prediction given in Corollary~\ref{cor:classification error}, for data vectors with $p=300$, $\bmu=\sqrt 2\cdot[\ones_{p/2}; 2\ones_{p/2}]/\sqrt p$ and $\C =\I_p+ 6\cdot[\zeros_{p/2};~\sqrt{2}\ones_{p/2}][\zeros_{p/2};~\ones_{p/2}]^\T/ p$, \textbf{(left)} as the function of the regularization penalty $\lambda$ for $n=900$ and logistic loss $\ell(t) = \ln(1+e^{-t})$, or \textbf{(right)} as a function of the data number $n$ for $\lambda=0$ and square loss $\ell(t) = (t-1)^2/2$. Empirical results are obtained by averaging over $500$ independent realizations. }
   		\label{fig:performance-compare-deterministic}
\end{figure} 

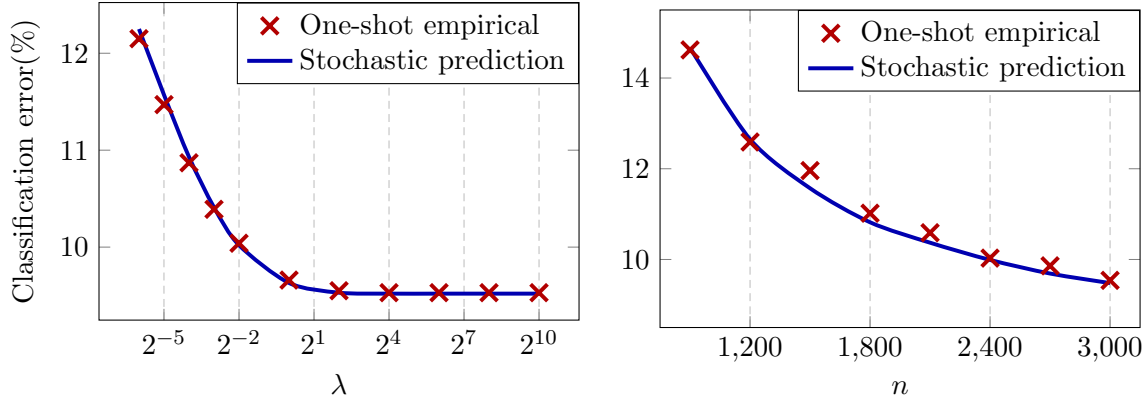
\begin{figure}
	\centering
		\begin{tabular}{cc}
			\begin{tikzpicture}[font=\normalsize]
			\renewcommand{\axisdefaulttryminticks}{4} 
			\pgfplotsset{every major grid/.style={densely dashed}}       
			\tikzstyle{every axis y label}+=[yshift=-10pt] 
			\tikzstyle{every axis x label}+=[yshift=5pt]
			\pgfplotsset{every axis legend/.append style={cells={anchor=west},fill=white, at={(1,1)}, anchor=north east, font=\normalsize }}
			\begin{axis}[
			width=.48\linewidth,
	     	height=.35\linewidth,
			 xmode=log,
			log basis x ={2},
			grid=major,
			ymajorgrids=false,
			scaled ticks=true,
			xtick={0.0312,0.25,2,16,128,1024},
			ytick={0.12,0.11,0.1},
			yticklabels = {$12$,$11$,$10$},
			xlabel={$\lambda$},
			ylabel={  Classification error($\%$) },
			]
			\addplot[only marks,mark=x,mark options={draw=RED,fill=white,mark size=4.5pt},RED,line width=1.5pt] plot coordinates{
			(0.01562,0.1215)(0.03125,0.1147)(0.0625,0.1087)(0.125,0.1039)(0.25,0.1004)(1,0.0966)(4,0.0955)(16,0.0953)(64,0.0953)(256,0.0953)(1024,0.0953)
			};
			\addlegendentry{{ One-shot empirical }};
        \addplot[smooth,BLUE,line width=1.5pt] plot coordinates{
			(0.01562,0.1225)(0.03125,0.1157)(0.0625,0.1093)(0.125,0.1041)(0.25,0.1002)(1,0.0963)(4,0.0953)(16,0.0952)(64,0.0952)(256,0.0952)(1024,0.0952)
			};
			\addlegendentry{{Stochastic prediction }};
			\end{axis}
			\end{tikzpicture}

&
			\begin{tikzpicture}[font=\normalsize]
			\renewcommand{\axisdefaulttryminticks}{4} 
			\pgfplotsset{every major grid/.style={densely dashed}}       
			\tikzstyle{every axis y label}+=[yshift=-10pt] 
			\tikzstyle{every axis x label}+=[yshift=5pt]
			\pgfplotsset{every axis legend/.append style={cells={anchor=west},fill=white, at={(1,1)}, anchor=north east, font=\normalsize }}
			\begin{axis}[
			width=.48\linewidth,
	     	height=.35\linewidth,
			xmin=750,
			ymin=0.085,
			xmax=3150,
			ymax=0.155,
			xtick={1200,1800,2400,3000},
			ytick={0.14,0.12,0.1},
            yticklabels = {$14$,$12$,$10$},
			grid=major,
			ymajorgrids=false,
			scaled ticks=true,
			xlabel={$n$},
			]
	       \addplot[only marks,mark=x,mark options={draw=RED,fill=white,mark size=4.5pt},RED,line width=1.5pt] plot coordinates{
	(900,0.1462)(1200,0.1259)(1500,0.1196)(1800,0.1102)(2100,0.1059)(2400,0.1003)(2700,0.0986)(3000,0.0954)
};
\addlegendentry{{ One-shot empirical }};
\addplot[smooth,BLUE,line width=1.5pt] plot coordinates{
	(900,0.1467)(1200,0.1265)(1500,0.1157)(1800,0.1082)(2100,0.1037)(2400,0.0999)(2700,0.0969)(3000,0.0948)
};
\addlegendentry{{Stochastic prediction }};
			\end{axis}
			\end{tikzpicture}
	\end{tabular}
		\caption{ Comparison between the \emph{one-shot} empirical classification error and its stochastic high dimensional prediction via \eqref{eq:Err asymtotic stochastic},  for data vectors with $p=300$, $\bmu=\sqrt{2/p}\cdot \ones_p$, $\C =\I_p$, \textbf{(left)} as the function of the regularization penalty $\lambda$ for $n=900$ and logistic loss $\ell(t) = \ln(1+e^{-t})$, or \textbf{(right)} as a function of the data number $n$ for $\lambda=0$ and square loss $\ell(t) = (t-1)^2/2$.}
		\label{fig:performance-compare-stochastic}
\end{figure}

As stated in Theorem~\ref{theo:main}, the three parameters $\theta,\eta,\gamma$ fully describing the high dimensional distribution of $\hat{\bbeta}$ can be accessed empirically from the training samples $\{(\x_i,y_i)\}_{i=1}^n$. Plugging the estimated values $\hat\theta,\hat\eta,\hat\gamma$ given in \eqref{eq:hat theta eta gamma} into the expression \eqref{eq:Err asymtotic} of classification error, we get 
\begin{equation}
\label{eq:Err asymtotic stochastic}
	{\rm Err}(\hat\bbeta_\ell(\lambda))= Q\left(\frac{\hat m}{\hat\sigma}\right)+o_P(1)
\end{equation}
where
\begin{align*}
\hat m = &\hat\eta\bmu^\T\left(\lambda\I_p+\hat\theta\C\right)^{-1}\bmu\\
\hat\sigma^2=&\hat\eta^2\bmu^\T\left(\lambda\I_p+\hat\theta\C\right)^{-1}\C\left(\lambda\I_p+\hat\theta\C\right)^{-1}\bmu+\hat\gamma\tr\left[\left(\lambda\I_p+\hat\theta\C\right)^{-1}\C\right]^2.
\end{align*} The numerical results displayed in Figure~\ref{fig:performance-compare-stochastic} show that the stochastic prediction of classification error given above matches very well the corresponding empirical value.

\begin{figure}
	\centering
		\begin{tabular}{cc}
			\begin{tikzpicture}[font=\normalsize]
			\renewcommand{\axisdefaulttryminticks}{4} 
			\pgfplotsset{every major grid/.style={densely dashed}}       
			\tikzstyle{every axis y label}+=[yshift=-10pt] 
			\tikzstyle{every axis x label}+=[yshift=5pt]
			\pgfplotsset{every axis legend/.append style={cells={anchor=west},fill=white, at={(1,1)}, anchor=north east, font=\normalsize }}
			\begin{axis}[
			width=.48\linewidth,
	     	height=.35\linewidth,
			xmin=-7,
			ymin=0,
			xmax=9,
			ymax=0.25,
			yticklabels = {},
			bar width=3pt,
			grid=major,
			ymajorgrids=false,
			scaled ticks=true,
			]
			\addplot+[ybar,mark=none,color=white,fill=blue!60!white,area legend] coordinates{
				(-7.329411, 0.000000)(-6.981611, 0.001872)(-6.633811, 0.000000)(-6.286011, 0.000000)(-5.938210, 0.000000)(-5.590410, 0.003744)(-5.242610, 0.000000)(-4.894810, 0.005616)(-4.547010, 0.011231)(-4.199210, 0.016847)(-3.851410, 0.031822)(-3.503610, 0.022463)(-3.155810, 0.022463)(-2.808010, 0.048669)(-2.460209, 0.056157)(-2.112409, 0.065516)(-1.764609, 0.073004)(-1.416809, 0.087979)(-1.069009, 0.117929)(-0.721209, 0.136648)(-0.373409, 0.132904)(-0.025609, 0.177829)(0.322191, 0.142263)(0.669991, 0.160982)(1.017792, 0.196548)(1.365592, 0.187188)(1.713392, 0.181573)(2.061192, 0.136648)(2.408992, 0.172213)(2.756792, 0.144135)(3.104592, 0.104826)(3.452392, 0.091722)(3.800192, 0.084235)(4.147992, 0.056157)(4.495793, 0.052413)(4.843593, 0.043053)(5.191393, 0.037438)(5.539193, 0.016847)(5.886993, 0.016847)(6.234793, 0.009359)(6.582593, 0.011231)(6.930393, 0.007488)(7.278193, 0.003744)(7.625993, 0.003744)(7.973794, 0.000000)(8.321594, 0.000000)(8.669394, 0.001872)(9.017194, 0.000000)(9.364994, 0.000000)(9.712794, 0.000000)
			};
			\addlegendentry{{ Empirical histogram of $r_i$ }}
			\addplot[smooth,RED,line width=1.5pt] plot coordinates{
				(-7.329411, 0.000267)(-6.981611, 0.000442)(-6.633811, 0.000719)(-6.286011, 0.001144)(-5.938210, 0.001784)(-5.590410, 0.002724)(-5.242610, 0.004074)(-4.894810, 0.005968)(-4.547010, 0.008563)(-4.199210, 0.012033)(-3.851410, 0.016563)(-3.503610, 0.022328)(-3.155810, 0.029483)(-2.808010, 0.038129)(-2.460209, 0.048298)(-2.112409, 0.059922)(-1.764609, 0.072816)(-1.416809, 0.086666)(-1.069009, 0.101031)(-0.721209, 0.115355)(-0.373409, 0.129005)(-0.025609, 0.141304)(0.322191, 0.151596)(0.669991, 0.159295)(1.017792, 0.163945)(1.365592, 0.165264)(1.713392, 0.163170)(2.061192, 0.157791)(2.408992, 0.149454)(2.756792, 0.138649)(3.104592, 0.125982)(3.452392, 0.112119)(3.800192, 0.097731)(4.147992, 0.083439)(4.495793, 0.069774)(4.843593, 0.057147)(5.191393, 0.045843)(5.539193, 0.036020)(5.886993, 0.027720)(6.234793, 0.020894)(6.582593, 0.015425)(6.930393, 0.011154)(7.278193, 0.007900)(7.625993, 0.005480)(7.973794, 0.003723)(8.321594, 0.002478)(8.669394, 0.001615)(9.017194, 0.001031)(9.364994, 0.000645)(9.712794, 0.000395)
			};
			\addlegendentry{{ Theory: $r \sim \mathcal{N}(m,\sigma^2$) } }
			\end{axis}
			\end{tikzpicture}

&
			\begin{tikzpicture}[font=\normalsize]
			\renewcommand{\axisdefaulttryminticks}{4} 
			\pgfplotsset{every major grid/.style={densely dashed}}       
			\tikzstyle{every axis y label}+=[yshift=-10pt] 
			\tikzstyle{every axis x label}+=[yshift=5pt]
			\pgfplotsset{every axis legend/.append style={cells={anchor=west},fill=white, at={(1,1)}, anchor=north east, font=\normalsize }}
			\begin{axis}[
			width=.48\linewidth,
	     	height=.35\linewidth,
			xmin=0,
			ymin=0,
			xmax=1.2,
			ymax=8,
			yticklabels = {},
			bar width=3pt,
			grid=major,
			ymajorgrids=false,
			scaled ticks=true,
			]
			\addplot+[ybar,mark=none,color=white,fill=blue!60!white,area legend] coordinates{
				(0.000132, 5.605215)(0.022432, 3.532453)(0.044733, 3.094546)(0.067034, 2.598251)(0.089334, 2.043568)(0.111635, 2.247925)(0.133936, 1.605661)(0.156236, 1.605661)(0.178537, 1.547273)(0.200838, 1.196947)(0.223138, 1.313722)(0.245439, 0.992590)(0.267740, 1.021784)(0.290040, 1.109365)(0.312341, 0.875815)(0.334642, 1.138559)(0.356942, 0.846621)(0.379243, 0.817427)(0.401544, 0.846621)(0.423844, 0.934203)(0.446145, 0.583877)(0.468445, 0.788233)(0.490746, 0.437907)(0.513047, 0.846621)(0.535347, 0.496295)(0.557648, 0.496295)(0.579949, 0.583877)(0.602249, 0.525489)(0.624550, 0.496295)(0.646851, 0.175163)(0.669151, 0.350326)(0.691452, 0.613070)(0.713753, 0.467101)(0.736053, 0.321132)(0.758354, 0.262744)(0.780655, 0.408714)(0.802955, 0.321132)(0.825256, 0.175163)(0.847557, 0.175163)(0.869857, 0.321132)(0.892158, 0.408714)(0.914459, 0.321132)(0.936759, 0.175163)(0.959060, 0.087581)(0.981361, 0.029194)(1.003661, 0.000000)(1.025962, 0.000000)(1.048263, 0.000000)(1.070563, 0.000000)(1.092864, 0.000000)
			};
			\addlegendentry{{ Empirical histogram of $c_i$ }}
			\addplot[smooth,RED,line width=1.5pt] plot coordinates{
				(0.000132, 7.026698)(0.022432, 4.536487)(0.044733, 3.562226)(0.067034, 2.524589)(0.089334, 2.082450)(0.111635, 1.803534)(0.133936, 1.595468)(0.156236, 1.472602)(0.178537, 1.340767)(0.200838, 1.225973)(0.223138, 1.146154)(0.245439, 1.097725)(0.267740, 1.017907)(0.290040, 0.967684)(0.312341, 0.933605)(0.334642, 0.896834)(0.356942, 0.858271)(0.379243, 0.815222)(0.401544, 0.791008)(0.423844, 0.753341)(0.446145, 0.730920)(0.468445, 0.698634)(0.490746, 0.673523)(0.513047, 0.625094)(0.535347, 0.601776)(0.557648, 0.582046)(0.579949, 0.544378)(0.602249, 0.514783)(0.624550, 0.497743)(0.646851, 0.466354)(0.669151, 0.419719)(0.691452, 0.403575)(0.713753, 0.393710)(0.736053, 0.359631)(0.758354, 0.354250)(0.780655, 0.354250)(0.802955, 0.340797)(0.825256, 0.316583)(0.847557, 0.301336)(0.869857, 0.280709)(0.892158, 0.246629)(0.914459, 0.218828)(0.936759, 0.193716)(0.959060, 0.147081)(0.981361, 0.096858)(1.003661, 0.058294)(1.025962, 0.026008)(1.048263, 0.000000)(1.070563, 0.000000)(1.092864, 0.000000)
			};
			\addlegendentry{{ Theory: $h(r)$ } }
			\end{axis}
			\end{tikzpicture}
	\end{tabular}
		\caption{ Comparison between the empirical histogram of $ r_i$ and $c_i$ with their theoretical prediction given in Proposition~\ref{prop:distribution c_i}. For logistic loss $\ell(t) = \ln(1+e^{-t})$, $\bmu = [1;~\mathbf{0}_{p-1}]$, $\C =2 \cdot \I_p$ and $p=256$, $n = 6p$.}
		\label{fig:distribution-c-r}
\end{figure}
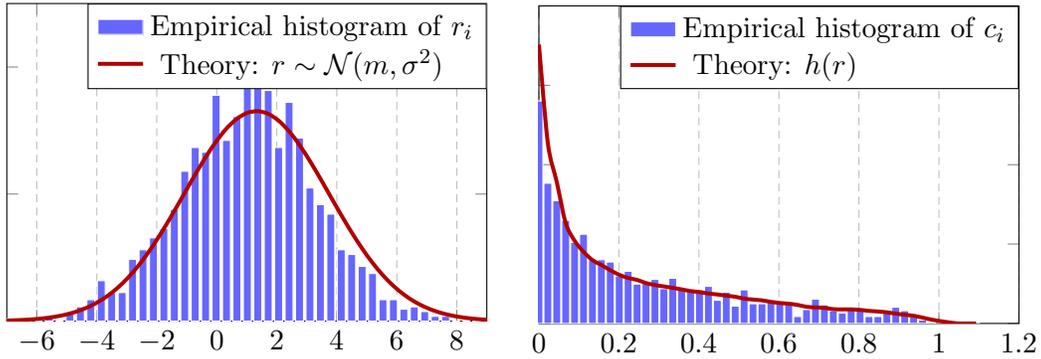

As discussed in Theorem~\ref{theo:main} and Proposition~\ref{prop:distribution c_i}, the deterministic and stochastic description of $\hat \bbeta$ are linked by the fact that $c_i$ has asymptotically the same distribution as $h(r)$ and $r_i$ as $r\sim\mathcal{N}(m,\sigma^2)$. We provide in Figure~\ref{fig:distribution-c-r} numerical confirmation of this result.

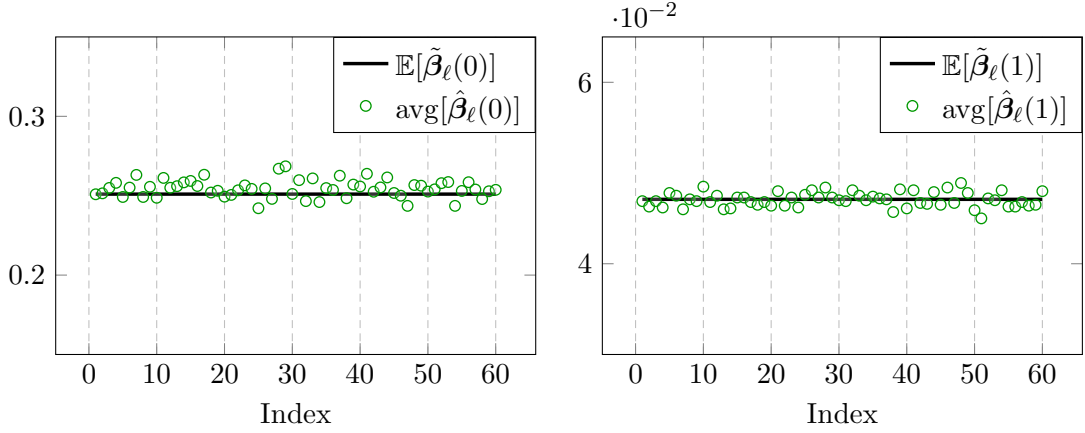
\begin{figure}
	\centering
		\begin{tabular}{cc}
			\begin{tikzpicture}[font=\normalsize]
			\renewcommand{\axisdefaulttryminticks}{4} 
			\pgfplotsset{every major grid/.style={densely dashed}}       
			\tikzstyle{every axis y label}+=[yshift=-10pt] 
			\tikzstyle{every axis x label}+=[yshift=5pt]
			\pgfplotsset{every axis legend/.append style={cells={anchor=west},fill=white, at={(1,1)}, anchor=north east, font=\normalsize }}
	   	    \begin{axis}[
			width=.48\linewidth,
	     	height=.35\linewidth,
			ymin=0.15,
			ymax=0.35,
			ytick={0.2,0.3},
			grid=major,
			ymajorgrids=false,
			scaled ticks=true,
			xlabel={Index},
			]
			\addplot[smooth,black,line width=1.3pt] coordinates{
				(1,0.2510)(2,0.2510)(3,0.2510)(4,0.2510)(5,0.2510)(6,0.2510)(7,0.2510)(8,0.2510)(9,0.2510)(10,0.2510)(11,0.2510)(12,0.2510)(13,0.2510)(14,0.2510)(15,0.2510)(16,0.2510)(17,0.2510)(18,0.2510)(19,0.2510)(20,0.2510)(21,0.2510)(22,0.2510)(23,0.2510)(24,0.2510)(25,0.2510)(26,0.2510)(27,0.2510)(28,0.2510)(29,0.2510)(30,0.2510)(31,0.2510)(32,0.2510)(33,0.2510)(34,0.2510)(35,0.2510)(36,0.2510)(37,0.2510)(38,0.2510)(39,0.2510)(40,0.2510)(41,0.2510)(42,0.2510)(43,0.2510)(44,0.2510)(45,0.2510)(46,0.2510)(47,0.2510)(48,0.2510)(49,0.2510)(50,0.2510)(51,0.2510)(52,0.2510)(53,0.2510)(54,0.2510)(55,0.2510)(56,0.2510)(57,0.2510)(58,0.2510)(59,0.2510)(60,0.2510)
			};
			\addlegendentry{{$\E[\tilde\bbeta_{\ell}(0)]$ }};
			\addplot[mark =*,only marks,mark options={draw=GREEN,fill=white,fill opacity=0.2},mark size=2pt,line width =0.5pt] plot coordinates{
				(1,0.2509)(2,0.2515)(3,0.2549)(4,0.2581)(5,0.2492)(6,0.2551)(7,0.2631)(8,0.2492)(9,0.2556)(10,0.2487)(11,0.2612)(12,0.2550)(13,0.2560)(14,0.2584)(15,0.2593)(16,0.2561)(17,0.2632)(18,0.2520)(19,0.2531)(20,0.2494)(21,0.2505)(22,0.2534)(23,0.2565)(24,0.2543)(25,0.2421)(26,0.2546)(27,0.2480)(28,0.2670)(29,0.2685)(30,0.2511)(31,0.2598)(32,0.2466)(33,0.2609)(34,0.2460)(35,0.2548)(36,0.2536)(37,0.2626)(38,0.2484)(39,0.2570)(40,0.2558)(41,0.2638)(42,0.2525)(43,0.2552)(44,0.2615)(45,0.2517)(46,0.2500)(47,0.2436)(48,0.2567)(49,0.2563)(50,0.2526)(51,0.2540)(52,0.2579)(53,0.2586)(54,0.2436)(55,0.2530)(56,0.2585)(57,0.2539)(58,0.2479)(59,0.2528)(60,0.2537)
			};
			\addlegendentry{{${\rm avg}[\hat\bbeta_{\ell}(0)]$ }};
			
			\end{axis}
			\end{tikzpicture}
&
			\begin{tikzpicture}[font=\normalsize]
			\renewcommand{\axisdefaulttryminticks}{4} 
			\pgfplotsset{every major grid/.style={densely dashed}}       
			\tikzstyle{every axis y label}+=[yshift=-10pt] 
			\tikzstyle{every axis x label}+=[yshift=5pt]
			\pgfplotsset{every axis legend/.append style={cells={anchor=west},fill=white, at={(1,1)}, anchor=north east, font=\normalsize }}
            \begin{axis}[
			width=.48\linewidth,
	     	height=.35\linewidth,
			ymin=0.03,
			ymax=0.065,
			ytick={0.04,0.06},
			grid=major,
			ymajorgrids=false,
			scaled ticks=true,
			xlabel={Index},
			]
			\addplot[smooth,black,line width=1.3pt] coordinates{
				(1,0.0471)(2,0.0471)(3,0.0471)(4,0.0471)(5,0.0471)(6,0.0471)(7,0.0471)(8,0.0471)(9,0.0471)(10,0.0471)(11,0.0471)(12,0.0471)(13,0.0471)(14,0.0471)(15,0.0471)(16,0.0471)(17,0.0471)(18,0.0471)(19,0.0471)(20,0.0471)(21,0.0471)(22,0.0471)(23,0.0471)(24,0.0471)(25,0.0471)(26,0.0471)(27,0.0471)(28,0.0471)(29,0.0471)(30,0.0471)(31,0.0471)(32,0.0471)(33,0.0471)(34,0.0471)(35,0.0471)(36,0.0471)(37,0.0471)(38,0.0471)(39,0.0471)(40,0.0471)(41,0.0471)(42,0.0471)(43,0.0471)(44,0.0471)(45,0.0471)(46,0.0471)(47,0.0471)(48,0.0471)(49,0.0471)(50,0.0471)(51,0.0471)(52,0.0471)(53,0.0471)(54,0.0471)(55,0.0471)(56,0.0471)(57,0.0471)(58,0.0471)(59,0.0471)(60,0.0471)
			};
			\addlegendentry{{$\E[\tilde\bbeta_{\ell}(1)]$ }};
			\addplot[mark =*,only marks,mark options={draw=GREEN,fill=white,fill opacity=0.2},mark size=2pt,line width =0.5pt] plot coordinates{
				(1,0.0469)(2,0.0463)(3,0.0469)(4,0.0462)(5,0.0478)(6,0.0475)(7,0.0460)(8,0.0471)(9,0.0469)(10,0.0485)(11,0.0468)(12,0.0475)(13,0.0460)(14,0.0461)(15,0.0473)(16,0.0473)(17,0.0468)(18,0.0465)(19,0.0468)(20,0.0464)(21,0.0480)(22,0.0464)(23,0.0473)(24,0.0462)(25,0.0476)(26,0.0481)(27,0.0473)(28,0.0484)(29,0.0473)(30,0.0470)(31,0.0469)(32,0.0481)(33,0.0475)(34,0.0470)(35,0.0474)(36,0.0472)(37,0.0471)(38,0.0457)(39,0.0482)(40,0.0461)(41,0.0481)(42,0.0467)(43,0.0466)(44,0.0479)(45,0.0465)(46,0.0484)(47,0.0467)(48,0.0489)(49,0.0478)(50,0.0459)(51,0.0450)(52,0.0472)(53,0.0470)(54,0.0481)(55,0.0463)(56,0.0463)(57,0.0468)(58,0.0464)(59,0.0465)(60,0.0480)
				
			};
			\addlegendentry{{${\rm avg}[\hat\bbeta_{\ell}(1)]$ }};
			\end{axis}
			\end{tikzpicture}
	\end{tabular}
		\caption{Comparison between the empirical average ${\rm avg}[\hat\bbeta]$ of $\hat\bbeta$ over $1\,000$ independent realizations and its high dimensional expectation $\E[\tilde\bbeta]$, for data vectors of $p=60$, $\bmu=\sqrt 2\cdot \ones_p/\sqrt p$, $\C =2\cdot\I_p$, and with $\ell(t) = \ln(1+e^{-t})$. \textbf{Left}: unregularized solution with $\lambda=0$. \textbf{Right}: regularized solution with $\lambda=1$.}
		\label{fig:beta identity matrix}
\end{figure}

It is interesting to observe from the left plot of Figure~\ref{fig:performance-compare-deterministic} and Figure~\ref{fig:performance-compare-stochastic} that  $\lambda$ is optimized at a finite value under the setting of Figure~\ref{fig:performance-compare-deterministic}, but in the limit of $\lambda \to \infty$ in the case of Figure~\ref{fig:performance-compare-stochastic}. The reason is that when $\C=a\I_p$ as in Figure~\ref{fig:performance-compare-stochastic}, we reduce the variance and in the same time introduce no directional bias by increasing $\lambda$ (as recalled from the discussion below Proposition~\ref{prop:distribution c_i}), the classification is thus minimized at extremely large $\lambda$. To further illustrate this phenomenon, we plot in Figure~\ref{fig:beta}~and~\ref{fig:beta identity matrix} the high dimensional expectation and the empirical average of $\hat\bbeta$ for data vectors of $\C\neq\I_p$ (Figure~\ref{fig:beta}) and $\C=\I_p$ (Figure~\ref{fig:beta identity matrix}). In the both settings of Figure~\ref{fig:beta}~and~\ref{fig:beta identity matrix}, the oracle solution $\bbeta_*=2\C^{-1}\bmu$ is aligned with $\ones_p$, and so are the expectation of unregularized solutions, as displayed in the left side of the figures. And we can see from the right side that while the expectation of regularized solution deviates from the direction of $\ones_p$ in Figure~\ref{fig:beta} where $\C\neq\I_p$.  it remains aligned with $\ones_p$ in Figure~\ref{fig:beta identity matrix}  where $\C=\I_p$.

\begin{figure}
	\centering
		\begin{tabular}{cc}
			\begin{tikzpicture}[font=\normalsize]
			\renewcommand{\axisdefaulttryminticks}{4} 
			\pgfplotsset{every major grid/.style={densely dashed}}       
			\tikzstyle{every axis y label}+=[yshift=-10pt] 
			\tikzstyle{every axis x label}+=[yshift=5pt]
			\pgfplotsset{every axis legend/.append style={cells={anchor=west},fill=white, at={(1,1)}, anchor=north east, font=\normalsize }}
	   		\begin{axis}[
	   		width=.48\linewidth,
	     	height=.35\linewidth,
	ymin=0.10,
	ymax=0.24,
   ytick={0.1,0.2},
	grid=major,
	ymajorgrids=false,
	scaled ticks=true,
	xlabel={Index},
	]
	\addplot[smooth,black,line width=1.3pt] coordinates{
	(1,0.1657)(2,0.1657)(3,0.1657)(4,0.1657)(5,0.1657)(6,0.1657)(7,0.1657)(8,0.1657)(9,0.1657)(10,0.1657)(11,0.1657)(12,0.1657)(13,0.1657)(14,0.1657)(15,0.1657)(16,0.1657)(17,0.1657)(18,0.1657)(19,0.1657)(20,0.1657)(21,0.1657)(22,0.1657)(23,0.1657)(24,0.1657)(25,0.1657)(26,0.1657)(27,0.1657)(28,0.1657)(29,0.1657)(30,0.1657)(31,0.1657)(32,0.1657)(33,0.1657)(34,0.1657)(35,0.1657)(36,0.1657)(37,0.1657)(38,0.1657)(39,0.1657)(40,0.1657)(41,0.1657)(42,0.1657)(43,0.1657)(44,0.1657)(45,0.1657)(46,0.1657)(47,0.1657)(48,0.1657)(49,0.1657)(50,0.1657)(51,0.1657)(52,0.1657)(53,0.1657)(54,0.1657)(55,0.1657)(56,0.1657)(57,0.1657)(58,0.1657)(59,0.1657)(60,0.1657)
	};
	\addlegendentry{{$\E[\tilde\bbeta_{\ell}(0)]$ }};
	\addplot[mark =*,only marks,mark options={draw=GREEN,fill=white,fill opacity=0.2},mark size=2pt,line width =0.5pt] plot coordinates{
		(1,0.1701)(2,0.1727)(3,0.1669)(4,0.1688)(5,0.1678)(6,0.1707)(7,0.1729)(8,0.1811)(9,0.1736)(10,0.1712)(11,0.1736)(12,0.1708)(13,0.1794)(14,0.1708)(15,0.1629)(16,0.1669)(17,0.1694)(18,0.1724)(19,0.1654)(20,0.1655)(21,0.1761)(22,0.1661)(23,0.1679)(24,0.1657)(25,0.1722)(26,0.1725)(27,0.1746)(28,0.1629)(29,0.1670)(30,0.1745)(31,0.1722)(32,0.1688)(33,0.1707)(34,0.1658)(35,0.1689)(36,0.1690)(37,0.1690)(38,0.1718)(39,0.1694)(40,0.1766)(41,0.1686)(42,0.1722)(43,0.1669)(44,0.1732)(45,0.1721)(46,0.1730)(47,0.1692)(48,0.1697)(49,0.1791)(50,0.1650)(51,0.1662)(52,0.1785)(53,0.1723)(54,0.1754)(55,0.1687)(56,0.1712)(57,0.1678)(58,0.1672)(59,0.1626)(60,0.1726)
	};
\addlegendentry{{${\rm avg}[\hat\bbeta_{\ell}(0)]$ }};
	\end{axis}
			\end{tikzpicture}
&
			\begin{tikzpicture}[font=\normalsize]
			\renewcommand{\axisdefaulttryminticks}{4} 
			\pgfplotsset{every major grid/.style={densely dashed}}       
			\tikzstyle{every axis y label}+=[yshift=-10pt] 
			\tikzstyle{every axis x label}+=[yshift=5pt]
			\pgfplotsset{every axis legend/.append style={cells={anchor=west},fill=white, at={(1,1)}, anchor=north east, font=\normalsize }}
		   	\begin{axis}[
		   	width=.48\linewidth,
	     	height=.35\linewidth,
		ymin=0.03,
		ymax=0.07,
	 ytick={0.04,0.06},
		grid=major,
		ymajorgrids=false,
		scaled ticks=true,
		xlabel={Index},
		]
	\addplot[smooth,black,line width=1.3pt] coordinates{
		(1,0.0372)(2,0.0372)(3,0.0372)(4,0.0372)(5,0.0372)(6,0.0372)(7,0.0372)(8,0.0372)(9,0.0372)(10,0.0372)(11,0.0372)(12,0.0372)(13,0.0372)(14,0.0372)(15,0.0372)(16,0.0372)(17,0.0372)(18,0.0372)(19,0.0372)(20,0.0372)(21,0.0372)(22,0.0372)(23,0.0372)(24,0.0372)(25,0.0372)(26,0.0372)(27,0.0372)(28,0.0372)(29,0.0372)(30,0.0372)
};
\addlegendentry{{$\E[\tilde\bbeta_{\ell}(1)]$ }};
\addplot[mark =*,only marks,mark options={draw=GREEN,fill=white,fill opacity=0.2},mark size=2pt,line width =0.5pt] plot coordinates{
(1,0.0367)(2,0.0366)(3,0.0362)(4,0.0378)(5,0.0368)(6,0.0377)(7,0.0361)(8,0.0367)(9,0.0383)(10,0.0377)(11,0.0367)(12,0.0364)(13,0.0374)(14,0.0369)(15,0.0380)(16,0.0360)(17,0.0362)(18,0.0378)(19,0.0373)(20,0.0367)(21,0.0386)(22,0.0367)(23,0.0370)(24,0.0376)(25,0.0376)(26,0.0369)(27,0.0359)(28,0.0370)(29,0.0380)(30,0.0378)(31,0.0555)(32,0.0556)(33,0.0556)(34,0.0559)(35,0.0552)(36,0.0546)(37,0.0560)(38,0.0551)(39,0.0546)(40,0.0541)(41,0.0551)(42,0.0542)(43,0.0548)(44,0.0549)(45,0.0536)(46,0.0554)(47,0.0545)(48,0.0555)(49,0.0554)(50,0.0551)(51,0.0566)(52,0.0549)(53,0.0554)(54,0.0549)(55,0.0551)(56,0.0541)(57,0.0556)(58,0.0552)(59,0.0548)(60,0.0548)
};
\addlegendentry{{${\rm avg}[\hat\bbeta_{\ell}(1)]$ }};

	\addplot[smooth,black,line width=1.3pt] coordinates{
(31,0.0550)(32,0.0550)(33,0.0550)(34,0.0550)(35,0.0550)(36,0.0550)(37,0.0550)(38,0.0550)(39,0.0550)(40,0.0550)(41,0.0550)(42,0.0550)(43,0.0550)(44,0.0550)(45,0.0550)(46,0.0550)(47,0.0550)(48,0.0550)(49,0.0550)(50,0.0550)(51,0.0550)(52,0.0550)(53,0.0550)(54,0.0550)(55,0.0550)(56,0.0550)(57,0.0550)(58,0.0550)(59,0.0550)(60,0.0550)
};
		\end{axis}
			\end{tikzpicture}
	\end{tabular}
		\caption{Comparison between the empirical average ${\rm avg}[\hat\bbeta]$ of $\hat\bbeta$ over $1\,000$ independent realizations and its high dimensional expectation $\E[\tilde\bbeta]$, for data vectors of $p=60$, $\bmu=\sqrt2 \cdot [\ones_{p/2};~2 \cdot \ones_{p/2}]/\sqrt p$, $\C =3\cdot \I_p+ 6 \cdot [\zeros_{p/2};~\sqrt{2}\ones_{p/2}][\zeros_{p/2};~\ones_{p/2}]^\T/p$, and with logistic loss $\ell(t) = \ln(1+e^{-t})$. \textbf{Left}: unregularized solution with $\lambda=0$. \textbf{Right}: regularized solution with $\lambda=1$. }
		\label{fig:beta}
\end{figure} 
 
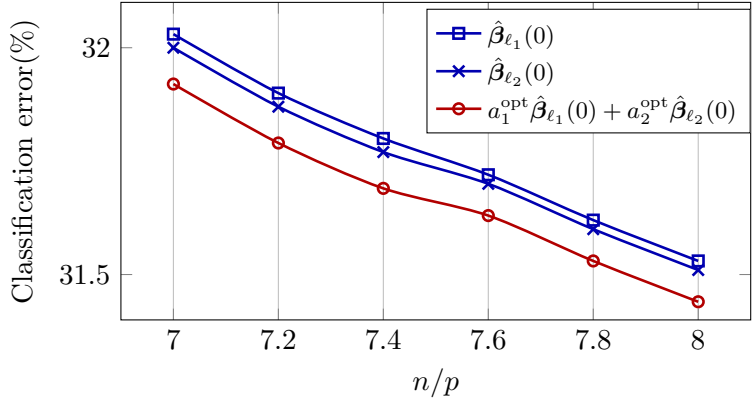
\begin{figure}
	\begin{center}
		\begin{tikzpicture}[font=\normalsize]
		\renewcommand{\axisdefaulttryminticks}{4} 
		\pgfplotsset{every axis legend/.append style={cells={anchor=west},fill=white, at={(1,0.98)}, anchor=north east, font=\footnotesize}}
		\pgfplotsset{/pgfplots/error bars/error bar style={thick}}
		\pgfplotsset{every axis plot/.append style={thick},}
		\begin{axis}[
		height=0.35\linewidth,
		width=0.6\linewidth,
		xmin=6.9,
		ymin=0.314,
		xmax=8.1,
		ymax=0.321,
		ytick={0.32,0.315},
		yticklabels = {$32$,$31.5$},
		grid=major,
		ymajorgrids=false,
		scaled ticks=true,
		xlabel={$n/p$},
		ylabel={Classification error($\%$) },
		]
		\addplot[smooth,mark=square,mark options={fill=white,mark size=2pt},BLUE,line width=1pt] plot coordinates{
			(7,0.3203)(7.2,0.3190)(7.4,0.3180)(7.6,0.3172)(7.8,0.3162)(8,0.3153)
		};
		\addlegendentry{{ $\hat\bbeta_{\ell_1}(0)$}};
		\addplot[smooth,mark=x,mark options={fill=white,mark size=3pt},BLUE,line width=1pt] plot coordinates{
			(7,0.3200)(7.2,0.3187)(7.4,0.3177)(7.6,0.3170)(7.8,0.3160)(8,0.3151)
		};
		\addlegendentry{{  $\hat\bbeta_{\ell_2}(0)$}};
		\addplot[smooth,mark=o,mark options={solid,draw=RED,fill=white,mark size=2pt},RED,line width=1pt] plot coordinates{
			(7,0.3192)(7.2,0.3179)(7.4,0.3169)(7.6,0.3163)(7.8,0.3153)(8,0.3144)
		};
		\addlegendentry{{$a_1^{\rm opt}\hat\bbeta_{\ell_1}(0)+a_2^{\rm opt}\hat\bbeta_{\ell_2}(0)$ }};
		\end{axis}
		\end{tikzpicture}
		\caption{Empirical classification errors given by the unregularized solutions $\hat\bbeta_{\ell_1}(0),\hat\bbeta_{\ell_2}(0)$ with $\ell_1(t) = \ln(1+e^{-t})$ and with the square root loss $\ell_1(t) = \sqrt{(t-1)^2+1} $, and the combined classifier $a_1^{\rm opt}\hat\bbeta_{\ell_1}(0)+a_2^{\rm opt}\hat\bbeta_{\ell_2}(0)$ with $a_1^{\rm opt},a_2^{\rm opt}$ given by \eqref{eq:opt linear comb} for $\bmu = [0.6;~\mathbf{0}_{p-1}]$, $\C = \I_p$ and $p = 250$.}
		\label{fig:improvement-ensemble}
	\end{center}
\end{figure}

With Theorem~\ref{theo:joint distribution} describing the joint distribution of a group of $\hat\bbeta$ obtained with different loss functions $\ell$, we can predict the performance of linear combined solutions of the form $\sum_{i=1}^M a_i\hat\bbeta_{\ell_i}$. In particular, when $\hat\bbeta_{\ell_i}$ have the same level of directional bias (for instance, when $\hat\bbeta_{\ell_i}$ are unregularized solutions with no directional bias),  we can find easily the optimal values of $a_i$ as stated in Corollary~\ref{cor:performance of combination unreg}--\ref{cor:optimal combination reg}. As validation of our results, a close match between the empirical performance and the theoretical prediction is observed in Figure~\ref{fig:ensemble-perf-compare} for combinations of unregularized solutions. The results in Figure~\ref{fig:improvement-ensemble} show that a non-negligible performance improvement can be achieved over individual unregularized classifiers $\hat\bbeta_{\ell_i}(0)$ by optimally combining them. However, as will be discussed in Section~\ref{sec:unregularized solutions}, this improvement is actually futile due to the fact that the unregularized solution of square loss can outperform any linearly combined unregularized solutions. We leave to future investigation the question  of whether this conclusion extends to regularized solutions.

\begin{figure}
	\centering
		\begin{tabular}{cc}
			\begin{tikzpicture}[font=\normalsize]
			\renewcommand{\axisdefaulttryminticks}{4} 
			\pgfplotsset{every major grid/.style={densely dashed}}       
			\tikzstyle{every axis y label}+=[yshift=-10pt] 
			\tikzstyle{every axis x label}+=[yshift=5pt]
			\pgfplotsset{every axis legend/.append style={cells={anchor=west},fill=white, at={(1,1)}, anchor=north east, font=\normalsize }}
			\begin{axis}[
            width=.48\linewidth,
	     	height=.35\linewidth,
			xmin=-4,
			ymin=0.27,
			xmax=4,
			ymax=0.33,
			grid=major,
			ymajorgrids=false,
			scaled ticks=true,
			ytick={0.33,0.31,0.29,0.27},
			yticklabels = {$33$,$31$,$29$,$27$},
			xlabel={$\rho$},
			ylabel={ Classification error($\%$) },
			]
						\addplot[only marks,mark=x,mark options={draw=RED,fill=white,mark size=4.5pt},RED,line width=1.5pt] plot coordinates{
				(-5.000000, 0.319900)(-4.000000, 0.301655)(-3.000000, 0.286985)(-2.000000, 0.277077)(-1.000000, 0.272209)(0.000000, 0.271778)(1.000000, 0.274679)(2.000000, 0.279737)(3.000000, 0.285979)(4.000000, 0.292704)(5.000000, 0.299465)
			};
			\addlegendentry{{ One-shot empirical }};
			\addplot[smooth,BLUE,line width=1.5pt] plot coordinates{
				(-5.000000, 0.303789)(-4.000000, 0.293736)(-3.000000, 0.284808)(-2.000000, 0.277747)(-1.000000, 0.273273)(0.000000, 0.271904)(1.000000, 0.273811)(2.000000, 0.278758)(3.000000, 0.286181)(4.000000, 0.295346)(5.000000, 0.305519)
			};
			\addlegendentry{{Stochastic prediction }};
			\end{axis}
			\end{tikzpicture}

&
			\begin{tikzpicture}[font=\normalsize]
			\renewcommand{\axisdefaulttryminticks}{4} 
			\pgfplotsset{every major grid/.style={densely dashed}}       
			\tikzstyle{every axis y label}+=[yshift=-10pt] 
			\tikzstyle{every axis x label}+=[yshift=5pt]
			\pgfplotsset{every axis legend/.append style={cells={anchor=west},fill=white, at={(1,1)}, anchor=north east, font=\normalsize }}
			\begin{axis}[
			width=.48\linewidth,
	     	height=.35\linewidth,
			xmin=-4,
			ymin=0.27,
			xmax=4,
			ymax=0.33,
			grid=major,
			ymajorgrids=false,
			scaled ticks=true,
			yticklabels = {},
			grid=major,
			ymajorgrids=false,
			scaled ticks=true,
			xlabel={$\rho$},
			]
			
			\addplot[only marks,mark=x,mark options={draw=RED,fill=white,mark size=4.5pt},RED,line width=1.5pt] plot coordinates{
				(-5.000000, 0.324650)(-4.000000, 0.316908)(-3.000000, 0.309793)(-2.000000, 0.303665)(-1.000000, 0.298890)(0.000000, 0.295786)(1.000000, 0.294562)(2.000000, 0.295283)(3.000000, 0.297849)(4.000000, 0.302026)(5.000000, 0.307490)
			};
			\addlegendentry{{ One-shot empirical }};
        \addplot[smooth,BLUE,line width=1.5pt] plot coordinates{
				(-5.000000, 0.317826)(-4.000000, 0.311145)(-3.000000, 0.305098)(-2.000000, 0.299973)(-1.000000, 0.296055)(0.000000, 0.293591)(1.000000, 0.292750)(2.000000, 0.293591)(3.000000, 0.296055)(4.000000, 0.299973)(5.000000, 0.305098)
			};
			\addlegendentry{{Stochastic prediction }};
			\end{axis}
			\end{tikzpicture}
	\end{tabular}
		\caption{Comparison between the \emph{one-shot} empirical classification error of combined classifier $\sum_{i=1}^2a_i\hat\bbeta_{\ell_i}(0)$ as a function of $\rho = \hat\theta_1^{-1}\hat\eta_1a_1/(\sum_{i=1}^2\hat\theta_i^{-1}\hat\eta_ia_i)$ and its stochastic high dimensional prediction ${\rm Err}(\bxi)$ with $\bxi$ given in Corollary~\ref{cor:performance of combination unreg}, for $p=256$ and $n=10p$. \textbf{Left}: $\ell_1(t) = \ln(1+e^{-t})$, $\ell_2(t) = \exp(-t)$, $\bmu = [1;~\mathbf{0}_{p-1}]$, $\C =2 \cdot \I_p$. \textbf{Right}: $\ell_1(t) = (t-1)^2/2$, $\ell_2(t) = \ln(1+e^{-t})$, $\bmu = [\ones_{p/2};~-\ones_{p/2}]/\sqrt{2p}$, $[\C]_{ij} =0.1^{|i-j|}$. }
		\label{fig:ensemble-perf-compare}
\end{figure}

	\section{Optimality Results}
	\label{sec:optimality}
	
	In Theorem~\ref{theo:main}, we gave an exact statistical characterization of the generalized solution $\hat\bbeta$ to the empirical risk minimization problem \eqref{eq:opt-origin}, for high dimensional data. This characterization involves three parameters $(\theta, \eta, \gamma)$ that can either be determined in an stochastic manner or an deterministic one. In this section, we will demonstrate with the theoretical results in Section~\ref{sec:statistical characterization} that the classification error given by the solution of square loss, associated with LDA classifier as explained in Section~\ref{sec:problem}, is the lowest among all smooth and convex loss functions under Assumption~\ref{ass:loss}, including the logistic loss yielding the maximal likelihood solution. It is remarkable that the optimality of the square loss is universal in the sense that it holds for all values of $n/p>0$  and is independent of the statistical parameters $\bmu,\C$ of the data distribution. 
		
	The optimality of the square loss is two-fold: (i) the square loss is the optimal choice in the absence of regularization; (ii) with regularization, the classification performance given by the square loss at optimal regularization $\lambda$ is better than any other loss at its respective optimal value of $\lambda$. We discuss the unregularized case in Section~\ref{sec:unregularized solutions}, before moving on to the regularized one in Section~\ref{sec:regularized solutions}. The mathematical arguments employed in Section~\ref{sec:unregularized solutions} concern the stochastic description of $\hat\bbeta$ provided in the second point of Theorem~\ref{theo:main}, while the proof in Section~\ref{sec:regularized solutions} makes use of the deterministic description in the first point. It worth pointing out that the optimality of the square loss in the unregularized scenario can be also proven as a special case of the derivation in Section~\ref{sec:regularized solutions}, however the approach of Section~\ref{sec:unregularized solutions} is stronger as it shows additionally that the unregularized solution obtained with the square loss yields the optimal classification error among all possible linear combinations of unregularized solutions given by different loss functions. Another important remark to be made on the different interests of these two approaches is that while the one in Section~\ref{sec:regularized solutions} does not cover linearly combined solutions, it provides more direct insights into what distinguishes the square loss and makes it the optimal choice.
	
	\subsection{Unregularized Solutions}
	\label{sec:unregularized solutions}
	 As pointed out in the discussion following Theorem~\ref{theo:main}, all unregularized solutions $\hat\bbeta_\ell(0)$ are unbiased in direction, in the sense that $\E[\hat\bbeta_\ell(0)]$ is aligned with the underlying true parameter vector $\bbeta_*$ in the limit of high dimensions. The optimal solution is thus the one suffering the least from the effect of variance. Recall the expression of the (expected) classification error from \eqref{eq:Err}, we get by applying \eqref{eq:tilde beta} and \eqref{eq:hat theta eta gamma} that 
	\begin{align}
			{\rm Err}\left(\hat\bbeta_\ell(0)\right)=Q\left(\frac{\bmu^\T\C^{-1}\bmu}{\sqrt{\bmu^\T\C^{-1}\bmu+\hat\eta^{-2}\hat\gamma^2}}\right)+o_P(1)=Q\left(\frac{\bmu^\T\C^{-1}\bmu}{\sqrt{\bmu^\T\C^{-1}\bmu+\frac{p\Vert\c\Vert^2}{(\ones_n^\T\c)^2}}}\right)+o_P(1).\label{eq:peformance unreg}
		\end{align}
		It follows immediately that 
		\begin{align}
		\label{eq:optimal unregularized solution condition c}
		\argmin_{\ell}{\rm Err}\left(\hat\bbeta_\ell(0)\right)=\argmin_{\ell}\frac{\Vert\c\Vert}{\ones_n^\T\c}
		\end{align}
		holds with high probability for large $p$, meaning that the loss function having the smallest value of $\Vert\c\Vert/(\ones_n^\T\c)$ gives the optimal unregularized solution for high dimensional data in term of classification error.

		Now notice that, for any loss function $\ell$, we always have  
		$\X_\y\c=\zeros_p$
		where $\X_\y=[y_1\x_1,\ldots,y_n\x_n]$, by cancelling the gradient of \eqref{eq:opt-origin} for $\lambda=0$. Denote by $\c_{{\rm LS}}$ the associated vector $\c$ for the unregularized solution $\hat\bbeta_{{\rm LS}}(0)$ obtained with the square loss. We are going to derive from \(\X_\y\c=\zeros_p\) the key result that for any $\c$ associated with some $\hat\bbeta_\ell(0)$, we have 
		\begin{align}
		\label{eq:relation c and cLS}
		\c= a \c_{{\rm LS}}+\c_{\perp}
		\end{align}
		for some $a\geq 0$ and $\c_{\perp}$ orthogonal to $\ones_n$. To obtain this result, let us consider the singular value decomposition 
		\begin{equation*}
		\X_\y = \U \bSigma \V^\T
		\end{equation*}
		for some unitary matrices \(\U\in\RR^{p\times p}\), \(\V\in\RR^{n\times n}\). As \(\X_\y \in \RR^{p \times n}\) is of rank \(p\) for \(n>p\) with probability one, one can arrange the unitary matrices $\U,\V$ such that \(\bSigma = \begin{bmatrix} \S & \mathbf{0} \end{bmatrix}\) with \( \S \in \RR^{p \times p}\) a diagonal matrix with positive diagonal entries. Write \(\V = \begin{bmatrix} \V_1 & \V_2 \end{bmatrix}\) with \(\V_1 \in \RR^{n \times p}\) and \(\V_2 \in \RR^{n \times (n-p)}\). It follows from \(\X_\y\c=\mathbf{0}\) that \(\V_1^\T \c=\mathbf{0}\). The vector \(\c\) thus lies in the subspace spanned by the column vectors of \(\V_2\). We are then able to write $\c$ as 
		$$\c=\V_2\bdeta$$
		with $\bdeta\in\RR^p$. Recall from \eqref{eq:beta_LS} that the unregularized solution obtained with the square loss has the following explicit form
		$$\hat\bbeta_{{\rm LS}}(0)=\left(\X\X^\T\right)^{-1}\X\y=\left(\X_\y\X_\y^\T\right)^{-1}\X_\y\ones_n.$$ We get thus
		$$\c_{{\rm LS}}=\ones_n-\X_\y^\T\hat\bbeta_{{\rm LS}}(0)=\V_2\V_2^\T\ones_n.$$
		Therefore, $\c_{{\rm LS}}$ is the projection of $\ones_n$ onto the subspace spanned by the column vectors of $\V_2$. Understandably, any vector that lies in that subspace and orthogonal to $\c_{{\rm LS}}$ is also orthogonal to $\ones_n$. Since any vector can be decomposed into two parts with one aligned and the other orthogonal to $\c_{{\rm LS}}$, we derive the key result given in \eqref{eq:relation c and cLS}. Explicitly, we have 
		\begin{align*}
		\c=\V_2\bdeta=\left(\frac{\bdeta^\T\V_2^\T\ones_n}{\Vert\V_2^\T\ones_n\Vert^2}\right)\c_{{\rm LS}}+\V_2\left(\bdeta-\frac{\bdeta^\T\V_2^\T\ones_n}{\Vert\V_2^\T\ones_n\Vert^2}\V_2^\T\ones_n\right)
		\end{align*}
		where we note that $\V_2\left(\bdeta-\frac{\bdeta^\T\V_2^\T\ones_n}{\Vert\V_2^\T\ones_n\Vert^2}\V_2^\T\ones_n\right)=\c_{\perp}$ is orthogonal to $\ones_n$. 
		
		As a direct consequence of \eqref{eq:relation c and cLS}, we have
		$$\frac{\Vert\c\Vert}{\ones_n^\T\c}=\frac{\sqrt{\Vert a\c_{{\rm LS}}\Vert^2+\Vert \c_{\perp}\Vert^2}}{a\ones_n^\T\c_{{\rm LS}}}\leq \frac{\Vert\c_{{\rm LS}}\Vert}{\ones_n^\T\c_{{\rm LS}}}.$$
		Combining the above inequality with \eqref{eq:optimal unregularized solution condition c} leads to the below theorem.
		\begin{Theorem}[Optimality of square loss for unregularized classifier]
		\label{theo:optimal-unreg}
			Let Assumptions~\ref{ass:growth-rate}~and~\ref{ass:loss} hold. Then, for any $n/p>1$, we have that
			\begin{align*}
			{\rm Err}\left(\hat\bbeta_{{\rm LS}}(0)\right)\leq \min_{\ell}{\rm Err}\left(\hat\bbeta_\ell(0)\right)
			\end{align*}
			holds with high probability, where we recall $\hat\bbeta_{{\rm LS}}(0)$ is the solution to \eqref{eq:opt-origin} with the square loss $\ell(t)=(1-t)^2/2$ at $\lambda=0$.
		\end{Theorem}
		
		Moreover, with the help of \eqref{eq:relation c and cLS} and Corollary~\ref{cor:performance of combination unreg}, we can show that $\hat\bbeta_{{\rm LS}}(0)$ outperforms not only all other unregularized solutions individually, but also all linear combination of them. As a result of Theorem~\ref{theo:joint distribution},  which characterizes the joint distribution of solutions obtained with different loss function, Corollary~\ref{cor:performance of combination unreg} states that the classification error given by some linear combination $\sum_{i=1}^M a_i\hat\bbeta_{\ell_i}(0)$ of $\hat\bbeta_{\ell_1}(0),\ldots,\hat\bbeta_{\ell_m}(0)$ can be accessed with the two quantities $\tau,\omega$ given in\eqref{eq:tau omega}. Similarly to the discussion on individual classifiers, we observe that the comparison of the high dimensional classification errors given $\hat\bbeta_{{\rm LS}}(0)$ and that $\sum_{i=1}^M a_i\hat\bbeta_{\ell_i}(0)$ reduces to that of $\frac{\Vert\c_{{\rm LS}}\Vert}{\ones_n^\T\c_{{\rm LS}}}$ and $\frac{\Vert\sum_{i=1}^M a_i\c_i\Vert}{\sum_{i=1}^M  a_i \ones_n^\T\c_i}$ where $\c_i$ is the vector $\c$ associated with $\hat\bbeta_{\ell_i}(0)$. Reapplying the relation \eqref{eq:relation c and cLS} between arbitrary $\c$ and $\c_{{\rm LS}}$, we deduce the superiority of $\hat\bbeta_{{\rm LS}}(0)$ over $\sum_{i=1}^M a_i \hat\bbeta_{\ell_i}(0)$, as stated in the following theorem.
		\begin{Theorem}[Error lower bound for linear combination of unregularized classifiers]
		\label{theo:optim-linear-combination}
			Let Assumptions~\ref{ass:growth-rate}~and~\ref{ass:loss} hold. Then, for any $n/p>1$, we have that
			\begin{align*}
			{\rm Err}\left(\hat\bbeta_{{\rm LS}}(0)\right)\leq \min_{\ell_1,\ldots,\ell_m}\min_{a_1,\ldots,a_m\in\RR}{\rm Err}\left(\sum_{i=1}^M a_i\hat\bbeta_{\ell_i}(0)\right)
			\end{align*}
			holds with high probability for any $m\in\mathbb{N}^*$, where we recall $\hat\bbeta_{{\rm LS}}(0)$ is the solution to \eqref{eq:opt-origin} with the square loss $\ell(t)=(1-t)^2/2$ at $\lambda=0$.
		\end{Theorem}

	\subsection{Regularized Solutions}
	\label{sec:regularized solutions}
	
	It is understandably not fair to compare loss functions at some fixed $\lambda\neq 0$, due to the different (directional) bias levels reflected by $\lambda/\theta$ as indicated in \eqref{eq:tilde beta}, where $\theta$ varies implicitly with the choice of loss function as described in \eqref{eq:theta eta gamma}. We start therefore by discussing the competitiveness of loss functions at the same bias level.  In other words, we are interested in comparing a (potentially infinite) set of classifiers  $\hat\bbeta_{\ell}(\lambda)$ satisfying
		\begin{align*}
			\lambda/\theta=\omega
		\end{align*}
	for some fixed $\omega\geq 0$. Recall from Corollary~\ref{cor:classification error} that the classification error yielded by $\hat\bbeta_{\ell}(\lambda)$ is asymptotically equal to $Q(m/\sigma)$, with $m,\sigma$ functions of $\theta,\eta,\gamma$ as indicated in \eqref{eq:m sigma}.

	Note importantly that $\theta,\eta,\gamma$ are expressed in \eqref{eq:theta eta gamma} as some statistics of random variables $r\sim\mathcal{N}(m,\sigma^2)$ and $h(r)$ with the function $h(\cdot)$ dependent of $\ell(\cdot)$ as specified in Theorem~\ref{theo:main}.
	Since
    \begin{align}
    \label{eq:inequality cov(hr,r)}
    -\cov[h(r),r]\leq \sqrt{\var[h(r)]\var[r]},
	\end{align}
	by Cauchy-Schwarz inequality, we get from \eqref{eq:theta eta gamma} and \eqref{eq:m sigma} that
	\begin{align}
	\label{eq:inequality theta}
	    \theta^2\leq \frac{\frac{n}{p}\gamma^2-\eta^2}{\sigma^2}.
	\end{align}
	We observe furthermore by plugging \eqref{eq:tilde beta} into \eqref{eq:m sigma} that
	\begin{align*}
		\sigma^2 =&\frac{\eta^2}{\theta^2}\bmu^\T\left(\frac{\lambda}{\theta}\I_p+\C\right)^{-1}\C\left(\frac{\lambda}{\theta}\I_p+\C\right)^{-1}\bmu+\frac{\gamma^2}{\theta^2p}\tr\left[\left(\frac{\lambda}{\theta}\I_p+\C\right)^{-1}\C\right]^2\\
		\geq&\frac{\eta^2}{\theta^2}\bmu^\T\left(\frac{\lambda}{\theta}\I_p+\C\right)^{-1}\C\left(\frac{\lambda}{\theta}\I_p+\C\right)^{-1}\bmu+\frac{\eta^2}{\theta^2n}\tr\left[\left(\frac{\lambda}{\theta}\I_p+\C\right)^{-1}\C\right]^2+\frac{\sigma^2}{n}\tr\left[\left(\frac{\lambda}{\theta}\I_p+\C\right)^{-1}\C\right]^2
		\end{align*} where the inequality is justified by \eqref{eq:inequality theta}. With the expression of $m$ also given in \eqref{eq:m sigma}, it follows immediately that
      \begin{align}
      \label{eq:inequality m over sigma}
		\frac{m^2}{\sigma^2}\geq e\left(\frac{\lambda}{\theta}\right)=e\left(\omega\right) 
		\end{align}
		with
			\begin{align}
			\label{eq:e}
				e\left(\omega\right)=\frac{\left\{1-\frac{1}{n}\tr\left[\left(\omega\I_p+\C\right)^{-1}\C\right]^2\right\}\left[\bmu^\T\left(\omega\I_p+\C\right)^{-1}\C\left(\omega\I_p+\C\right)^{-1}\bmu\right]^2}{\bmu^\T\left(\omega\I_p+\C\right)^{-1}\C\left(\omega\I_p+\C\right)^{-1}\bmu+\frac{1}{n}\tr\left[\left(\omega\I_p+\C\right)^{-1}\C\right]^2}.
			\end{align}
    Since the inequality in \eqref{eq:inequality m over sigma} stems from \eqref{eq:inequality cov(hr,r)}, a necessary and sufficient condition for $m^2/\sigma^2=e\left(\omega\right)$ is 
	\begin{align}
	\label{eq:equality theta}
		-\cov[h(r),r]=\sqrt{\var[h(r)]\var[r]}
	\end{align}
	A key observation in our derivation is that the condition \eqref{eq:equality theta} holds for  the square loss $\ell(t)=(1-t)^2/2$.  Recall from Theorem~\ref{theo:main} that the function $h(\cdot)$ is given as $h(\cdot) = \frac{\ell'(g_{\kappa, \ell} (r)-r}{\kappa}$ for some constant $\kappa>0$ and  $g_{\kappa, \ell} (\cdot)$ as defined in \eqref{eq:def-g}.  Letting $\ell(t)=(1-t)^2/2$, we get then
	\begin{align*}
		h(r) =\frac{1-r}{1+\kappa}.
	\end{align*}
	The condition \eqref{eq:equality theta} is thus satisfied since
	\begin{align*}
		-\cov[h(r),r]=(1+2\kappa)^{-1}\var[r]=\sqrt{\var[h(r)]\var[r]},
		\end{align*}
    which concludes the proof of the following theorem.
    \begin{Theorem}[Error lower bound of regularized classifiers at fixed directional bias]
    	\label{theo:optimality reg}
    	Under the conditions and notations of Theorem~\ref{theo:main}, for any classifier $\hat\bbeta_{\ell}(\lambda)$ satisfying $\lambda/\theta=\omega$ with some fixed $\omega\geq 0$, we have that
        \begin{align*}
        	{\rm Err}(\hat\bbeta_\ell(\lambda))\geq Q\left(\sqrt{e(\omega)}\right)
        \end{align*} 
        with high probability, where $e(\omega)$ is given in \eqref{eq:e}; moreover, if there exists a positive value $\lambda_{\rm LS}$ such that $\lambda_{\rm LS}/\theta_{\rm LS}=\omega$ where $\theta_{\rm LS}$ is the value of $\theta$ associated with $\hat\bbeta_{\rm LS}(\lambda_{\rm LS})$, then
        \begin{align*}
        	{\rm Err}(\hat\bbeta_{\rm LS}(\lambda_{\rm LS}))=Q\left(\sqrt{e(\omega)}\right)+o_P(1).
        \end{align*}
    \end{Theorem}
     
 In plain words, Theorem~\ref{theo:optimality reg} tells us that all regularized classifiers $\hat\bbeta_{\ell}(\lambda)$ having the same directional bias, in the sense that they all asymptotically align in expectation with $\left(\omega\I_p+\C\right)^{-1}\bmu$ for some fixed $\omega\geq 0$, yield classification errors lower bounded by $Q\left(\sqrt{e(\omega)}\right)$, achievable by the regularized classifier of square loss $\hat\bbeta_{\rm LS}(\lambda_{\rm LS})$ with the same directional bias (if it exists). To demonstrate that, for a certain $n/p>0$,  $\hat\bbeta_{\rm LS}(\lambda_{\rm LS})$ yields the best achievable performance over the space of hyperparameters (i.e., $\ell$ and $\lambda$) with an optimally set $\lambda_{\rm LS}$, it suffices to show that $\hat\bbeta_{\rm LS}(\lambda_{\rm LS})$ gives the widest range of $\omega$ as $\lambda_{\rm LS}$ goes from zero to infinity. We refer to Appendix~\ref{sm:proof-of-proposition-omega} for the proof of this result, which is formulated in Proposition~\ref{prop:omega} below.
     \begin{Proposition}[Attainable range of $\lambda/\theta$]
     	\label{prop:omega}
     	 Under the conditions and notations of Theorem~\ref{theo:main}, we have that, for any \(n/p>0\), 
     	     \begin{align*}
     		\min_{\lambda_{\rm LS}\geq 0}  \lambda_{\rm LS}/\theta_{\rm LS}\leq \min_{\ell,\lambda\geq 0}\lambda/\theta
     		\end{align*}
     		and 
     		\begin{align*}
     		\max_{\lambda_{\rm LS}\geq 0}  \lambda_{\rm LS}/\theta_{\rm LS}\geq \max_{\ell,\lambda\geq 0}\lambda/\theta
     		\end{align*}
     		hold with high probability, where $\theta_{\rm LS}$ stands for the value of $\theta$ associated with $\hat\bbeta_{\rm LS}(\lambda_{\rm LS})$.
     		
     \end{Proposition}
     
     
     This leads to the following result on the generalized optimality of square loss in the presence of ridge regularization.
      \begin{Theorem}[Optimality of square loss for regularized classifier]
     	\label{theo:optimality reg 2}
     	Let Assumptions~\ref{ass:growth-rate}~-~\ref{ass:loss} hold. Then, for any \(n/p>0\), we have that
     \begin{align*}
    \min_{\lambda\geq 0} {\rm Err}\left(\hat\bbeta_{{\rm LS}}(\lambda)\right)\leq \min_{\ell}\min_{\lambda\geq 0}{\rm Err}\left(\hat\bbeta_\ell(\lambda)\right)
     \end{align*}
     holds with high probability; and
     \begin{align*}
     	 \min_{\lambda\geq 0} {\rm Err}\left(\hat\bbeta_{{\rm LS}}(\lambda)\right)=\min_{\omega\geq 0}Q\left(e(\omega)\right)+o_P(1).
     \end{align*}
     \end{Theorem}
     
     \section{Extension to non-Gaussian settings}
     \label{sec:universality}
     
     Many results in random matrix theory, while first derived under the Gaussianity assumption, were found later to hold universally over a much larger family of distributions under relatively mild conditions \citep{tao2010random,bai2008clt}. The major difference between these results and ours is that our object of study are implicit non-linear mappings of random matrices. In spite of that, we discover that most of our results, including the precise error and the optimality of square loss, hold for non-Gaussian mixture data, under some additional non-sparsity condition on $\bmu$ with respect to the eigenspace of $\C$, as detailed in Assumption~\ref{ass:non-sparsity}.
     \begin{Assumption}[Non-sparsity of $\bmu$ with respect to $\C$]
     \label{ass:non-sparsity}
         Denote $\C=\V\bLambda^2\V^\T$ the spectral decomposition of $\C$ and $\{\v_d\}_{d=1}^p$ the column vectors of $\V$. For $d \in \{1,\ldots,p\}$, we have  $\bmu^\T\v_d=O(\Vert\bmu\Vert/\sqrt{p})$.
     \end{Assumption}
     
     Before presenting the universality results, we define the following general framework of mixture data:
     \begin{equation}
         \label{eq:general mixture}
         	\begin{cases}
	y = -1 &\Leftrightarrow \x= -\bmu+\V\bLambda\z\\
	y = +1 &\Leftrightarrow \x = +\bmu+\V\bLambda\z,
	\end{cases}\end{equation}
	for some unitary matrix $\V\in\RR^{p\times p}$, positive diagonal matrix $\bLambda\in\RR^{p\times p}$, and random vector $\z\in\RR^{p}$ having independent entries of zero mean, unit variance and bounded fourth moment. We remark that $\C=\V\bLambda^2\V^\T$ is the covariance matrix of $\x$.
	
	\begin{Theorem}[Universality beyond Gaussian distribution]
	\label{theo:universality over non-gaussian data}
	Let Assumptions~\ref{ass:growth-rate}~-~\ref{ass:non-sparsity} hold, and training samples $\{(\x_i,y_i)\}_{i=1}^n$ be drawn independently and uniformly from the ``generic'' mixture model in \eqref{eq:general mixture}, we have
	\begin{equation*}
	    {\rm Err}(\hat\bbeta_\ell(\lambda))= Q\left(\frac{m}{\sigma}\right)+o_P(1)
	\end{equation*}
	with $m,\sigma^2$ given in \eqref{eq:m sigma}. 
	Consequently, Theorem~\ref{theo:optimal-unreg},~\ref{theo:optimality reg}~and~\ref{theo:optimality reg 2} also hold beyond the Gaussian setting.
	\end{Theorem}
	We refer to Appendix~\ref{sm:proof-of-theorem-unitersality} for the proof of the above theorem.

	\section{Conclusion}
	\label{sec:conclusion}
	
	In this article, we investigated the problem of high dimensional classification under the general algorithmic framework of empirical risk minimization and the classical setting of Gaussian mixture data. We showed that the high dimensional distribution of the classifier parameter $\hat \bbeta$ given in \eqref{eq:opt-origin} is a Gaussian vector controlled by three deterministic constants that depend on the statistical parameters of the mixture model and the number of training samples. These constants can also be estimated empirically from the training samples. The full statistical description of $\hat \bbeta$ allows us to understand on a deep level the learning performance of the empirical risk minimization approach. Based on this result, we proceeded further to the study of the optimal loss and explained how the statistical description of $\hat\bbeta$ provided in Theorem~\ref{theo:main} leads to the identification of square loss as the optimal choice in the unregularized or regularized case. Our analysis served also to statistically characterize linear combinations of classifiers learned with different loss functions, allowing for the important conclusion on the limitation of this ensemble learning approach that the improved performance of linearly combining unregularized solutions is upper bounded by that of the square loss solution alone.
	
	\medskip
	
	
	
	Although our proof uses the condition of smooth loss functions,  our results can be applied to non-smooth losses with the help of proximal mappings, as explained in the paragraphs above Theorem~\ref{theo:main}. To go around the technical hurdle, one may consider a non-smooth loss as the limit of a series of smooth losses. The present study can also be further developed to encompass other forms of regularization, a necessary step to see if the optimality of square loss is specific to the case of ridge-regularization or it is a more universal phenomenon. It is also of interest to track the evolution dynamics of the underlying optimization problem, for instance as a function of the number of descent steps when solved with gradient-based methods, which is closely related to the training of modern neural networks \citep{saxe2013exact,liao2018dynamics}.
	

		
\vskip 0.2in
\bibliography{RMT4LR}
	
	\clearpage
	

\appendix
	
\paragraph{Notations.} Before getting into the proofs, we first introduce the following asymptotic notations. The big $O$ notation \(O(u_n)\) is understood here in probability. We specify that when multidimensional objects are concerned, $O(u_n)$ is understood entry-wise. The notation \(O_{\Vert\cdot\Vert}(\cdot)\) is understood as follows: for a vector $\v$, $\v=O_{\Vert\cdot\Vert}(u_n)$ means its Euclidean norm is $O(u_n)$ and for a square matrix $\mathbf{M}$, $\mathbf{M}=O_{\Vert\cdot\Vert}(u_n)$ means that the operator norm of $\mathbf{M}$ is $O(u_n)$. The small $o$ notation is understood likewise. Note that under Assumption~\ref{ass:growth-rate} it is equivalent to use either $O(u_n)$ or $O(u_p)$ since $n, p$ scales linearly. In the following we shall use constantly $O(u_p)$ for simplicity of exposition. The symbol $\simeq$ is used in the following sense: for a scalar $s=O(1)$, $s\simeq \tilde s$ indicates that $(s-\tilde s)=o(1)$, and for a vector $\v$ with $\Vert \v\Vert=O(1)$, $\v\simeq \tilde \v$ means $\Vert \v-\tilde \v\Vert=o(1)$. For random variable $r\sim\mathcal{N}(m,\sigma^2)$ with potentially random $m$ and $\sigma^2$, the expectation $\E[f(r)]$ should be understood as conditioned on $m,\sigma^2$, so that, $\E[r]$ is equal to $m$ instead of $\E[m]$. When parametrized functions $f_{\tau}(\cdot)$ are involved, $\E[f_{\tau}(r)]$ is computed by taking the integral over $r$.

	\section{Proof of Theorem~\ref{theo:main}}\label{sm:proof-of-theorem-main}

		{\bf Main idea and key steps:} As the optimization problem \eqref{eq:opt-origin}  does not generally assume an explicit solution, we turn our attention to the stationary-point expression (where we use the shortcut notation  $\hat \bbeta $ for $\hat \bbeta_{\ell}(\lambda)$) 
		\begin{equation}\label{eq:lambda-beta}
		\lambda \hat \bbeta = \frac1n \sum_{i=1}^n -\ell'(y_i\x_i^\T\hat\bbeta)y_i \x_i,
		\end{equation}
		obtained by cancelling the gradient. To characterize the behavior of $ \hat \bbeta$ from the above expression, we need to address the statistics of $y_i\x_i^\T\hat\bbeta$, which is for the moment not directly tractable due to the implicit dependence of $\hat \bbeta$ on all $y_i\x_i$. To this end, an important concept in our proof is a ``leave-one-out'' version of $\hat \bbeta$, denoted $\hat \bbeta_{-i}$, obtained by solving \eqref{eq:opt-origin} with all the remaining $n-1$ training samples $(\x_j,y_j)$ for $j \neq i$. This leave-one-out solution $\hat \bbeta_{-i}$ has two crucial properties: (i) it is by definition \emph{independent} of the left-out data sample $(\x_i,y_i)$; and (ii) it is close to the original $\hat \bbeta$ as removing one among $n$ training samples has a negligible effect as $n \to \infty$. Naturally, the (asymptotic) generalization performance of $\hat\bbeta$ is reflected by the probability of $y_i\x_i^\T\hat\bbeta_{-i}>0$ (since again, $y_i\x_i$ is independent of $\hat\bbeta_{-i}$ and can be seen as a new test datum), and the training performance by the probability of $y_i\x_i^\T\hat\bbeta>0$. The first key result in our derivation is the following \emph{nonlinear} relation between $y_i\x_i^\T\hat\bbeta$ and $y_i\x_i^\T\hat\bbeta_{-i}$:
		\begin{align}
		\label{eq:leave-one-out prediction}
			y_i\x_i^\T\hat\bbeta_{-i}\simeq y_i\x_i^\T\hat\bbeta+\kappa \ell'(y_i\x_i^\T\hat\bbeta)
		\end{align}
		for some constant $\kappa>0$. We can thus approximate $y_i\x_i^\T\hat\bbeta$ as a function of  $	y_i\x_i^\T\hat\bbeta_{-i}$ by writing:
		\begin{align*}
			y_i\x_i^\T\hat\bbeta \simeq g_{\kappa,\ell}\left(y_i\x_i^\T\hat\bbeta_{-i}\right)
		\end{align*}
		where $t\mapsto g_{\kappa,\ell}(t)$ is the inverse mapping of $t\mapsto t+\kappa\ell'(t)$ as stated in \eqref{eq:def-g} that is guaranteed to exist and be unique. Immediately, we have
		\begin{align*}
		 c_i\simeq \tilde c_i 
		\end{align*}
		where $c_i=-\ell'(y_i\x_i^\T\hat\bbeta)$ as defined in \eqref{eq:def-c} and $\tilde c_i=\frac{g_{\kappa,\ell}\left(y_i\x_i^\T\hat\bbeta_{-i}\right)-y_i\x_i^\T\hat\bbeta_{-i}}{\kappa}=h\left(y_i\x_i^\T\hat\bbeta_{-i}\right)$ with $h(t)=\frac{g_{\kappa,\ell}(t)-t}{\kappa}$ as defined in Theorem~\ref{theo:main} .
		We can therefore approximate \eqref{eq:lambda-beta} as
		\begin{align}
		\label{eq:approx-beta}
				\lambda \hat \bbeta = \frac1n \sum_{i=1}^n c_iy_i \x_i \simeq \frac1n \sum_{i=1}^n \tilde c_iy_i \x_i 
		\end{align}
		where we trade $y_i\x_i^\T\hat\bbeta$ for the more tractable $y_i\x_i^\T\hat\bbeta_{-i}$, which is simply the product of two independent vectors $y_i\x_i$ and $\hat\bbeta_{-i}$. The remaining steps aim to determine the asymptotic distribution of $\hat\bbeta$ from \eqref{eq:approx-beta}.  Letting $y_i\x_i=\bmu+\w_i$ with $\w_i\sim\mathcal{N}(\zeros_p,\C)$, we decompose $\tilde c_iy_i\x_i$ as
		\begin{align*}
	   \tilde c_iy_i\x_i=\tilde c_i\bmu+\E[\tilde c_i\w_i~\vert~\hat\bbeta_{-i}]+(\tilde c_i\w_i-\E[\tilde c_i\w_i~\vert~\hat\bbeta_{-i}]).
		\end{align*}
		By showing that 
		\begin{align}
		    \E[\tilde c_i\w_i~\vert~\hat\bbeta_{-i}]&=\frac{\E[\tilde c_i\w_i^\T\hat\bbeta_{-i}~\vert~\hat\bbeta_{-i}]}{\hat\bbeta_{-i}^\T\C\hat\bbeta_{-i}}\C\hat\bbeta_{-i}\label{eq:expectation ciwi}
		\end{align}
		and
		\begin{align}
		    \frac{1}{n}\sum_{i=1}^n(\tilde c_i\w_i-\E[\tilde c_i\w_i~\vert~\hat\bbeta_{-i}])\simeq \bnu, \quad \bnu\sim\mathcal{N}\left(\zeros_p, \frac{\sum_{i=1}^n\tilde c_i^2}{n}\C\right)\label{eq:v},
		\end{align} we have that $\hat\bbeta$ is asymptotically close (in norm) to the following Gaussian vector
		\begin{equation}
		    \left(\lambda\I_p-\frac{\E[\tilde c_i\w_i^\T\hat\bbeta_{-i}~\vert~\hat\bbeta_{-i}]}{\hat\bbeta^\T\C\hat\bbeta}\C\right)^{-1} \cdot \mathcal{N}\left(\frac{\sum_{i=1}^n\tilde c_i}{n}\bmu,\frac{\sum_{i=1}^n\tilde c_i^2}{n^2}\C\right).
		\end{equation}
    By demonstrating that 
    \begin{align}
    &\frac{1}{n}\sum_{i=1}^n\tilde c_i\simeq\frac{1}{n}\sum_{i=1}^n c_i\simeq\E[h(r)]\label{eq:E hr}\\
    &\frac{1}{n}\sum_{i=1}^n\tilde c_i^2\simeq\frac{1}{n}\sum_{i=1}^n c_i^2\simeq\E[h^2(r)]\label{eq:E squre hr}
    \end{align}
   with\footnote{There is a slight difference between the random variable $r$ defined here and the one in Theorem~\ref{theo:main}. This is reconciled by the fact that $\bmu^\T\hat\bbeta\simeq \bmu^\T\E[\hat\bbeta]$ and $\hat\bbeta^\T\C\hat\bbeta\simeq\E[\hat\bbeta^\T\C\hat\bbeta]$, which is a consequence of \eqref{eq:v}--\eqref{eq:var r}.} $r\sim\mathcal{N}\left(\bmu^\T\hat\bbeta,\hat\bbeta^\T\C\hat\bbeta\right)$, $h(\cdot)=-\ell'\left(g_{\kappa,\ell}\left(\cdot\right)\right)$, and also
   \begin{align}
   &\E[\tilde c_i\w_i^\T\hat\bbeta_{-i}~\vert~\hat\bbeta_{-i}] \simeq \frac{1}{n}\sum_{i=1}^n c_i\left(r_i-\sum_{i=1}^n r_i\right)\simeq\cov[h(r),r]\label{eq:cov hr r}\\
  & \hat\bbeta^\T\C\hat\bbeta\simeq\frac{1}{n}\sum_{i=1}^n \left(r_i-\sum_{i=1}^n r_i\right)^2\simeq \var[r]\label{eq:var r}
   \end{align}
  for $r_i= y_i\x_i^\T\hat\bbeta-\kappa c_i=y_i\x_i^\T\hat\bbeta_{-i}+O(p^{-\frac{1}{2}})$, we will  have all the ingredients to obtain the results of Theorem~\ref{theo:main} except the missing characterization  of $\kappa$, which will be determined in the subsequent detailed  proof, along with the demonstration of the key arguments~\eqref{eq:leave-one-out prediction}--\eqref{eq:var r}.
  
		\bigskip
			
	{\bf Detailed arguments:} As mentioned above, our first step is to demonstrate the key result in \eqref{eq:leave-one-out prediction}.  Let us start by remarking that $\Vert\hat\bbeta\Vert=O(1)$ for $\lambda>0$, which is the case that we focus on before discussing the unregularized solutions by taking $\lambda\to0$. Indeed, since $\hat\bbeta$ is the minimizer of \eqref{eq:opt-origin} for some non-negative loss $\ell(\cdot)$ with finite $\ell(0)$ under Assumption~\ref{ass:loss}, we have that
	\begin{align*}
		 \frac1n \sum_{i=1}^n \ell(y_i\x_i^\T \hat\bbeta) + \frac{\lambda}2 \| \hat\bbeta \|^2\leq  \frac1n \sum_{i=1}^n \ell(0)
	\end{align*}
	and therefore $\| \hat\bbeta \|^2\leq  \frac{2}{\lambda n} \sum_{i=1}^n \ell(0)=O(1)$ for $\lambda > 0$.
	
	As the leave-one-out solution $\hat \bbeta_{-i}$ is obtained by solving  \eqref{eq:opt-origin} without the $i$-th training sample $(\x_i,y_i)$, similar to \eqref{eq:lambda-beta}, it follows that 
		\begin{equation}
		    \lambda \hat \bbeta_{-i} = \frac1n \sum_{j \neq i} - \ell'(y_j \x_j^\T \hat \bbeta_{-i}) y_j \x_j.
		\end{equation}
		Taking the difference of the above equation with \eqref{eq:lambda-beta}, we get
		\begin{align*}
		\lambda (\hat \bbeta - \hat \bbeta_{-i}) = &\frac1n \sum_{j \neq i} -\left(\ell'(y_j \x_j^\T \hat \bbeta ) - \ell'(y_j\x_j^\T \hat \bbeta_{-i}) \right) y_j\x_j - \frac1n \ell'(y_i \x_i^\T \hat \bbeta) y_i \x_i\\
		=& \frac1n \sum_{j \neq i} -\ell''(t_{-i,j}) (y_j \x_j^\T \hat \bbeta-y_j\x_j^\T \hat \bbeta_{-i})y_j\x_j - \frac1n \ell'(y_i \x_i^\T \hat \bbeta) y_i \x_i
		\end{align*}
		where $\ell''(t_{-i,j}) \geq 0$ for some $t_{-i,j}$ between $y_j \x_j^\T \hat \bbeta$ and $y_j\x_j^\T \hat \bbeta_{-i}$ by the convexity of $\ell(\cdot)$ and the mean value theorem. We thus remark that
		\begin{equation}\label{eq:dif-beta-beta_i}
			\hat \bbeta - \hat \bbeta_{-i}=\frac{- \ell'(y_i \x_i^\T \hat \bbeta) }{n}\left(\lambda\I_p+\frac{1}{n}\sum_{j \neq i}\ell''(t_{-i,j})\x_j\x_j^\T\right)^{-1}y_i \x_i
		\end{equation}
		with $y_j^2 = 1$. As direct consequences, we have
		\begin{equation}\label{eq:bound-norm-dif-beta}
			\Vert\hat \bbeta - \hat \bbeta_{-i}\Vert \leq \frac{\vert\ell'(y_i \x_i^\T \hat \bbeta)\vert}{n\lambda}\Vert\x_i\Vert=O(p^{-\frac{1}{2}}).
		\end{equation}
		Also, since 
		\begin{align*}
		    &\frac{1}{n}\sum_{j\neq i}\ell''(t_{-i,j})\left(y_j \x_j^\T \hat \bbeta-y_j\x_j^\T \hat \bbeta_{-i}\right)^2\\
		    =&\frac{\ell'(y_i \x_i^\T \hat \bbeta)^2 }{n^2}\x_i^\T\left(\lambda\I_p+\frac{1}{n}\sum_{j \neq i}\ell''(t_{-i,j})\x_j\x_j^\T\right)^{-1}\frac{1}{n}\sum_{j \neq i}\ell''(t_{-i,j})\x_j\x_j^\T\left(\lambda\I_p+\frac{1}{n}\sum_{j \neq i}\ell''(t_{-i,j})\x_j\x_j^\T\right)^{-1} \x_i\\
		    =&\frac{ \ell'(y_i \x_i^\T \hat \bbeta)^2 }{n^2}\x_i^\T\left[\left(\lambda\I_p+\frac{1}{n}\sum_{j \neq i}\ell''(t_{-i,j})\x_j\x_j^\T\right)^{-1}-\lambda\left(\lambda\I_p+\frac{1}{n}\sum_{j \neq i}\ell''(t_{-i,j})\x_j\x_j^\T\right)^{-2}\right] \x_i\\
		    \leq &\frac{ \ell'(y_i \x_i^\T \hat \bbeta)^2 }{n^2\lambda}\x_i^\T\x_i=O(p^{-1}),
		\end{align*}
		we get immediately by the interchangeability of $\x_j$ with $j\neq i$ that
		\begin{align*}
		     \left(y_j \x_j^\T \hat \bbeta- y_j \x_j^\T \hat \bbeta_{-i}\right)^2=O(p^{-1})
		\end{align*}
       We obtain thus by Taylor's expansion of $\ell'(\cdot)$ that
   \begin{align*}
      	\ell''(t_{-i,j})=\ell''(y_j\x_j^\T \hat \bbeta_{-i})+O(p^{-\frac{1}{2}}).
   \end{align*}
   Projecting \eqref{eq:dif-beta-beta_i} against $y_i \x_i$ we further deduce
   \begin{align*}
   y_i \x_i^\T \hat \bbeta-y_i\x_i^\T \hat \bbeta_{-i}=-\ell'(y_i \x_i^\T \hat \bbeta) \cdot \kappa_i+O(p^{-\frac{1}{2}}).
   \end{align*}
   with
   \begin{equation}
       \kappa_i= \frac1n \x_i^\T\Q_{-i}\x_i, \quad \Q_{-i}=\left(\lambda\I_p+\frac{1}{n}\sum_{j \neq i}\ell''(y_j\x_j^\T \hat \bbeta_{-i})\x_j\x_j^\T\right)^{-1}.
   \end{equation}
   It is important to remark that all $\kappa_i$ are concentrated around the same value. Indeed, since $\Q_{-i}$ is independent of $\x_i$ with bounded norm,
   $$\frac{1}{n}\x_i^\T\Q_{-i}\x_i=\frac{1}{n}\tr(\Q_{-i}\C)+O(p^{-\frac{1}{2}})$$
   by standard concentration arguments.  We get then
   	\begin{align*}
   	\kappa_i-\kappa_j&=\frac{1}{n}\tr\left[(\Q_{-i}-\Q_{-j})\C\right]+O(p^{-\frac12})\\
   	&=\frac{1}{n^2}\tr\bigg\{\Q_{-i}\Big[\ell''(y_i\x_i^\T \hat \bbeta_{-j})\x_i\x_i^\T-\ell''(y_j\x_j^\T \hat \bbeta_{-i})\x_j\x_j^\T\\
   	&+\sum_{l \neq i,j}\big(\ell''(y_l\x_l^\T \hat \bbeta_{-j})-\ell''(y_l\x_l^\T \hat \bbeta_{-i})\big)\x_l\x_l^\T\Big]\Q_{-j}\C\bigg\}+O(p^{-\frac12}) = O(p^{-\frac{1}{2}})
   	\end{align*} where the second equality is obtained by using  the resolvent identity $\mathbf{A}^{-1} - \mathbf{B}^{-1} = \mathbf{A}^{-1} (\mathbf{B}-\mathbf{A})\mathbf{B}^{-1}$. This allows us to write
   	\begin{align*}
   	    \kappa_i=\kappa+O(p^{-\frac{1}{2}})
   	\end{align*}
   	for some $\kappa$ constant over $i$ that will be determined later.
   	
    We arrive therefore at the key relation \eqref{eq:leave-one-out prediction} that formally writes
    \begin{align}
    \label{eq:relation kappa}
        y_i\x_i^\T\hat\bbeta_{-i}=y_i\x_i^\T\hat\bbeta+\kappa \ell'(y_i\x_i^\T\hat\bbeta)+O(p^{-\frac{1}{2}})
    \end{align}
    or, alternatively the following (approximate) mapping form $y_i\x_i^\T\hat\bbeta_{-i}$ to $y_i\x_i^\T\hat\bbeta$:
   $$y_i\x_i^\T\hat\bbeta = g_{\kappa,\ell}\left(y_i\x_i^\T\hat\bbeta_{-i}\right)+O(p^{-\frac{1}{2}})$$
   where we recall  $t\mapsto g_{\kappa,\ell}(t)$ is the inverse mapping of $t\mapsto t+\kappa\ell'(t)$ as defined in \eqref{eq:def-g}. Plugging the above relation into \eqref{eq:lambda-beta} leads to 
   \begin{align*}
   \lambda\hat\bbeta =\frac{1}{n}\sum_{i=1}^nc_i \bmu+\frac{1}{n}\sum_{i=1}^n c_i\w_i=\frac{1}{n}\sum_{i=1}^n\tilde c_i \bmu+\frac{1}{n}\sum_{i=1}^n\tilde c_i\w_i+O_{\Vert\cdot\Vert}(p^{-\frac{1}{4}})
   \end{align*}
   with $c_i= -\ell'(y_i\x_i^\T\hat\bbeta)$ and 
   \begin{equation}
       \tilde c_i=h\left(y_i\x_i^\T\hat\bbeta_{-i}\right)=c_i+O(p^{-\frac{1}{2}})
   \end{equation}
   for $h(t)=\frac{g_{\kappa,\ell}(t)-t}{\kappa} = -\ell'(g_{\kappa, \ell}(t))$ per \eqref{eq:def-prox}, and $\w_i=y_i\x_i-\bmu\sim\mathcal{N}(\zeros_p,\C)$. Indeed, since $\tilde c_i=c_i+O(p^{-\frac{1}{2}})$, we have
   \begin{align*}
       \left\Vert\frac{1}{n}\sum_{i=1}^nc_i \bmu-\frac{1}{n}\sum_{i=1}^n\tilde c_i \bmu\right\Vert^2=\left(\frac{1}{n}\sum_{i=1}^n (c_i -\tilde c_i) \right)^2\Vert\bmu\Vert^2=O(p^{-1})
   \end{align*}
   with $\| \bmu \| = O(1)$ under Assumption~\ref{ass:growth-rate} and 
   \begin{align*}
       \left\Vert\frac{1}{n}\sum_{i=1}^nc_i \w_i-\frac{1}{n}\sum_{i=1}^n\tilde c_i \w_i\right\Vert^2=\frac{1}{n^2 }\sum_{i,j=1}^n(c_i-\tilde c_i)(c_j-\tilde c_j)\w_i^\T\w_j=O(p^{-\frac{1}{2}})
   \end{align*}
   as $\w_i^\T\w_i=O(p)$ and $\w_i^\T\w_j=O(\sqrt p)$.
   
   As laid out  in ``Main idea and key steps'', in order to derive the asymptotic distribution of $\hat\bbeta$, it now remains to demonstrate \eqref{eq:expectation ciwi}~-~\eqref{eq:var r}. We will start by showing \eqref{eq:E hr}~-~\eqref{eq:var r}. Since \eqref{eq:E hr}, \eqref{eq:E squre hr}, \eqref{eq:cov hr r} and \eqref{eq:var r} can be obtained with similar reasoning, we give only the detailed derivation for  \eqref{eq:E hr}. Firstly, using again the fact that
    \begin{align*}
   \frac{1}{n}\sum_{i=1}^n c_i=\frac{1}{n}\sum_{i=1}^n \tilde c_i+O(p^{-\frac{1}{2}})
   \end{align*}
   we have
   \begin{equation}
    \frac{1}{n}\sum_{i=1}^n \tilde c_i \overset{(A)}{=} \frac{1}{n}\sum_{i=1}^n \E\left[h(y_i\x_i^\T\hat\bbeta_{-i})~\vert~\hat\bbeta_{-i} \right]+O(p^{-\frac{1}{4}}) \overset{(B)}{=} \E\left[h(r) \right]+O(p^{-\frac{1}{4}}).
    \end{equation}
    where for $(B)$ we recall that $\tilde c_i= h(y_i\x_i^\T\hat\bbeta_{-i})$ and use the fact that, conditioned on $\hat\bbeta_{-i}$, we have $y_i\x_i^\T\hat\bbeta_{-i}\cd \mathcal{N}(\bmu^\T\hat\bbeta_{-i},\hat\bbeta_{-i}^\T\C\hat\bbeta_{-i})$ (since $\hat\bbeta_{-i}$ is by definition, independent of $y_i\x_i$), so that $\E [h(y_i\x_i^\T\hat\bbeta_{-i})~\vert~\hat\bbeta_{-i}]=\E\left[h(r) \right]+O(p^{-1/2})$ for $r\sim\mathcal{N} (\bmu^\T\hat\bbeta,\hat\bbeta^\T\C\hat\bbeta )$ with  $\Vert \hat\bbeta-\hat\bbeta_{-i}\Vert=O(p^{-1/2})$ per \eqref{eq:bound-norm-dif-beta}. 
    To establish $(A)$, we need to control the difference
    \begin{equation}
        \frac{1}{n}\sum_{i=1}^nh(y_i\x_i^\T\hat\bbeta_{-i}) - \frac{1}{n}\sum_{i=1}^n \E\left[h(y_i\x_i^\T\hat\bbeta_{-i})~\vert~\hat\bbeta_{-i} \right]
    \end{equation}
    by evaluating, for example, its second moment as:
    	 \begin{align*}
    	&\E\left[\left(\frac{1}{n}\sum_{i=1}^nh(y_i\x_i^\T\hat\bbeta_{-i})-\frac{1}{n}\sum_{i=1}^n \E\left[h(y_i\x_i^\T\hat\bbeta_{-i})~\big\vert~\hat\bbeta_{-i} \right]\right)^2\right]\\
    	=&\frac{1}{n^2}\sum_{i=1}^{n}\sum_{j \neq i}\E\bigg[\left(h(y_i\x_i^\T\hat\bbeta_{-i})-\E\left[h(y_i\x_i^\T\hat\bbeta_{-i})~\big\vert~\hat\bbeta_{-i} \right]\right)\left(h(y_j\x_j^\T\hat\bbeta_{-j})-\E\left[h(y_j\x_j^\T\hat\bbeta_{-j})~\big\vert~\hat\bbeta_{-j} \right]\right)\bigg]\\
    	&+\frac{1}{n^2}\sum_{i=1}^{n}\E\left[\left(h(y_i\x_i^\T\hat\bbeta_{-i})-\E\left[h(y_i\x_i^\T\hat\bbeta_{-i})~\big\vert~\hat\bbeta_{-i} \right]\right)^2\right]\\
    	=&\frac{1}{n^2}\sum_{i=1}^{n}\sum_{j \neq i}\E\bigg[\left(h(y_i\x_i^\T\hat\bbeta_{-ij})-\E\left[h(y_i\x_i^\T\hat\bbeta_{-ij})~\big\vert~\hat\bbeta_{-ij} \right]\right)\left(h(y_j\x_j^\T\hat\bbeta_{-ij})-\E\left[h(y_j\x_j^\T\hat\bbeta_{-ij})~\big\vert~\hat\bbeta_{-ij} \right]\right)\bigg]\\
    	&+O(p^{-\frac{1}{2}}) = O(p^{-\frac{1}{2}})
    	\end{align*} where $\hat\bbeta_{-ij}$ stands for the solution of \eqref{eq:opt-origin} without the $i$-th and $j$-th training samples. Here, the ``leave-one-out''argument is applied to obtain $\hat\bbeta_{-ij}$ that is (i) independent of $(\x_i,y_i)$ and $(\x_j, y_j)$ so as to facilitate the treatment of the conditional expectation and (ii) is asymptotically close to $\hat \bbeta_{-i}, \hat \bbeta_{-i}$ and $\hat \bbeta$ per \eqref{eq:bound-norm-dif-beta}.  Ultimately, we notice that for $i \neq j$
    \begin{align*}
    \E\left[h(y_i\x_i^\T\hat\bbeta_{-ij})h(y_j\x_j^\T\hat\bbeta_{-ij})~\big\vert~ \hat\bbeta_{-ij} \right]= \E\left[h(y_i\x_i^\T\hat\bbeta_{-ij})~\big\vert~ \hat\bbeta_{-ij} \right]  \E\left[h(y_j\x_j^\T\hat\bbeta_{-ij})~\big\vert~ \hat\bbeta_{-ij} \right],
    \end{align*}
    which gives the final line. To obtain  \eqref{eq:E squre hr}, \eqref{eq:cov hr r} and \eqref{eq:var r}, it suffices to adapt  the above reasoning  respectively for $\frac{1}{n}\sum_{i=1}^nc_i^2$, $ \frac{1}{n}\sum_{i=1}^n c_i\left(r_i-\sum_{i=1}^n r_i\right)$  and $\frac{1}{n}\sum_{i=1}^n \left(r_i-\sum_{i=1}^n r_i\right)^2$.

    \smallskip
    Now we turn to the discussion of  \eqref{eq:expectation ciwi}. Notice first that, conditioning on $\hat\bbeta_{-i}$, $$\tilde c_i=h\left(\bmu^\T\hat\bbeta_{-i}+\hat \bbeta_{-i}^\T\w_i\right)$$ is a deterministic function of $\hat\bbeta_{-i}^\T\w_i \sim \mathcal N(\mathbf{0}_p, \hat \bbeta_{-i}^\T \C \hat \bbeta_{-i})$. Further note that we can decompose $\w_i$ as 
    \begin{equation}
        \w_i = \frac{\hat\bbeta_{-i}^\T\w_i}{\hat\bbeta_{-i}^\T\C\hat\bbeta_{-i}}\C\hat\bbeta_{-i} + \left(\w_i-\frac{\hat\bbeta_{-i}^\T\w_i}{\hat\bbeta_{-i}^\T\C\hat\bbeta_{-i}}\C\hat\bbeta_{-i} \right)
    \end{equation}
    with the second term uncorrelated and thus independent of $\hat\bbeta_{-i}^\T\w_i$ (since uncorrelated joint Gaussian variables are independent) as
    $$\E\left[\hat\bbeta_{-i}^\T\w_i\left(\w_i-\frac{\hat\bbeta_{-i}^\T\w_i}{\hat\bbeta_{-i}^\T\C\hat\bbeta_{-i}}\C\hat\bbeta_{-i}\right)~\Bigg\vert~\hat\bbeta_{-i}\right]=\zeros_p.$$ We get thus
    \begin{align*}
       \E[\tilde c_i\w_i~\vert~\hat\bbeta_{-i}]&=\E\left[\tilde c_i\frac{\hat\bbeta_{-i}^\T\w_i}{\hat\bbeta_{-i}^\T\C\hat\bbeta_{-i}}\C\hat\bbeta_{-i}~\Bigg\vert~\hat\bbeta_{-i}\right]+\E\left[\tilde c_i\left(\w_i-\frac{\hat\bbeta_{-i}^\T\w_i}{\hat\bbeta_{-i}^\T\C\hat\bbeta_{-i}}\C\hat\bbeta_{-i}\right)~\Bigg\vert~\hat\bbeta_{-i}\right]\\
       &=\frac{\E[\tilde c_i\w_i^\T\hat\bbeta_{-i}~\vert~\hat\bbeta_{-i}]}{\hat\bbeta_{-i}^\T\C\hat\bbeta_{-i}}\C\hat\bbeta_{-i}+\E[\tilde c_i~\vert~\hat\bbeta_{-i}]\E\left[\w_i-\frac{\hat\bbeta_{-i}^\T\w_i}{\hat\bbeta_{-i}^\T\C\hat\bbeta_{-i}}\C\hat\bbeta_{-i}~\Bigg\vert~\hat\bbeta_{-i}\right]\\
       &=\frac{\E[\tilde c_i\w_i^\T\hat\bbeta_{-i}~\vert~\hat\bbeta_{-i}]}{\hat\bbeta_{-i}^\T\C\hat\bbeta_{-i}}\C\hat\bbeta_{-i}.
    \end{align*}

    \smallskip
    From the already established \eqref{eq:expectation ciwi} and \eqref{eq:cov hr r} and the fact that $\Vert\hat\bbeta-\hat\bbeta_{-i}\Vert=O(p^{-\frac{1}{2}})$, it is easy to observe that
    \begin{align*}
        \frac{1}{n}\sum_{i=1}^n\left(\tilde c_i\w_i-\E[\tilde c_i\w_i~\vert~\hat\bbeta_{-i}]\right)&=\frac{1}{n}\sum_{i=1}^n\tilde c_i\w_i+\theta\C\hat\bbeta+O_{\Vert\cdot\Vert}(p^{-\frac{1}{2}})\\ 
        &=\C^{\frac{1}{2}}\left(\frac{1}{n}\sum_{i=1}^n\tilde c_i\z_i+\theta\C^{\frac{1}{2}}\hat\bbeta\right)+O_{\Vert\cdot\Vert}(p^{-\frac{1}{2}})
    \end{align*}
    with $\z_i \sim \mathcal N(\zeros_p, \I_p)$ and $\theta=-\cov[h(r),r]/\var[r]$. To obtain \eqref{eq:v}, it suffices to establish
    \begin{equation}\label{eq:dif-u}
        \left\Vert\frac{1}{n}\sum_{i=1}^n\tilde c_i\z_i+\theta\C^{\frac{1}{2}}\hat\bbeta-\gamma\u\right\Vert=o(1)
    \end{equation}
    for $\gamma=\sqrt{\frac{p\E[h^2(r)]}{n}}$ and $\u\sim\mathcal{N}(\zeros_p,\I_p/p)$. Since $\u\sim\mathcal{N}(\zeros_p,\I_p/p)$, we have $\v^\T\u\sim\mathcal{N}(0,1/p)$ for any deterministic vector $\v\in\RR^p$ of unit norm. As such, it suffices to show that for any deterministic unitary matrix $\V=[\v_1,\ldots,\v_p]\in\RR^{p\times p}$,
    $$\sum_{d=1}^p\left(\frac{1}{n}\sum_{i=1}^n\tilde c_i\v_d^\T\z_i+\theta\v_d^\T\C^{\frac{1}{2}}\hat\bbeta-\gamma s_d\right)^2=o(1)$$
    for some \emph{independent} $s_1,\ldots,s_p\sim\mathcal{N}(0,1/p)$ (due to the orthogonality of the $\v_d$'s). Let $\mathcal{A}$ stand for the set of $d\in\{1,\ldots,p\}$ such that $\v_d^\T\C^{\frac{1}{2}}\hat\bbeta=O(p^{-1/2})$ and $\mathcal{A}^{\rm c}$ its complementary set. We start by observing that, for $d\in\mathcal{A}^{\rm c}$,
    \begin{align}
    \label{eq:set Ac}
        \left(\frac{1}{n}\sum_{i=1}^n\tilde c_i\v_d^\T\z_i+\theta\v_d^\T\C^{\frac{1}{2}}\hat\bbeta-\gamma s_d\right)^2=o\left((\v_d^\T\C^{\frac{1}{2}}\hat\bbeta)^2\right).
    \end{align}  
    Let us write
    \begin{equation}\label{eq:def-x_i_d}
        \x_i^{\{d\}}=y_i\bmu+y_i\C^{\frac{1}{2}}\sum_{d'\neq d}(\v_{d'}^\T\z_i)\v_{d'},
    \end{equation}
    and denote by $\hat\bbeta^{\{d(i)\}}$ the solution to \eqref{eq:opt-origin} with the $i$-th training sample $(\x_i,y_i)$ replaced by $(\x_i^{\{d\}},y_i)$. As $\x_i^{\{d\}}$ is independent of $\v_{d}^\T\z_i$ (again by orthogonality), so is $\hat\bbeta^{\{d(i)\}}$. Similarly to the discussion on the difference between $\hat\bbeta$ and $\hat\bbeta_{-i}$, we have for $\hat\bbeta$ and $\hat\bbeta^{\{d(i)\}}$ that
    \begin{align*}
		\lambda \left(\hat \bbeta - \hat \bbeta^{\{d(i)\}}\right) = &\frac1n \sum_{j \neq i} -\left(\ell'(y_j \x_j^\T \hat \bbeta ) - \ell'\left(y_j\x_j^\T \hat \bbeta^{\{d(i)\}}\right) \right) y_j\x_j - \frac1n\ell'(y_i \x_i^\T \hat \bbeta) y_i \x_i\\
		&+\frac1n\ell'\left(y_i \x_i^{\{d\}\T} \hat\bbeta^{\{d(i)\}}\right) y_i \x_i^{\{d\}}\\
		=& \frac1n \sum_{j \neq i} -\ell''\left(t_j^{\{d(i)\}}\right)\x_j\x_j^\T \left(\hat \bbeta-\hat\bbeta^{\{d(i)\}}\right)- \frac1n \ell''\left(t_i^{\{d(i)\}}\right)\x_i^{\{d\}}\x_i^{\{d\}\T} \left( \hat \bbeta-\hat\bbeta^{\{d(i)\}}\right)\\
		&-\frac1n \ell''\left(t_i^{\{d(i)\}}\right)(\v_{d}^\T\z_i)(\v_{d}^\T\C^{\frac{1}{2}}\hat \bbeta)\x_i^{\{d\}}-\frac1n\ell'(y_i \x_i^\T \hat\bbeta)(\v_{d}^\T\z_i)\C^{\frac{1}{2}}\v_{d}
		\end{align*}
		where $\ell''(t_j^{\{d(i)\}})\geq 0$ for some $t_j^{\{d(i)\}}$ between $y_j \x_j^\T \hat \bbeta$ and $y_j\x_j^\T \hat \bbeta^{\{d(i)\}}$ by the mean value theorem and $\ell''(t_i^{\{d(i)\}})\geq 0$ for some $t_i^{\{d(i)\}}$ between $y_i \x_i^\T \hat \bbeta$ and $y_i \x_i^{\{d\}\T} \hat\bbeta^{\{d(i)\}}$.
        Rearranging the last equation gives
        \begin{align*}
            \hat \bbeta - \hat \bbeta^{\{d(i)\}}=&\left[\lambda\I_p+\frac1n \sum_{j \neq i} \ell''\left(t_j^{\{d(i)\}}\right)\x_j\x_j^\T+ \frac1n \ell''\left(t_i^{\{d(i)\}}\right)\x_i^{\{d\}}\x_i^{\{d\}\T} \right]^{-1}\\
            &\left(-\frac1n \ell''\left(t_i^{\{d(i)\}}\right)(\v_{d}^\T\z_i)(\v_{d}^\T\C^{\frac{1}{2}}\hat \bbeta)\x_i^{\{d\}}-\frac1n\ell'(y_i \x_i^\T \hat\bbeta)(\v_{d}^\T\z_i)\C^{\frac{1}{2}}\v_{d}\right),
        \end{align*}
        from which we deduce that 
        \begin{align}
        \label{eq:beta^d(i)}
            \Vert\hat \bbeta - \hat \bbeta^{\{d(i)\}}\Vert=O(\v_d^\T\C^{\frac{1}{2}}\hat\bbeta/\sqrt p).
        \end{align}
       We then denote by $\bbeta^{\{d(i)\}}_{-j}$ the leave-one-out version of $\hat \bbeta^{\{d(i)\}}$, which is, by definition, independent of both $\v_d^\T\z_i$ and $\v_d^\T\z_j$, and satisfies $\Vert\hat \bbeta_{-j} - \hat \bbeta^{\{d(i)\}}_{-j}\Vert=O(\v_d^\T\C^{\frac{1}{2}}\hat\bbeta/\sqrt p)$. 
    Since $\E[\tilde c_i\v_d^\T\z_i+\theta\v_d^\T\C^{\frac{1}{2}}\hat\bbeta]=O(p^{-1/2})$ (by \eqref{eq:expectation ciwi}, \eqref{eq:cov hr r}~and~\eqref{eq:var r}), we have that
    \begin{align}
        \left(\frac{1}{n}\sum_{i=1}^n\tilde c_i\v_d^\T\z_i+\theta\v_d^\T\C^{\frac{1}{2}}\hat\bbeta\right)^2=O(\v_d^\T\C^{\frac{1}{2}}\hat\bbeta/\sqrt p)\label{eq:civdzi Ac}
    \end{align}
    from
    \begin{align*}
        &\E\left[\left(\frac{1}{n}\sum_{i=1}^n\tilde c_i\v_d^\T\z_i+\theta\v_d^\T\C^{\frac{1}{2}}\hat\bbeta\right)^2\right]\\
        =&\frac{1}{n^2}\sum_{i\neq j}\E\left[\left(\tilde c_i\v_d^\T\z_i+\theta\v_d^\T\C^{\frac{1}{2}}\hat\bbeta\right)\left(\tilde c_j\v_d^\T\z_j+\theta\v_d^\T\C^{\frac{1}{2}}\hat\bbeta\right)\right]+O(p^{-1})\\
        =&\frac{1}{n^2}\sum_{i\neq j}\E\left[\left(h(y_i\x_i^\T\hat \bbeta_{-i}^{\{d(j)\}})\v_d^\T\z_i++\theta\v_d^\T\C^{\frac{1}{2}}\hat\bbeta^{\{d(ij)\}}\right)\left(h(y_j\x_j^\T\hat \bbeta_{-j}^{\{d(i)\}})\v_d^\T\z_j+\theta\v_d^\T\C^{\frac{1}{2}}\hat\bbeta^{\{d(ij)\}}\right)\right]\\
        &+O(\v_d^\T\C^{\frac{1}{2}}\hat\bbeta/\sqrt p) = O(\v_d^\T\C^{\frac{1}{2}}\hat\bbeta/\sqrt p).
    \end{align*} As $s_d\sim\mathcal{N}(0,1/p)=o\left((\v_d^\T\C^{\frac{1}{2}}\hat\bbeta)^2\right)$, we get directly that for $d \in \mathcal{A}^{\rm c}$,
    \begin{equation}
        \left(\frac{1}{n}\sum_{i=1}^n\tilde c_i\v_d^\T\z_i+\theta\v_d^\T\C^{\frac{1}{2}}\hat\bbeta-\gamma s_d\right)^2=O(\v_d^\T\C^{\frac{1}{2}}\hat\bbeta/\sqrt p)
    \end{equation}

      We proceed to demonstrate that, for $d\in\mathcal{A}$,
      \begin{align}
          \label{eq:set A}
             \left(\frac{1}{n}\sum_{i=1}^n\tilde c_i\v_d^\T\z_i+\theta\v_d^\T\C^{\frac{1}{2}}\hat\bbeta-\gamma s_d\right)^2=o(p^{-1}).
      \end{align}
   Let $\hat\bbeta^{\{d\}}$ stand for the solution to \eqref{eq:opt-origin} with all $(\x_i,y_i)$ replaced by $(\x_i^{\{d\}},y_i)$, for $\x_i^{\{ d\}}$ defined in \eqref{eq:def-x_i_d}. Taking again the the difference of the stationary point equations \eqref{eq:lambda-beta}, we have
     \begin{align*}
		\lambda \left(\hat \bbeta - \hat \bbeta^{\{d\}}\right) = &-\frac1n \sum_{i=1}^n \left(\ell'(y_i \x_i^\T \hat \bbeta ) - \ell'\left(y_i\x_i^{\{d\}\T} \hat \bbeta^{\{d\}}\right) \right) y_i\x_i- \frac1n \sum_{i=1}^n\frac1n\ell'(y_i \x_i^{\{d\}\T} \hat \bbeta^{\{d\}})(\v_{d}^\T\z_i)\C^{\frac{1}{2}}\v_{d}\\
		=&-\frac1n \sum_{i=1}^n  \ell''\left(t_i^{\{d\}}\right)\x_i\x_i^{\T} \left( \hat \bbeta-\hat\bbeta^{\{d\}}\right)-\frac1n\sum_{i=1}^n  \ell''\left(t_i^{\{d\}}\right)(\v_{d}^\T\z_i)(\v_{d}^\T\C^{\frac{1}{2}}\hat \bbeta)\x_i\\
		&-\frac1n\sum_{i=1}^n \ell'(y_i \x_i^{\{d\}\T} \hat \bbeta^{\{d\}})(\v_{d}^\T\z_i)\C^{\frac{1}{2}}\v_{d}
		\end{align*}
		where $\ell''\left(t_i^{\{d\}}\right)\geq0$ for some $t_i^{\{d\}}$ between $y_i \x_i^\T \hat \bbeta$ and $y_i\x_i^{\{d\}\T} \hat \bbeta^{\{d\}}$ by the mean value theorem. We are going to show from the above equation that 
		$$\Vert\hat \bbeta - \hat \bbeta^{\{d\}}\Vert=O(p^{-\frac{1}{2}}).$$
		Since $\ell'(y_1 \x_1^{\{d\}\T} \hat \bbeta^{\{d\}}),\ldots,\ell'(y_n \x_n^{\{d\}\T} \hat \bbeta^{\{d\}})$ are independent of $\v_{d}^\T\z_1,\ldots,\v_{d}^\T\z_n$ (again by orthogonality), we have
		\begin{align*}
		    \E\left[\left(\frac1n\sum_{i=1}^n \ell'(y_i \x_i^{\{d\}\T} \hat \bbeta^{\{d\}})(\v_{d}^\T\z_i)\right)^2\right]=&\frac{1}{n^2}\sum_{i,j=1}^n\E\left[ \ell'(y_i \x_i^{\{d\}\T} \hat \bbeta^{\{d\}}) \ell'(y_j \x_j^{\{d\}\T} \hat \bbeta^{\{d\}})\right]\E\left[ \v_{d}^\T\z_i \v_{d}^\T\z_j\right]\\
		    =&\frac{1}{n^2}\sum_{i}^n\E\left[ \ell'(y_i \x_i^{\{d\}\T} \hat \bbeta^{\{d\}}) \ell'(y_i \x_i^{\{d\}\T} \hat \bbeta^{\{d\}})\right]\E\left[ \v_{d}^\T\z_i \v_{d}^\T\z_i\right]\\
		    =&O(p^{-1}).
		\end{align*}
	    As $t_i^{\{d\}}$ is between $y_i \x_i^\T \hat \bbeta$ and $y_i\x_i^{\{d\}\T} \hat \bbeta^{\{d\}}$, and 
		\begin{align*}
		    y_i \x_i^\T \hat \bbeta&=g_{\kappa,\ell}(y_i \x_i^\T \hat \bbeta_{-i})+O(p^{-\frac{1}{2}})\\
		    y_i \x_i^{\{d\}\T} \hat \bbeta^{\{d\}}&=g_{\kappa,\ell}(y_i \x_i^{\{d\}\T} \hat \bbeta_{-i}^{\{d\}})+O(p^{-\frac{1}{2}})
		\end{align*}
		where $\hat \bbeta^{\{d\}}_{-i}$ stands for the leave-one-out version of $\hat \bbeta^{\{d\}}$,
		we remark that
		\begin{align*}
		    t_i^{\{d\}}=g_{\kappa,\ell}(y_i \x_i^{\{d\}\T} \hat \bbeta_{-i}^{\{d\}})+O\left(\max\{\Vert\hat \bbeta_{-i}-\hat \bbeta_{-i}^{\{d\}}\Vert,p^{-\frac{1}{2}}\}\right),
		\end{align*}
		and consequently
		\begin{align*}
		    \ell''\left(t_i^{\{d\}}\right)=\ell''\left(g_{\kappa,\ell}(y_i \x_i^{\{d\}\T} \hat \bbeta_{-i}^{\{d\}})\right)+O\left(\max\{\Vert\hat \bbeta-\hat \bbeta^{\{d\}}\Vert,p^{-\frac{1}{2}}\}\right).
		\end{align*}
	Using again the fact that  $\ell'(y_1 \x_1^{\{d\}\T} \hat \bbeta^{\{d\}}),\ldots,\ell'(y_n \x_n^{\{d\}\T} \hat \bbeta^{\{d\}})$ are independent of $\v_{d}^\T\z_1,\ldots,\v_{d}^\T\z_n$, we have that
	\begin{align*}
	     &\E\left[\left\Vert\frac1n\sum_{i=1}^n\ell''\left(g_{\kappa,\ell}(y_i \x_i^{\{d\}\T} \hat \bbeta_{-i}^{\{d\}})\right)(\v_{d}^\T\z_i)\x_i\right\Vert^2\right]\\
	     =&\frac{1}{n^2}\sum_{i,j=1}^n\E\left[\ell''\left(g_{\kappa,\ell}(y_i \x_i^{\{d\}\T} \hat \bbeta_{-i}^{\{d\}})\right)\ell''\left(g_{\kappa,\ell}(y_j \x_j^{\{d\}\T} \hat \bbeta_{-j}^{\{d\}})\right)(\v_{d}^\T\z_i)(\v_{d}^\T\z_j)\x_i^\T\x_j\right]\\
	     =&\frac{1}{n^2}\sum_{i,j=1}^n\E\left[\ell''\left(g_{\kappa,\ell}(y_i \x_i^{\{d\}\T} \hat \bbeta_{-i}^{\{d\}})\right)\ell''\left(g_{\kappa,\ell}(y_j \x_j^{\{d\}\T} \hat \bbeta_{-j}^{\{d\}})\right)\x_i^{\{d\}\T}\x_j^{\{d\}}\right]\E\left[(\v_{d}^\T\z_i)(\v_{d}^\T\z_j)\right]+O(1)\\
	     =&\frac{1}{n^2}\sum_{i=1}^n\E\left[\ell''\left(g_{\kappa,\ell}(y_i \x_i^{\{d\}\T} \hat \bbeta_{-i}^{\{d\}})\right)\ell''\left(g_{\kappa,\ell}(y_i \x_i^{\{d\}\T} \hat \bbeta_{-i}^{\{d\}})\right)\x_i^{\{d\}\T}\x_i^{\{d\}}\right]\E\left[(\v_{d}^\T\z_i)(\v_{d}^\T\z_i)\right]+O(1)=O(1)
	\end{align*}
	Therefore, for $d\in\mathcal{A}$, we observe 
		\begin{align}
		\label{eq:beta^d}
		    \left\Vert\hat \bbeta - \hat \bbeta^{\{d\}}\right\Vert=O(p^{-\frac{1}{2}})
		\end{align}
      from 
	\begin{align*}
	    \hat \bbeta^{\{d\}}-\hat \bbeta =&\left(\lambda\I_p+\frac1n \sum_{i=1}^n  \ell''\left(t_i^{\{d\}}\right)\x_i\x_i^\T\right)^{-1}\frac{1}{n}\sum_{i=1}^n\Bigg( \ell''\left(t_i^{\{d\}}\right)(\v_{d}^\T\z_i)(\v_{d}^\T\C^{\frac{1}{2}}\hat \bbeta)\x_i\\
	    &+\ell'(y_i \x_i^{\{d\}\T} \hat \bbeta^{\{d\}})(\v_{d}^\T\z_i)\C^{\frac{1}{2}}\v_{d}\Bigg)\\
	    =&\left(\lambda\I_p+\frac1n \sum_{i=1}^n  \ell''\left(t_i^{\{d\}}\right)\x_i\x_i^\T\right)^{-1}\frac{1}{n}\sum_{i=1}^n\Bigg( \ell''\left(g_{\kappa,\ell}(y_i \x_i^{\{d\}\T} \hat \bbeta_{-i}^{\{d\}})\right)(\v_{d}^\T\z_i)(\v_{d}^\T\C^{\frac{1}{2}}\hat \bbeta)\x_i\\
	    &+\ell'(y_i \x_i^{\{d\}\T} \hat \bbeta^{\{d\}})(\v_{d}^\T\z_i)\C^{\frac{1}{2}}\v_{d}\Bigg)+O_{\Vert\cdot\Vert}\left(\max\{\Vert\hat \bbeta-\hat \bbeta^{\{d\}}\Vert,O(p^{-\frac{1}{2}})\}\right).
		\end{align*}

	    Our next step is to show that 
	    \begin{align}
	    \label{eq:civdzi}
	        \frac{1}{n}\sum_{i=1}^n\tilde c_i\v_d^\T\z_i+\theta\v_d^\T\C^{\frac{1}{2}}\hat\bbeta-\frac{1}{n}\sum_{i=1}^n h(y_i \x_i^{\{d\}\T} \hat \bbeta^{\{d\}}_{-i})\v_d^\T\z_i=O(p^{-\frac{3}{4}})
	    \end{align}
		where $\x_i^{\{d\}\T} \hat \bbeta^{\{d\}}_{-i}$ is independent of all $\v_d^\T\z_j$, $j\in\{1,\ldots,n\}$ and $h(y_i \x_i^{\{d\}\T} \hat \bbeta^{\{d\}}_{-i})=\tilde c_i+O(p^{-\frac{1}{2}})$ since $\Vert\hat \bbeta_{-i} - \hat \bbeta_{-i}^{\{d\}}\Vert=O(p^{-\frac{1}{2}})$.
		To this aim, we begin by rewriting $\tilde c_i\v_d^\T\z_i$ as
		\begin{align*}
		    \tilde c_i\v_d^\T\z_i=h(y_i\x_i^{\T}\hat\bbeta_{-i}) \v_d^\T\z_i
		    =\left[h(y_i\x_i^{\{d\}\T}\hat\bbeta_{-i})+h'(y_i\x_i^{\{d\}\T}\hat\bbeta_{-i})(\v_d^\T\C^{\frac{1}{2}}\hat\bbeta)(\v_d^\T\z_i)\right]\v_d^\T\z_i+O(p^{-1}).
		\end{align*}
		Following the same reasoning for establishing \eqref{eq:E hr}, we have
		\begin{align*}
		   \frac{1}{n}\sum_{i=1}^n h'(y_i\x_i^{\{d\}\T}\hat\bbeta_{-i})(\v_d^\T\z_i)^2=&\E\left[h'(y_i\x_i^{\{d\}\T}\hat\bbeta_{-i})\right]\E\left[(\v_d^\T\z_i)^2\right]=\E\left[h'(r)\right]+O(p^{-\frac{1}{2}})\\
		   =&\cov\left[h(r),r\right]/\var[r]=\theta+O(p^{-\frac{1}{4}})
		\end{align*}
		which leads to
		\begin{align*}
		     \frac{1}{n}\sum_{i=1}^n\tilde c_i\v_d^\T\z_i=\frac{1}{n}\sum_{i=1}^n h(y_i\x_i^{\{d\}\T}\hat\bbeta_{-i})\v_d^\T\z_i+\theta\v_d^\T\C^{\frac{1}{2}}\hat\bbeta+O(p^{-\frac{3}{4}}).
		\end{align*}
		Therefore, it suffices to show 
		\begin{align*}
		    \frac{1}{n}\sum_{i=1}^n \left(h(y_i\x_i^{\{d\}\T}\hat\bbeta_{-i})-h(y_i \x_i^{\{d\}\T} \hat \bbeta^{\{d\}}_{-i})\right)\v_d^\T\z_i=O(p^{-\frac{3}{4}})
		\end{align*}
        in order to prove \eqref{eq:civdzi}. Reapplying the approximation and independence properties of $\hat\bbeta^{\{d\}}$ and $\hat\bbeta^{\{d(i)\}}$, we have
		\begin{align*}
		    &\E\left[\left(h(y_i\x_i^{\{d\}\T}\hat\bbeta_{-i})-h(y_i \x_i^{\{d\}\T} \hat \bbeta^{\{d\}}_{-i})\right)(\v_d^\T\z_i)\left(h(y_j\x_j^{\{d\}\T}\hat\bbeta_{-j})-h(y_j \x_j^{\{d\}\T} \hat \bbeta^{\{d\}}_{-j})\right)(\v_d^\T\z_i)\right]\\
		    =&\E\left[\left(h(y_i\x_i^{\{d\}\T}\hat\bbeta_{-i}^{\{d(j)\}})-h(y_i \x_i^{\{d\}\T} \hat \bbeta^{\{d\}}_{-i})\right)\left(h(y_j\x_j^{\{d\}\T}\hat\bbeta_{-j}^{\{d(i)\}})-h(y_j \x_j^{\{d\}\T} \hat \bbeta^{\{d\}}_{-j})\right)\right]\E[\v_d^\T\z_i]\E[\v_d^\T\z_j]+O(p^{-\frac{3}{2}})\\
		    =&O(p^{-\frac{3}{2}})
		\end{align*}
		for $i\neq j$. It follows straightforwardly that
		\begin{align*}
		    &\E\left[\left(\frac{1}{n}\sum_{i=1}^n \left(h(y_i\x_i^{\{d\}\T}\hat\bbeta_{-i})-h(y_i \x_i^{\{d\}\T} \hat \bbeta^{\{d\}}_{-i})\right)\v_d^\T\z_i\right)^2\right]\\
		    =&\frac{1}{n^2}\sum_{i=1}^n\sum_{j\neq i}\E\left[\left(h(y_i\x_i^{\{d\}\T}\hat\bbeta_{-i})-h(y_i \x_i^{\{d\}\T} \hat \bbeta^{\{d\}}_{-i})\right)(\v_d^\T\z_i)\left(h(y_j\x_j^{\{d\}\T}\hat\bbeta_{-j})-h(y_j \x_j^{\{d\}\T} \hat \bbeta^{\{d\}}_{-j})\right)(\v_d^\T\z_i)\right]\\
		    &+\frac{1}{n^2}\sum_{i=1}^n\E\left[\left(h(y_i\x_i^{\{d\}\T}\hat\bbeta_{-i})-h(y_i \x_i^{\{d\}\T} \hat \bbeta^{\{d\}}_{-i})\right)^2(\v_d^\T\z_i)^2\right] = O(p^{-\frac{3}{2}}).
		\end{align*}
	We get thus \eqref{eq:civdzi}. Again, as $h(y_1 \x_1^{\{d\}\T} \hat \bbeta^{\{d\}}_{-1}),\ldots,h(y_n \x_n^{\{d\}\T} \hat \bbeta^{\{d\}}_{-n})$ are independent of $\v_d^\T\z_1,\ldots,\v_d^\T\z_n$, conditioning on $h(y_1 \x_1^{\{d\}\T} \hat \bbeta^{\{d\}}_{-1}),\ldots,h(y_n \x_n^{\{d\}\T} \hat \bbeta^{\{d\}}_{-n})$, we have by the central limit theorem that
	$$\frac{1}{n}\sum_{i=1}^n h(y_i \x_i^{\{d\}\T} \hat \bbeta^{\{d\}}_{-i})\v_d^\T\z_i=\sqrt{\frac{p}{n}\sum_{i=1}h^2(y_i \x_i^{\{d\}\T} \hat \bbeta^{\{d\}}_{-i})} \cdot s_d+o(p^{-\frac{1}{2}})=\gamma s_d+o(p^{-\frac{1}{2}}) $$
	for some $s_d\sim\mathcal{N}(0,1/p)$. We obtain finally \eqref{eq:set A}.

    Combining \eqref{eq:set A} and \eqref{eq:set Ac}, we have
    \begin{align*}
        \sum_{d=1}^p\left(\frac{1}{n}\sum_{i=1}^n\tilde c_i\v_d^\T\z_i+\theta\v_d^\T\C^{\frac{1}{2}}\hat\bbeta-\gamma s_d\right)^2=&\sum_{d\in\mathcal{A}^{\rm c}}\left(\frac{1}{n}\sum_{i=1}^n\tilde c_i\v_d^\T\z_i+\theta\v_d^\T\C^{\frac{1}{2}}\hat\bbeta-\gamma s_d\right)^2\\
        &+\sum_{d\in\mathcal{A}}\left(\frac{1}{n}\sum_{i=1}^n\tilde c_i\v_d^\T\z_i+\theta\v_d^\T\C^{\frac{1}{2}}\hat\bbeta-\gamma s_d\right)^2\\
        =&o(\Vert\hat\bbeta\Vert)+o(1)=o(1),
    \end{align*}
    which completes the proof of \eqref{eq:v}.
    
    \medskip
    
    We return to the determination of $\kappa$. From the arguments demonstrated above, we remark that
    \begin{align*}
        \hat\bbeta&=\left(\lambda\I+\theta\C\right)^{-1}\left[\eta\bmu+\frac{1}{n}\sum_{i=1}^n(\tilde c_i\w_i-\E[\tilde c_i\w_i~\vert~\hat\bbeta_{-i}])\right]+o_{\Vert\cdot\Vert}(1)
    \end{align*}
    with $\eta=\E[h(r)]$. As $\x_j$ is independent of $\w_i$, we have 
    \begin{align*}
        \E\left[\x_j^\T(\tilde c_i\w_i-\E[\tilde c_i\w_i~\vert~\hat\bbeta_{-i}])\right]=0
    \end{align*}
    for $j\neq i$, by conditioning first on $\hat\bbeta_{-i}$ and $\x_j$. This leads to
    \begin{align*}
        \frac{1}{n}\sum_{j=1}^n\E\left[\x_j^\T\hat\bbeta\right] =&\E\left\{\frac{1}{n}\sum_{j=1}^n\x_j^\T\left(\lambda\I+\theta\C\right)^{-1}\left[\eta\bmu+\frac{1}{n}\sum_{i=1}^n(\tilde c_i\w_i-\E[\tilde c_i\w_i~\vert~\hat\bbeta_{-i}])\right]\right\}+o(1)\\
        =&\frac{1}{n}\sum_{j=1}^n\E\left\{\x_j^\T\left(\lambda\I+\theta\C\right)^{-1}\left[\eta\bmu+\frac{1}{n}(\tilde c_j\w_j-\E[\tilde c_j\w_j~\vert~\hat\bbeta_{-j}])\right]\right\}+o(1)\\
        =&\eta\bmu^\T\left(\lambda\I+\theta\C\right)^{-1}\bmu+\E[\tilde c_i]\tr\left(\lambda\I+\theta\C\right)^{-1}\C/n+o(1).
    \end{align*}
    Combining the above result with \eqref{eq:relation kappa}, we observe that
    \begin{align*}
        \E\left[\x_j^\T\hat\bbeta\right]=&\E\left[\x_j^\T\hat\bbeta_{-j}\right]-\kappa\E[\ell'(y_j \x_j^\T \hat\bbeta)]+o(1)\\
        =&\eta\bmu^\T\left(\lambda\I+\theta\C\right)^{-1}\bmu+\E[\tilde c_j] \cdot \kappa +o(1)\\
        =&\eta\bmu^\T\left(\lambda\I+\theta\C\right)^{-1}\bmu+\E[\tilde c_j] \cdot \tr\left(\lambda\I+\theta\C\right)^{-1}\C/n+o(1),
    \end{align*}
    which entails 
    \begin{align*}
        \kappa=\frac{1}{n}\tr\left(\lambda\I+\theta\C\right)^{-1}\C.
    \end{align*}
    \smallskip
    
    Before putting together all the arguments to retrieve the results of Theorem~\ref{theo:main}, we need to prove the positiveness of the key parameters $\theta,\eta,\gamma$ defined in \eqref{eq:theta eta gamma}.
    Note first that $\gamma$ is positive by definition.
    To see that $\theta$ is positive, recall from \eqref{eq:theta eta gamma} that
    \begin{align*}
	   \theta=\frac{-\cov[h(r),r]}{\var[r]}=\frac{\E[-h(m+z)z]}{\var[z]}
	\end{align*}
	with $r = m + z$ for some $z\sim\mathcal{N}(0,\sigma^2)$. Since $\kappa>0$ and $\ell(t)$ is a convex function, it is easy to see that $-h(m+z)=\ell'(g_{\kappa,\ell}(m+z))$ is an increasing function of $z$, and therefore $\E[-h(m+z)z]\geq0$ (since $-h(m+z)z + (-h(m-z)) (-z) = z (h(m-z) - h(m+z)) \geq 0$ for all $z \geq 0$),  entailing that $\theta\geq 0$. As for $\eta$, its positiveness must hold for any reasonable loss function $\ell$, otherwise the resulting classification error would be worse than random classification. In our investigation, the positiveness of $\eta$ is ensured by the condition $\ell'(0)<0$ in Assumption~\ref{ass:loss}. Indeed, if $\eta<0$, then $m<0$, and we get from the previously proven relation
	$$y_j \x_j^\T \hat \bbeta-y_j\x_j^\T \hat \bbeta_{-i}= \kappa c_i+O(p^{-\frac{1}{2}})$$ that
	\begin{equation}\label{eq:eta<0}
	    \frac{1}{n}\sum_{j=1}^n y_j \x_j^\T \hat \bbeta=m+\kappa\eta+O(p^{-\frac{1}{2}})<0
	\end{equation}
	with high probability for large $p$. Denote by $\mathcal{S}_+$ the set of index $i\in\{1,\ldots,n\}$ such that $y_i \x_i^\T \hat \bbeta\geq 0$, and $\mathcal{S}_-$ its complementary set, we then have, by the mean value theorem that
	\begin{align*}
	\sum_{i\in\mathcal{S}_+}\left[\ell(y_i \x_i^\T \hat \bbeta)-\ell(0)\right]&=\sum_{i\in\mathcal{S}_+}\ell'(a_i)y_i \x_i^\T \hat \bbeta>\ell'(0)\sum_{i\in\mathcal{S}_+}y_i \x_i^\T \hat\bbeta\\
	\sum_{i\in\mathcal{S}_-}\left[\ell(y_i \x_i^\T \hat \bbeta)-\ell(0)\right]&=\sum_{i\in\mathcal{S}_-}\ell'(a_i)y_i \x_i^\T \hat \bbeta>\ell'(0)\sum_{i\in\mathcal{S}_-}y_i \x_i^\T \hat\bbeta
	\end{align*}
	where $a_i$ is some value between $y_i \x_i^\T \hat \bbeta$ and $0$, and the inequalities are due to the fact that $\ell'(t)$ is an increasing function. We thus have $$\frac{1}{n}\sum_{i=1}^n\ell(y_i \x_i^\T \hat \bbeta)-\ell(0)>\ell'(0)\frac{1}{n}\sum_{i=1}^ny_i \x_i^\T \hat\bbeta.$$ Since $\ell'(0)<0$, we find that for $\eta < 0$ (so that \eqref{eq:eta<0} holds), $\frac{1}{n}\sum_{i=1}^n\ell(y_i \x_i^\T \hat \bbeta)+\frac{\lambda}{2}\Vert\hat\bbeta\Vert^2$ is (asymptotically) strictly greater than
	$\frac{1}{n}\sum_{i=1}^n\ell(y_i \x_i^\T\zeros_p)+\frac{\lambda}{2}\Vert\zeros_p\Vert^2$. As the objective function $\frac{1}{n}\sum_{i=1}^n\ell(y_i \x_i^\T  \bbeta)+\frac{\lambda}{2}\Vert\bbeta\Vert^2$ is supposed to reach its minimum at $\bbeta=\hat\bbeta$, we prove by contradiction that $\eta\geq 0$.
	
	\smallskip
	
	Ultimately, we note that the unregularized solution $\hat\bbeta(0)$ is retrieved as a limiting case of regularized solution $\hat\bbeta(\lambda)$ at $\lambda\to0$, if $\hat\bbeta(0)$ is unique and well-defined with bounded norm. Since
	\begin{align*}
	    \frac{1}{n}\sum_{i=1}^n\ell'(y_i\x_i^\T\hat\bbeta(\lambda))y_i\x_i+\lambda \hat\bbeta(\lambda)&=\zeros_p,\\
	    \frac{1}{n}\sum_{i=1}^n\ell'(y_i\x_i^\T\hat\bbeta(0)) y_i\x_i&=\zeros_p,
	\end{align*}
	we have
	\begin{align*}
	    \left(\frac{1}{n}\sum_{i=1}^n \ell''(t_{\lambda,i})\x_i\x_i^\T\right)(\hat\bbeta(0)-\hat\bbeta(\lambda))=\lambda \hat\bbeta(\lambda)
	\end{align*}
	where $\ell''(t_{\lambda,i})>0$ for some $t_{\lambda,i}$ between $y_i \x_i^\T \hat\bbeta(0)$ and $y_i\x_i^\T\hat\bbeta(\lambda)$. When $n/p> 1$, the smallest eigenvalue $q_{\rm min}$ of $\frac{1}{n}\sum_{i=1}^n \ell''(t_{\lambda,i})\x_i\x_i^\T$ is bounded away from zero with high probability \citep{bai2008limit}. We get therefore from 
	$$q_{\rm min}\Vert\hat\bbeta(0)-\hat\bbeta(\lambda)\Vert\leq \lambda\Vert\hat\bbeta(\lambda)\Vert$$ that
	$$\lim_{\lambda\to 0}\Vert\hat\bbeta(0)-\hat\bbeta(\lambda)\Vert\to 0.$$
   Recall that $\kappa$ defined in \eqref{eq:kappa} equals the deterministic limit of $\frac1n \tr\C\left(\frac{1}{n}\sum_{i=1}^n\ell''(y_i \x_i^\T \hat\bbeta(\lambda))\x_i\x_i^\T+\lambda\I\right)^{-1}$. We notice that when $n/p>1$, $\kappa$ is well defined at $\lambda=0$ and bounded away from zero. Since
    $$\theta=\frac{-\cov[h(r),r]}{\var[r]}=\frac{1}{\kappa}-\frac{\cov[g_{\kappa,\ell}(r),r]}{\kappa\var[r]},$$ 
    we remark that $\theta$ is also bounded away from zero at $\lambda=0$. The vector $\tilde\bbeta(0)$ is thus well-defined, with $\Vert\hat\bbeta(0)-\tilde\bbeta(0)\Vert=o(1)$.

\section{Proof of Theorem~\ref{theo:joint distribution}}

It can be summarized from the arguments in the proof of Theorem~\ref{theo:main} that
\begin{align*}
\frac{1}{n}\sum_{i=1}^n\tilde c_i\z_i+\theta\C^{\frac{1}{2}}\hat\bbeta=\gamma\u+o_{\Vert\cdot\Vert}(1).
\end{align*}
where constants $\gamma,\theta$ and random vector $\u\sim\mathcal{N}(\zeros_p,\I_p/p)$ are as understood in Theorem~\ref{theo:main}, with $$\tilde c_i=h(y_i\x_i\hat\bbeta_{-i})=c_i+o(1)$$
as defined in the proof of Theorem~\ref{theo:main}. Let us write $\tilde\c=[\tilde c_1,\ldots,\tilde c_n]$ and $\Z = [\z_1,\ldots,\z_n]$.  Under the notations of Theorem~\ref{theo:joint distribution}, we denote by $\tilde\c_{[k]}=[\tilde c_{[k]1},\ldots,\tilde c_{[k]n}]$ the vector $\tilde\c$ for $\hat\bbeta_{\ell_k}(\lambda_k)$,which we abbreviate to $\hat\bbeta_{[k]}$. The same is understood for $\theta_k$ and $\gamma_k$. For convenience, we focus on the pair $\left(\hat\bbeta_{[1]},\hat\bbeta_{[2]}\right)$ without loss of generality. As Theorem~\ref{theo:joint distribution} holds if 
\begin{align*}
    \begin{bmatrix}\Z\tilde\c_{[1]}/n\\\Z\tilde\c_{[2]}/n\end{bmatrix}+\begin{bmatrix}\theta_1\C^{\frac{1}{2}}\hat\bbeta_{[1]}\\\theta_2\C^{\frac{1}{2}}\hat\bbeta_{[1]}\end{bmatrix}+o_{\Vert\cdot\Vert}(1)=\begin{bmatrix}\gamma_1\u_{[1]}\\\gamma_2\u_{[2]}\end{bmatrix}\sim\mathcal{N}\left(\zeros_{2p},\begin{bmatrix}\gamma_1^2\I_p/p&\rho_{k_1k_2}\gamma_1\gamma_2\I_p/p\\\rho_{k_1k_2}\gamma_1\gamma_2\I_p/p&\gamma_2^2\I_p/p\end{bmatrix}\right),
\end{align*}
to prove Theorem~\ref{theo:joint distribution}, it suffices to show that for any deterministic unitary matrix $\V=[\v_1,\ldots,\v_{2p}]\in\RR^{2p\times 2p}$,
\begin{align}
    \sum_{d=1}^p\left(\v_d^\T\begin{bmatrix}\Z\tilde\c_{[1]}/n\\\Z\tilde\c_{[2]}/n\end{bmatrix}+\v_d^\T\begin{bmatrix}\theta_1\C^{\frac{1}{2}}\hat\bbeta_{[1]}\\\theta_2\C^{\frac{1}{2}}\hat\bbeta_{[1]}\end{bmatrix}-s_d\right)^2=o(1)\label{eq:proof joint distr}
\end{align}
for some $s_d\sim\mathcal{N}\left(0,\v_d^\T\begin{bmatrix}\gamma_1^2\I_p/p&\rho_{k_1k_2}\gamma_1\gamma_2\I_p/p\\\rho_{k_1k_2}\gamma_1\gamma_2\I_p/p&\gamma_2^2\I_p/p\end{bmatrix}\v_d\right)$, $d\in\{1,\ldots,p\}$. Let us write $\v_d=\begin{bmatrix}\v_{[1]d}\\\v_{[2]d}\end{bmatrix}$ where $\v_{[1]d},\v_{[2]d}\in\RR^p$, and denote by  $\mathcal{A}$  the set of $d\in\{1,\ldots,p\}$ such that $$\left(\v_{[1]d}^\T\C^{\frac{1}{2}}\hat\bbeta_{[1]}\right)^2+\left(\v_{[2]d}^\T\C^{\frac{1}{2}}\hat\bbeta_{[2]}\right)^2=O(p^{-1})$$ and $\mathcal{A}^{\rm c}$ its complementary set. We notice first that
\begin{align}
\label{eq:set Ac joint distr}
    \left(\v_d^\T\begin{bmatrix}\Z\tilde\c_{[1]}/n\\\Z\tilde\c_{[2]}/n\end{bmatrix}+\v_d^\T\begin{bmatrix}\theta_1\C^{\frac{1}{2}}\hat\bbeta_{[1]}\\\theta_2\C^{\frac{1}{2}}\hat\bbeta_{[1]}\end{bmatrix}-s_d\right)^2=o\left(\left(\v_{[1]d}^\T\C^{\frac{1}{2}}\hat\bbeta_{[1]}\right)^2+\left(\v_{[2]d}^\T\C^{\frac{1}{2}}\hat\bbeta_{[2]}\right)^2\right)
\end{align}
for $d\in\mathcal{A}^{\rm c}$. Indeed, applying \eqref{eq:beta^d(i)}, we have that \begin{align*}
    \left\Vert\hat\bbeta_{[k]}- \hat\bbeta_{[k]}^{\{d(i)\}}\right\Vert^2=O\left(p^{-1}\left(\v_{[k]d}^\T\C^{\frac{1}{2}}\hat\bbeta_{[k]}\right)^2\right)
\end{align*}
for $k=\{1,2\}$, where $\hat\bbeta_{[k]}^{\{d(i)\}}$ stands for approximations of $\hat\bbeta_{[k]}$ obtained by replacing $i$-th feature vector $\x_i$ with $y_i\bmu+y_i\C^{\frac{1}{2}}\P_{\v_{[k]d}}\z_i$ for $\P_{\v_{[k]d}}$ the projection matrix orthogonal to $\v_{[k]d}$. We denote by $\left(\hat\bbeta_{[1]-j}^{\{d(i)\}},\hat\bbeta_{[2]-j}^{\{d(i)\}}\right)$ the leave-one-out version of $\left(\hat\bbeta_{[1]}^{\{d(i)\}},\hat\bbeta_{[2]}^{\{d(i)\}}\right)$. Obviously, $\hat\bbeta_{[k]-j}^{\{d(i)\}}$ is independent of both $\v_{[k]d}^\T\z_i$ and $\v_{[k]d}^\T\z_j$, and $$\left\Vert\hat\bbeta_{[k]-j}- \hat\bbeta_{[k]-j}^{\{d(i)\}}\right\Vert^2=O\left(p^{-1}\left(\v_{[k]d}^\T\C^{\frac{1}{2}}\hat\bbeta_{[k]}\right)^2\right).$$ 
As $\E[\tilde c_{[k]}\v_{[k]d}^\T\z_i+\theta_k\v_{[k]d}^\T\C^{\frac{1}{2}}\hat\bbeta_{[k]}]=O(p^{-\frac{1}{2}})$, similarly to \eqref{eq:civdzi Ac}, we observe that
    \begin{align*}
        \left(\frac{1}{n}\sum_{i=1}^n\tilde c_{[k]}\v_{[k]d}^\T\z_i+\theta_k\v_{[k]d}^\T\C^{\frac{1}{2}}\hat\bbeta_{[k]}\right)^2=O(p^{-\frac{1}{2}}\v_{[k]d}^\T\C^{\frac{1}{2}}\hat\bbeta_{[k]}).
    \end{align*} Since $s_d==O(p^{-1})$, we obtain directly \eqref{eq:set Ac joint distr}. 
    
We are now interested in showing that, for $d\in\mathcal{A}$,
\begin{align}
\label{eq:set A joint distr}
    \left(\v_d^\T\begin{bmatrix}\Z\tilde\c_{[1]}/n\\\Z\tilde\c_{[2]}/n\end{bmatrix}+\v_d^\T\begin{bmatrix}\theta_1\C^{\frac{1}{2}}\hat\bbeta_{[1]}\\\theta_2\C^{\frac{1}{2}}\hat\bbeta_{[1]}\end{bmatrix}-s_d\right)^2=o\left(p^{-1}\right)
\end{align} Remark from the equation~\eqref{eq:tilde beta} in Theorem~\ref{theo:main} that 
$$\v^\T\C^{\frac{1}{2}}\hat\bbeta_{[k]}=O\left(\max\{\v^\T\C^{-\frac{1}{2}}\bmu,p^{-\frac{1}{2}}\}\right).$$ It follows that, for $d\in\mathcal{A}$, $\v_{[1]d}^\T\C^{-\frac{1}{2}}\bmu=O(p^{-\frac{1}{2}})$ and $\v_{[2]d}^\T\C^{-\frac{1}{2}}\bmu=O(p^{-\frac{1}{2}})$, which entails
\begin{align*}
    \v_{[k_1]d}^\T\C^{\frac{1}{2}}\hat\bbeta_{[k_2]}=O(p^{-\frac{1}{2}})
\end{align*}
for $k_1,k_2=\{1,2\}$. We have thus from \eqref{eq:beta^d}
that \begin{align*}
    \Vert\hat \bbeta_{[k]} - \hat \bbeta_{[k]}^{\{d\}}\Vert=O(p^{-\frac{1}{2}})
\end{align*}
where $\hat \bbeta_{[k]}^{\{d\}}$ is obtained by replacing all $\x_i$ with $y_i\bmu+y_i\C^{\frac{1}{2}}\P_{\v_{[1]d},\v_{[2]d}}\z_i$ for $\P_{\v_{[1]d},\v_{[2]d}}$ the projection matrix orthogonal to $\v_{[1]d},\v_{[2]d}$. Clearly $\hat \bbeta_{[1]}^{\{d\}},\hat \bbeta_{[2]}^{\{d\}}$ are independent of $\v_{[1]d}^\T\z_i,\v_{[2]d}^\T\z_i$ for all $i\in\{1,\ldots,n\}$. We get straightforwardly by adapting \eqref{eq:civdzi} that
\begin{align*}
	\frac{1}{n}\sum_{i=1}^n\tilde c_{[k]i}\v_{[k]d}^\T\z_i+\theta_k\v_{[k]d}^\T\C^{\frac{1}{2}}\hat\bbeta_{[k]}-\frac{1}{n}\sum_{i=1}^n h\left( \hat \bbeta^{\{d\}\T}_{[k]-i}\left(\bmu+\C^{\frac{1}{2}}\P_{\v_{[1]d},\v_{[2]d}}\z_i\right)\right)\v_{[k]d}^\T\z_i=O(p^{-\frac{3}{4}}).
\end{align*}
As all $h\left( \hat \bbeta^{\{d\}\T}_{[k]-i}\left(\bmu+\C^{\frac{1}{2}}\P_{\v_{[1]d},\v_{[2]d}}\z_i\right)\right)$ are independent of all $\v_{[k]d}^\T\z_i$, we get by the central limit theorem that
\begin{align*}
\sum_{k=1}^2\frac{1}{n}\sum_{i=1}^n\tilde c_{[k]i}\v_{[k]d}^\T\z_i+\theta_k\v_{[k]d}^\T\C^{\frac{1}{2}}\hat\bbeta_{[k]}=&\sum_{k=1}^2\frac{1}{n}\sum_{i=1}^n h\left( \hat \bbeta^{\{d\}\T}_{[k]-i}\left(\bmu+\C^{\frac{1}{2}}\P_{\v_{[1]d},\v_{[2]d}}\z_i\right)\right)\v_{[k]d}^\T\z_i+O(p^{-\frac{3}{4}})\\
=&s_d+o(p^{-\frac{1}{2}}),
\end{align*}
leading directly to \eqref{eq:set A joint distr}. We get therefore \eqref{eq:proof joint distr} by combining \eqref{eq:set Ac joint distr} and \eqref{eq:set A joint distr}.

\section{Proof of Proposition~\ref{prop:omega}}
	\label{sm:proof-of-proposition-omega}
	Recall from the proof of Theorem~\ref{theo:main} in Appendix~\ref{sm:proof-of-theorem-main} that $\theta$ is positive. We proceed now to establish $\max_{\lambda_{\rm LS}\geq 0}  \lambda_{\rm LS}/\theta_{\rm LS}\geq \max_{\ell,\lambda\geq 0}\lambda/\theta.a$. As it is easy to check $\lim_{\lambda\to\infty}\Vert\hat\bbeta\Vert\to0$ and see from \eqref{eq:kappa} that $\lim_{\lambda\to\infty}\kappa\to0$, which implies $\lim_{\lambda\to\infty}h(t)-(-\ell'(t))\to0$ for bounded $t$, we observe from \eqref{eq:kappa} that $\lim_{\lambda\to\infty}\theta<+\infty$. Consequently, $\lambda/\theta$ goes to infinity with $\lambda$, we prove thus $$\max_{\lambda_{\rm LS}\geq 0}  \lambda_{\rm LS}/\theta_{\rm LS}\geq \max_{\ell,\lambda\geq 0}\lambda/\theta$$.
    
    It remains to show that $\min_{\lambda_{\rm LS}\geq 0}  \lambda_{\rm LS}/\theta_{\rm LS}\leq \min_{\ell,\lambda\geq 0}\lambda/\theta$. First, notice from \eqref{eq:kappa} that when $n/p>1$, $\kappa$ has a bounded positive value at $\lambda=0$ and with the square loss $\ell(t)=(1-t)^2/2$, for which we also have $$h(r)=\frac{1-r}{1+\kappa},$$
    leading to 
    $$\theta=\frac{-\cov[h(r),r]}{\var[r]}=\frac{1}{1+\kappa}>0.$$ Since $\lambda/\theta$ is never negative, it is obvious that $ \lambda_{\rm LS}/\theta_{\rm LS}$ achieves the minimal possible value of zero at $\lambda_{\rm LS}=0$ in the case of $n/p>0$, i.e.,
    $$\min_{\lambda_{\rm LS}\geq 0}  \lambda_{\rm LS}/\theta_{\rm LS}=0\leq \min_{\ell,\lambda\geq 0}\lambda/\theta$$
    holds for $n/p>0$. We are now left with the case of $n/p\leq 1$. It is important to remark that when $n/p\leq 1$, $\kappa$ goes to infinity as $\lambda$ goes to zero, regardless of the choice of $\ell$. To see this, we suppose that the contrary is true, then at $\lambda=0$, we have from \eqref{eq:kappa} that
    $$	\kappa =\frac{p}{n} \theta=\kappa\frac{p}{n} \left(1-\frac{\cov[g_{\kappa, \ell} (r),r]}{\var[r]}\right)^{-1}$$
    since 
    $$	\theta=\frac{-\cov[h(r),r]}{\var[r]}=\frac{1}{\kappa}-\frac{\cov[g_{\kappa,\ell}(r),r]}{\kappa\var[r]}.$$
    This implies $1-\frac{\cov[g_{\kappa, \ell} (r),r]}{\var[r]}\geq 1$ since $n/p\leq 1$, which does not hold as $\cov[g_{\kappa, \ell} (r),r]>0$ since $g_{\kappa, \ell} (r)$ is an increasing function of $r$. We note thus by contradiction that when $n/p\leq 1$, $\kappa$ indeed always goes to infinity in the limit of $\lambda\to 0$. To discuss the minimum of $\lambda/\theta$ for $n/p\leq 1$, it is useful to have the following inequalities:
    \begin{align}
    &\theta\leq \frac{1}{\kappa}\label{eq:inequality theta kappa}\\
    &\kappa\lambda\geq (\kappa\lambda)^*\label{eq:inequality kappa_lambda}
    \end{align}
    where $(\kappa\lambda)^*>0$ is give by $$ 1 = \frac1n \tr \C \left[\C +(\kappa\lambda)^* \I_p \right]^{-1}.$$
    It is easy to see \eqref{eq:inequality theta kappa} from
    \begin{align*}
    	\theta=\frac{-\cov[h(r),r]}{\var[r]}=\frac{1}{\kappa}-\frac{\cov[g_{\kappa,\ell}(r),r]}{\kappa\var[r]}.
    \end{align*}
    as $g_{\kappa,\ell}(r)$ is an increasing function of $r$, $\cov[g_{\kappa,\ell}(r),r]\geq 0$. As for \eqref{eq:inequality kappa_lambda}, it is a direct consequence of \eqref{eq:kappa}. Indeed, rewriting \eqref{eq:kappa} as $$ \kappa = \kappa\frac1n \tr \C \left[ \left(1-\frac{\cov[g_{\kappa, \ell} (r),r]}{\var[r]}\right)\C + \kappa\lambda\I_p \right]^{-1},$$ it is easy to see that the inequality \eqref{eq:kappa} holds since $$\frac1n \tr \C \left[ \left(1-\frac{\cov[g_{\kappa, \ell} (r),r]}{\var[r]}\right)\C + \kappa\lambda\I_p \right]^{-1}=1$$ and $1-\frac{\cov[g_{\kappa, \ell} (r),r]}{\var[r]}\leq 1$. Combining the two inequalities, we get $$\frac{\lambda}{\theta}\geq \kappa\lambda\geq (\kappa\lambda)^*.$$
   As $\kappa\to\infty$ at $\lambda\to 0$, it is a simple matter to check that $$\lim_{\lambda_{\rm LS}\to0}\frac{\lambda_{\rm LS}}{\theta_{\rm LS}}=(\kappa\lambda)^*.$$ The proof of Proposition~\ref{prop:omega} is thus completed.  

\section{Proof of Theorem~\ref{theo:universality over non-gaussian data}}
	\label{sm:proof-of-theorem-unitersality}
	
	As many intermediary results demonstrated in the proof of Theorem~\ref{theo:main} (provided in Appendix~\ref{sm:proof-of-theorem-main}) do not require the Gaussianity of data, we will make use of them to prove that the non-Gaussian and Gaussian data have the same expression of classification error in the limit of large $p$. We start by retrieving the following ``leave-one-out'' results that apply directly to non-Gaussian data:
	\begin{align*}
	   &\Vert\hat\bbeta_{-i}-\hat\bbeta\Vert=O(p^{-\frac{1}{2}})\\
	   &y_i\x_i^\T\hat\bbeta = g_{\kappa,\ell}\left(y_i\x_i^\T\hat\bbeta_{-i}\right)+O(p^{-\frac{1}{2}})\\
	   &c_i=-\ell'(y_i\x_i^\T\hat\bbeta)\simeq \tilde c_i=h\left(y_i\x_i^\T\hat\bbeta_{-i}\right)
	\end{align*}
	where we recall that $\hat\bbeta_{-i}$ is the leave-one-out version of $\hat\bbeta$, obtained by removing the $i$-th sample $(\x_i, y_i)$ from the training set, and $h(t)=\frac{g_{\kappa,\ell}(t)-t}{\kappa}$.
	
	The universality over non-Gaussian distributions relies on the key fact that all the elements in the vector $\V\hat\bbeta$ are of the order $O(p^{-1/2})$, where we recall $\C=\V\bLambda^2\V^\T$ is the eigen-decomposition of the covariance matrix $\C$. This is ensured by Assumption~\ref{ass:non-sparsity}. Indeed, since for any $d\in\{1,\ldots,p\}$
	\begin{align*}
		 \frac1n \sum_{i=1}^n \ell(y_i\x_i^\T \hat\bbeta) + \frac{\lambda}2 \| \hat\bbeta \|^2\leq  \frac1n \sum_{i=1}^n \ell\bigg(y_i\x_i^\T \sum_{d'\neq d}\v_{d'}\v_{d'}^\T\hat\bbeta\bigg) + \frac{\lambda}2  \sum_{d'\neq d}(\v_{d'}^\T\hat\bbeta)^2,
	\end{align*}
	we have that
	\begin{align*}
	    \frac{\lambda}2(\v_{d}^\T\hat\bbeta)^2\leq  \frac1n \sum_{i=1}^n \ell\bigg(y_i\x_i^\T \sum_{d'\neq d}\v_{d'}\v_{d'}^\T\hat\bbeta\bigg)-\frac1n \sum_{i=1}^n \ell(y_i\x_i^\T \hat\bbeta).
	\end{align*}
	By the mean value theorem, we observe that
	\begin{align*}
	    \frac1n \sum_{i=1}^n \ell\bigg(y_i\x_i^\T \sum_{d'\neq d}\v_{d'}\v_{d'}^\T\hat\bbeta\bigg)-\frac1n \sum_{i=1}^n \ell(y_i\x_i^\T \hat\bbeta=-\frac1n \sum_{i=1}^n \ell'(a_i) y_i\x_i^\T\v_{d}\v_{d}^\T\hat\bbeta
	\end{align*}
for  $a_i$ some value between $y_i\x_i^\T \hat\bbeta$ and $y_i\x_i^\T \sum_{d'\neq d}\v_{d'}\v_{d'}^\T\hat\bbeta$. Also by the fact that $\ell'(t)$ is an increasing function, we observe that 
	\begin{equation*}
	    -\frac1n \sum_{i=1}^n \ell'(a_i) y_i\x_i^\T\v_{d}\v_{d}^\T\hat\bbeta\leq -\frac1n \sum_{i=1}^n \ell'\bigg(y_i\x_i^\T \sum_{d'\neq d}\v_{d'}\v_{d'}^\T\hat\bbeta\bigg) y_i\x_i^\T\v_{d}\v_{d}^\T\hat\bbeta
	\end{equation*}
	
  A similarly reasoning for showing $y_i\x_i^\T\hat\bbeta_{-i}=y_i\x_i^\T\hat\bbeta+\kappa \ell'(y_i\x_i^\T\hat\bbeta)+O(p^{-\frac{1}{2}})$ leads directly to \begin{align*}
      y_i\x_i^\T\sum_{d'\neq d}\v_{d'}\v_{d'}^\T\hat\bbeta = g_{\kappa,\ell}\left(y_i\x_i^\T\sum_{d'\neq d}\v_{d'}\v_{d'}^\T\hat\bbeta_{-i}\right)+O(p^{-\frac{1}{2}})
  \end{align*}
  Notice that under \eqref{eq:general mixture},
  \begin{align*}
      y_i\x_i^\T\v_{d}=\bmu^\T\v_{d}+[\bLambda]_{dd}y_i[\z_i]_d,
  \end{align*}
  $y_i\x_i^\T\sum_{d'\neq d}\v_{d'}\v_{d'}^\T\hat\bbeta_{-i}$ is thus independent of $y_i\x_i^\T\v_{d}$ as $y_i\x_i^\T\sum_{d'\neq d}\v_{d'}$ is independent of $y_i\x_i^\T\v_{d}$ by the independence between $[\z_i]_1,\ldots,[\z_i]_p$ and $\hat\bbeta_{-i}$ is independent of $\x_i$ by definition. Write
  \begin{align*}
      \x^{\{d\}}_i=y_i\bmu+\V\bLambda{\rm diag}([\underbrace{1,\ldots,1}_{d-1},0,\underbrace{1,\ldots,1}_{p-d-1}])\z_i
  \end{align*}
  and define $\hat\bbeta^{\{d(i)\}}$ be obtained by replacing $\x_i$ with $\x^{\{d\}}_i$. We retrieve again from the proof of Theorem~\ref{theo:main} that
  \begin{align*}
      \Vert\hat\bbeta-\hat\bbeta^{\{d(i)\}}\Vert=O(\v_d^\T\hat\bbeta/\sqrt{p}).
  \end{align*} Let $\hat\bbeta^{\{d(i)\}}_{-j}$ be the leave-one-out version of $\hat\bbeta^{\{d(i)\}}$ (which enjoys similarly the properties of leave-one-out solutions), we obtain that
  \begin{align*}
      &\E\left[\left(\frac1n \sum_{i=1}^n \ell'\bigg(g_{\kappa,\ell}\Big(y_i\x_i^\T\sum_{d'\neq d}\v_{d'}\v_{d'}^\T\hat\bbeta_{-i}\Big)\bigg)y_i[\z_i]_d\right)^2\right]\\
      =&\frac{1}{n^2}\sum_{i,j=1}^n\E\left[ \ell'\bigg(g_{\kappa,\ell}\Big(y_i\x_i^\T\sum_{d'\neq d}\v_{d'}\v_{d'}^\T\hat\bbeta_{-i}\Big)\bigg)\ell'\bigg(g_{\kappa,\ell}\Big(y_j\x_j^\T\sum_{d'\neq d}\v_{d'}\v_{d'}^\T\hat\bbeta_{-j}\Big)\bigg)y_iy_j[\z_i]_d[\z_j]_d\right]\\
      =&\frac{1}{n^2}\sum_{i\neq j}\E\left[ \ell'\bigg(g_{\kappa,\ell}\Big(y_i\x_i^\T\sum_{d'\neq d}\v_{d'}\v_{d'}^\T\hat\bbeta^{\{d(j)\}}_{-i}\Big)\bigg)\ell'\bigg(g_{\kappa,\ell}\Big(y_j\x_j^\T\sum_{d'\neq d}\v_{d'}\v_{d'}^\T\hat\bbeta^{\{d(i)\}}_{-j}\Big)\bigg)y_iy_j[\z_i]_d[\z_j]_d\right]\\
      &+O(\v_d^\T\hat\bbeta/\sqrt{p})=O(\v_d^\T\hat\bbeta/\sqrt{p}).
  \end{align*}
  Therefore
  \begin{align*}
      \frac1n \sum_{i=1}^n \ell'\bigg(y_i\x_i^\T \sum_{d'\neq d}\v_{d'}\v_{d'}^\T\hat\bbeta\bigg) y_i\x_i^\T\v_{d}=&\frac1n \sum_{i=1}^n \ell'\bigg(g_{\kappa,\ell}\Big(y_i\x_i^\T\sum_{d'\neq d}\v_{d'}\v_{d'}^\T\hat\bbeta_{-i}\Big)\bigg) y_i\x_i^\T\v_{d}+O(p^{-\frac{1}{2}})\\
      =&\frac1n \sum_{i=1}^n \ell'\bigg(g_{\kappa,\ell}\Big(y_i\x_i^\T\sum_{d'\neq d}\v_{d'}\v_{d'}^\T\hat\bbeta_{-i}\Big)\bigg) [\bLambda]_{dd}y_i[\z_i]_d+O(p^{-\frac{1}{2}})\\
      =&O\left(\max\{p^{-\frac{1}{2}},p^{-\frac{1}{4}}\v_d^\T\hat\bbeta^{\frac{1}{2}}\}\right).
  \end{align*}
  We get finally that
  \begin{align}
  \label{eq:vbeta}
      \v_d^\T\hat\bbeta=O(p^{-\frac{1}{2}})
  \end{align}
  for any $d\in\{1,\ldots,p\}$. With \eqref{eq:vbeta}, it is easy to see that, when conditioned on $\hat\bbeta_{-i}$,  $$y_i\x_i^\T\hat\bbeta_{-i}=r+O(p^{-\frac{1}{2}})$$
  for some $r\sim \mathcal{N}(\bmu^\T\hat\bbeta,\hat\bbeta^\T\C\hat\bbeta)$. This allows us to retrieve \eqref{eq:E hr}--\eqref{eq:var r} by the same reasoning in Appendix~\ref{sm:proof-of-theorem-main}. To obtain \eqref{eq:expectation ciwi}, notice first that \begin{align*}
      \E[\tilde c_i\w_i~\vert~\hat\bbeta_{-i}]=\V\bLambda\E\left[h\left(y_i\x_i^\T\hat\bbeta_{-i}\right)y_i\z_i~\vert~\hat\bbeta_{-i}\right].
  \end{align*}
  where we recall $\w_i=y_i\x_i-\bmu$. Since $\v_d^\T\hat\bbeta=O(p^{-\frac{1}{2}})$,  we have that
  \begin{align*}
      &\E\left[h(y_i\x_i^\T\hat\bbeta_{-i})y_i[\z_i]_d~\vert~\hat\bbeta_{-i}\right]\\
      =&\E\left[\left(h\Big(y_i\x_i^\T\sum_{d'\neq d}\v_{d'}\v_{d'}^\T\hat\bbeta_{-i}\Big)+h'\Big(y_i\x_i^\T\sum_{d'\neq d}\v_{d'}\v_{d'}^\T\hat\bbeta_{-i}\Big)(\v_{d}^\T\bmu+[\bLambda]_{dd}y_i[\z_i]_d)\right)\v_{d}^\T\hat\bbeta_{-i}y_i[\z_i]_d~\vert~\hat\bbeta_{-i}\right]\\
      &+O(p^{-1})\\
      =&\E\left[h'\Big(y_i\x_i^\T\sum_{d'\neq d}\v_{d'}\v_{d'}^\T\hat\bbeta_{-i}\Big)[\bLambda]_{dd}y_i[\z_i]_d y_i[\z_i]_d\v_{d}^\T\hat\bbeta_{-i}~\vert~\hat\bbeta_{-i}\right]+O(p^{-1})\\
      =&\E[h'(r)][\bLambda]_{dd}\v_{d}^\T\hat\bbeta+O(p^{-1}),
  \end{align*}
  leading to  
  \begin{align*}
      \E[\tilde c_i\w_i~\vert~\hat\bbeta_{-i}]=\E[h'(r)]\V\bLambda^2\V^\T\hat\bbeta+O(p^{-1}).
  \end{align*}
  As 
  $$\E[h'(r)]=\E[h(r)](r-\E[r])/\var[r]$$
  by the fact that $r$ is normally distributed, we get directly \eqref{eq:expectation ciwi}. 
  
  To prove Theorem~\ref{theo:universality over non-gaussian data}, it remains to show that
  $$\left\Vert\frac{1}{n}\sum_{i=1}^n(\tilde c_i\w_i-\E[\tilde c_i\w_i~\vert~\hat\bbeta_{-i}])\right\Vert^2=\gamma^2\tr\C/p+o(1),$$
  which is equivalent to
  $$\left\Vert\frac{1}{n}\sum_{i=1}^n(\tilde c_i\z_i-\theta\bLambda\V^\T\hat\bbeta\right\Vert^2=\gamma^2+o(1).$$
  Let $\hat\bbeta^{\{d\}}$ stand for the solution to \eqref{eq:opt-origin} with all $(\x_i,y_i)$ replaced by $(\x_i^{\{d\}},y_i)$, we retrieve again from \eqref{eq:beta^d} in Appendix~\ref{sm:proof-of-theorem-main} that
  $$\left\Vert\hat\bbeta-\hat\bbeta^{\{d\}}\right\Vert=O(p^{-\frac{1}{2}})$$. Following by the same derivation as in \eqref{eq:civdzi}, we have that
  $$\frac{1}{n}\sum_{i=1}^n\tilde c_i[\z_i]_d+\theta[\bLambda]_{dd}\v_{d}^\T\hat\bbeta-\frac{1}{n}\sum_{i=1}^n h(y_i \x_i^{\{d\}\T} \hat \bbeta^{\{d\}}_{-i})[\z_i]_d=O(p^{-\frac{3}{4}}).$$
  As $h(y_i \x_i^{\{d\}\T} \hat \bbeta^{\{d\}}_{-i})[\z_i]_d$, $d\in\{1,\ldots,p\}$, are independent, by the central limit theorem,
  $$\left(\frac{1}{n}\sum_{i=1}^n\tilde c_i[\z_i]_d+\theta[\bLambda]_{dd}\v_{d}^\T\hat\bbeta\right)^2=\frac{1}{n}\E[h^2(y_i \x_i^{\{d\}\T} \hat \bbeta^{\{d\}}_{-i})]+O(p^{-\frac{3}{2}})=\frac{1}{p}\gamma^2+O(p^{-\frac{3}{2}}),$$
  which completes the proof of Theorem~\ref{theo:universality over non-gaussian data}.

\end{document}